\documentclass[3p,times, twocolumn]{elsarticle}

\usepackage{graphicx}
\usepackage[english]{babel}
\usepackage[T1]{fontenc}
\usepackage{amsmath}
\usepackage{subfig}
\usepackage{caption}
\usepackage{amsfonts}
\usepackage{amssymb}
\usepackage{flushend} % <- to "balance" the last page
\usepackage{amsthm}
\usepackage{cases}
\usepackage{cool}
\usepackage{hyperref}
\usepackage[dvipsnames]{xcolor}
\usepackage{tikz}
\usepackage{epsfig}
\usepackage{cool} % for the pderiv command
\usepackage{nicefrac}
\usepackage{algpseudocode}
\usepackage{algorithm}
\usepackage{mathtools}
\usepackage{bigints}
\usepackage{multirow}
\algnewcommand{\LeftComment}[1]{\Statex \(\triangleright\) #1}

\newcommand{\RR}{\mathbb{R}}

\newcommand{\br}[1]{\left( #1 \right)}
\newcommand{\vect}[1]{\mathbf{#1}}
\newcommand{\mtrx}[1]{\mathbf{#1}}
\newcommand{\norm}[1]{\left\| #1 \right\|}

\newcommand{\set}[1]{\left\{ #1 \right\}}

\newcommand{\diff}{\mathop{}\!d}
\newcommand{\cond}{\mathop{\normalfont{\text{cond}}}}
\newcommand{\defas}{\stackrel{\mathclap{\normalfont\mbox{\tiny{def}}}}{=}}
\usepackage{bm}
\newcommand{\vectgreek}[1]{\boldsymbol{#1}}
\newcommand{\transpose}{^{\mathsf{T}}}
\newcommand{\ee}{\mathop{\normalfont{e}}}

\def\mK{\mathbf{K}}
\def\mD{\mathbf{D}}

\def\midentity{\text{\normalfont{\textbf{I}d}}}

\theoremstyle{remark}
\newtheorem{rmk}{Remark}

\begin{document}

\begin{frontmatter}

\title{Overcoming Some Drawbacks of Dynamic Movement Primitives}

\author[label1]{Michele Ginesi\corref{cor1}}
\author[label1]{Nicola Sansonetto}
\author[label1]{Paolo Fiorini}
\address[label1]{Department of Computer Science, University of Verona, Strada Le Grazie 15, Verona, Italy}
\cortext[cor1]{Corresponding author}

\begin{abstract}
    Dynamic Movement Primitives (DMPs) is a framework for learning a point-to-point trajectory from a demonstration.
    Despite being widely used, DMPs still present some shortcomings that may limit their usage in real robotic applications.
    Firstly, at the state of the art, mainly Gaussian basis functions have been used to perform function approximation.
    Secondly, the adaptation of the trajectory generated by the DMP heavily depends on the choice of hyperparameters and the new desired goal position.
    Lastly, DMPs are a framework for `one-shot learning', meaning that they are constrained to learn from a unique demonstration.
    In this work, we present and motivate a new set of basis functions to be used in the learning process, showing their ability to accurately approximate functions while having both analytical and numerical advantages w.r.t. Gaussian basis functions.
    Then, we show how to use the invariance of DMPs w.r.t. affine transformations to make the generalization of the trajectory robust against both the choice of hyperparameters and new goal position, performing both synthetic tests and experiments with real robots to show this increased robustness.
    Finally, we propose an algorithm to extract a common behavior from multiple observations, validating it both on a synthetic dataset and on a dataset obtained by performing a task on a real robot.
\end{abstract}


\end{frontmatter}

% ---------------------------------------------------------------------------- %
% INTRODUCTION
% ---------------------------------------------------------------------------- %

\section{Introduction}

The recent improvements in robot dexterity have given rise to increasing attention to Learning from Demonstration (LfD) approaches to make robots faithfully mimic human motions.

\noindent
Dynamic Movement Primitive (DMP) \cite{INS02, INS03, Sch06, INHPS13} is one of the most used frameworks for trajectory learning from a single demonstration.
They are based on a system of second-order Ordinary Differential Equations (ODEs), in which a \emph{forcing term} can be ``learned'' to encode the desired trajectory.
This approach has already been proven effective in teaching a robot how to perform some human task as, for instance, (table) tennis swing \cite{INS02, MHM10}, to play drums \cite{UGAM10}, to write \cite{KNTW12, WWCT16}, and to perform surgical-related tasks \cite{GMNRF19, GMRSF20a}.
The framework of DMPs has been shown to be flexible and robust enough to allow learning sensory experience \cite{PRKS11, PKRS12, KPKWS15}, handling obstacle avoidance \cite{PHPS08, HPPS09, GMCDSF19, GMRSF20}, describing bi-manual tasks \cite{GNIU14}, learning orientations \cite{UNPM14, SFL19}, and working in scenarios with human-robot interaction \cite{ANKKP14}.

Despite their wide use, DMPs approach has still some shortcomings that need to be fixed to obtain a more robust framework.

In this work, we discuss three aspects of DMPs and propose modifications that guarantee a more robust implementation of the framework.
In more detail, we discuss different classes of basis functions to be used instead of the Gaussian radial basis functions.
Then, we show how to exploit the invariance of DMPs with respect to particular transformations in order to make the generalization of the learned trajectory more robust.
Finally, we present an algorithm to learn a unique DMP from multiple observations without the need to rely on probabilistic approaches or additional parameters.

\noindent
We remark that probabilistic approaches for trajectory learning can deal with generalization and learning from multiple demonstrations.
These approaches include, for instance, \emph{Probabilistic Movement Primitives} \cite{PDPN13}, \emph{Kernelized Movement Primitives} \cite{HRSC19}, and \emph{Gaussian Mixture Models} \cite{Cal16}.
However, these methods necessarily require multiple trajectories to extract a common behavior, and, differently from DMPs, cannot be used as one-shot learning frameworks.
Moreover, generalization is heavily limited by the quality of the dataset.
Indeed, if the dataset is not descriptive enough of the task and all the different scenarios, the learned behavior may fail to generalize in certain situations.
Finally, these methods have not only a probabilistic learning phase but also a stochastic execution.
This aspect makes them not suitable in some critical scenarios, such as Robotic Minimally Invasive Surgery.
On the other hand, DMP is a completely deterministic approach.
Thus, the improvements we present in this work allow to extend the DMP framework adding some of the strong points of probabilistic frameworks while maintaining the strengths of a deterministic approach.

The work is organized as follows.
In Section~\ref{sec:overview} we review the two classical formulations of DMPs, emphasizing their shortcomings.
In Section~\ref{sec:basis} we review multiple classes of basis functions and introduce a new one, which has both the desirable properties of being smooth and compactly supported.
We then test and compare various numerical aspects of all the sets of basis functions presented, showing that our proposed one gives rise to a numerically more stable and faster learning phase.
In Section~\ref{sec:linear_invariance} we show how to generalize the DMPs to any new starting and goal positions by exploiting the invariance property of DMPs under affine transformations.
In Section~\ref{sec:regression} we present and test a novel algorithm to learn a unique DMP from a set of observations.
Lastly, in Section~\ref{sec:conclusions} we present the conclusions.

Our implementation of DMPs, written in Python 3.8, is available at the link \url{https://github.com/mginesi/dmp_pp}.
Together with the implementation of DMPs and our extensions, the repository contains the scripts to run all the tests presented in Sections~\ref{sec:basis_results},~\ref{sec:invariance_results}, and \ref{sec:regression_results}, together with the tests that are mentioned but not shown.

% ---------------------------------------------------------------------------- %
% OVERVIEW
% ---------------------------------------------------------------------------- %

\section{Dynamic Movement Primitives: an Overview}\label{sec:overview}

DMPs are used to model both periodic and discrete movements \cite{INS02, INS03, Sch06}.
However, in this work, we will focus on the latter.

\noindent
They consist of a system of second-order ODEs (one for each dimension of the ambient space) of mass-spring-damper type with a forcing term.
DMPs aim to model the forcing term in such a way to be able to generalize the trajectory to new start and goal positions while maintaining the shape of the learned trajectory.

The one-dimensional formulation of DMPs is \cite{INS02, INS03, Sch06}:
\begin{subnumcases}{\label{eqs:dmp_old_form}}
    \tau \dot{v} = K (g - x) - Dv + (g- x_0)f(s) \label{eq:dmp_old_form_acc} \\
    \tau \dot{x} = v
\end{subnumcases}
where $ x , v \in \RR $ are, respectively, position and velocity of a prescribed point of the system.
$x_0 \in \RR$ is the initial position, and $ g \in \RR $ is the \emph{goal}.
Constants $K, D \in \RR^+$ are, respectively, the spring and damping terms, chosen in such a way that the associated homogeneous system is critically damped: \( D = 2 \sqrt{K} \).
Parameter $\tau \in \RR^+$ is a temporal scaling factor, and $f$ is a real-valued, non-linear \emph{forcing} (also called \emph{perturbation}) \emph{term}.
$s \in (0, 1] $ is a re-parametrization of time $t \in [0, T]$, governed by the so-called \emph{canonical system}:
\begin{equation}\label{eq:canonical_system}
    \tau \dot{s} = -\alpha s.
\end{equation}
$\alpha \in \RR^+$ determines the exponential decay of canonical system \eqref{eq:canonical_system}.
The initial value is set to $ s(0) = 1 $.

\noindent
The forcing term $f$ in \eqref{eq:dmp_old_form_acc} is written in terms of basis functions as
\begin{equation}
    \label{eq:forcing_term}
    f (s) = \frac{ \sum_{i=0}^N \omega_i \, \psi_i(s) }{ \sum_{i=0}^N \psi_i(s) } s,
\end{equation}
where
\begin{equation}
    \psi_i (s) = \exp \br{- h_i \, (s - c_i) ^ 2} \label{eq:gaussian_basis_def}
\end{equation}
are \emph{Gaussian Basis Functions} (GBFs) with centers $c_i$ and widths $h_i$.
Centers $ c_i $ are defined as:
\begin{equation}
    \label{eq:basis_centers_def}
    c_i = \exp {\br{ -\alpha \, i \, \frac{T}{N} }}, \quad i = 0, 1, \ldots, N,
\end{equation}
so that they are equispaced in time interval $ [0, T] $.
Widths $ h_i $ \cite{SD20} are defined as
\begin{equation}
    \label{eq:basis_width_gaussian_def}
    \begin{aligned}
        & h_i = \frac{ \tilde{h} }{(c_{i+1} - c_i) ^ {2}}, & i = 0,1,\ldots, N - 1, \\
        & h_{N} = h_{N-1},
    \end{aligned}
\end{equation}
where $ \tilde{h} > 0 $ controls the overlapping between the basis functions.
Usually, $ \tilde{h} $ is set to one, \( \tilde{h} = 1 \) \cite{UNPM14,AK20}.

\noindent
The \emph{learning process} focuses on the computation of the weights $\omega_i \in \RR$ that best approximate the desired forcing term, obtained by solving \eqref{eq:dmp_old_form_acc} for $f$.
Given a desired trajectory $ x(t) , t\in[0, T]$ with velocity $ v(t) $, we set $ \tau=1 , x_0 = x(0),$ and $g = x(T)$.
The forcing term is then computed as
\begin{equation}
    \label{eq:forcin_term_old}
    f(s(t)) = \frac{1}{g-x_0} \Big( \dot{v}(t) - K \big(g - x(t)\big) + Dv \Big),
\end{equation}
where $ s(t) = \exp(-\alpha t) $.
Then, the vector $ \bm{\omega} = [\omega_0, \omega_1, \ldots, \omega_N]\transpose $ can be computed by solving the linear system
\begin{equation} \label{eq:lin_p_1}
    \mtrx{A} \bm{\omega} = \vect{b} ,
\end{equation}
where $\mtrx{A} = [a_{hk}] $ is a $ (N+1) \times (N+1) $ matrix and $ \vect{b} = [b_h] $ is a vector in $ \RR^{N+1} $ with components, respectively,
\begin{subequations}
    \label{eqs:integrals_min_problem}
    \begin{align}
        a_{h k} & = \int_{s(0)}^{s(T)} \frac{\psi_h(s) \psi_k(s)}{\br{ \sum_{i=0}^N \psi_i(s) }^2}\, s^2 \, \diff s , \label{eq:A_component}\\
        b_h & = \int_{s(0)}^{s(T)} \frac{\psi_h(s) }{ \sum_{i=0}^N \psi_i(s) }\, f(s)\, s \, \diff s . \label{eq:b_component}
    \end{align}
\end{subequations}

\begin{figure*}[t]
    \centering
    \subfloat[Slight change in goal position.\label{subfig:ovn_high}]{\includegraphics[width=0.9\columnwidth]{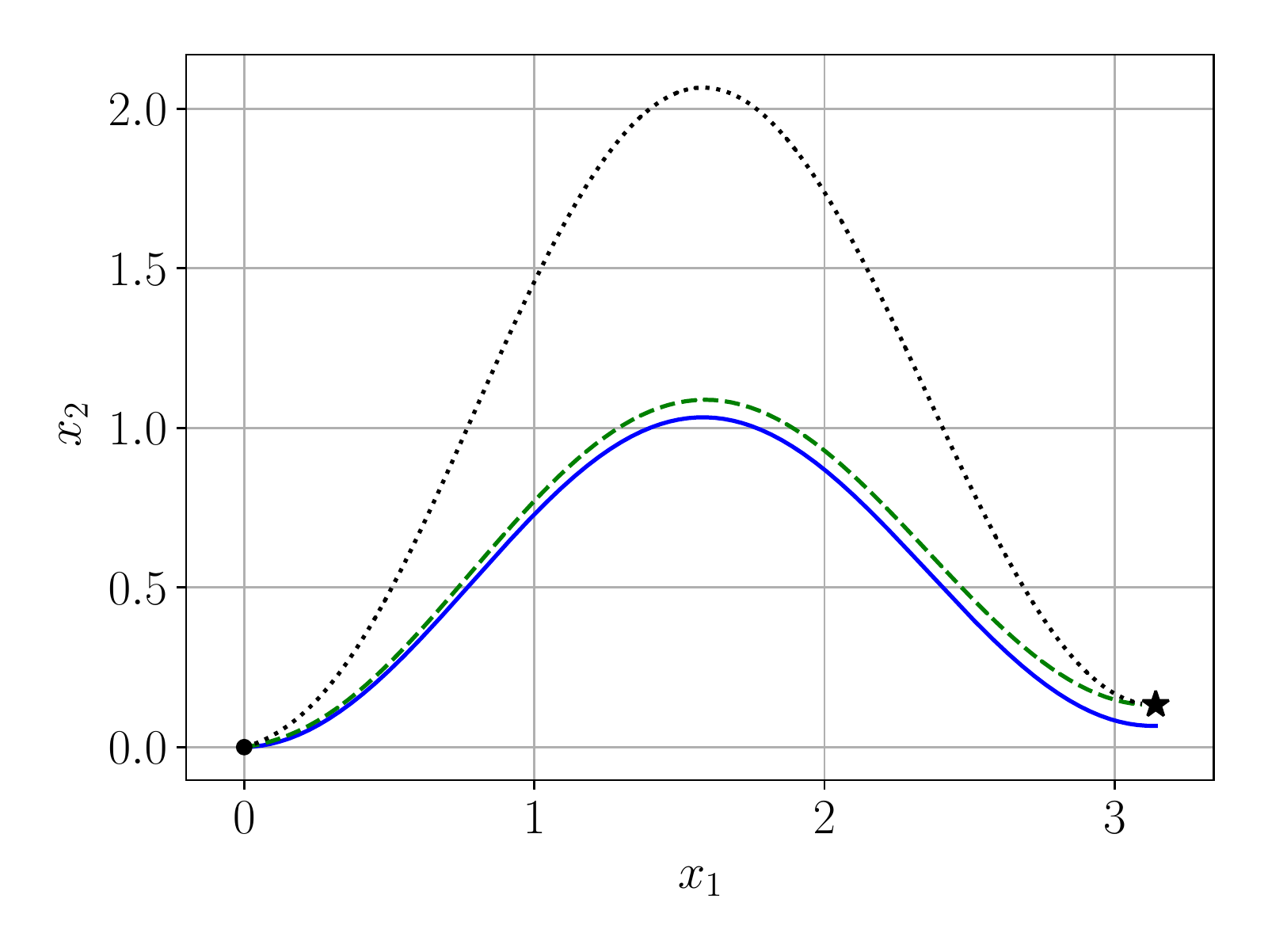}}
    \hspace{1.25cm}
    \subfloat[Change of sign in goal position.\label{subfig:ovn_under}]{\includegraphics[width=0.9\columnwidth]{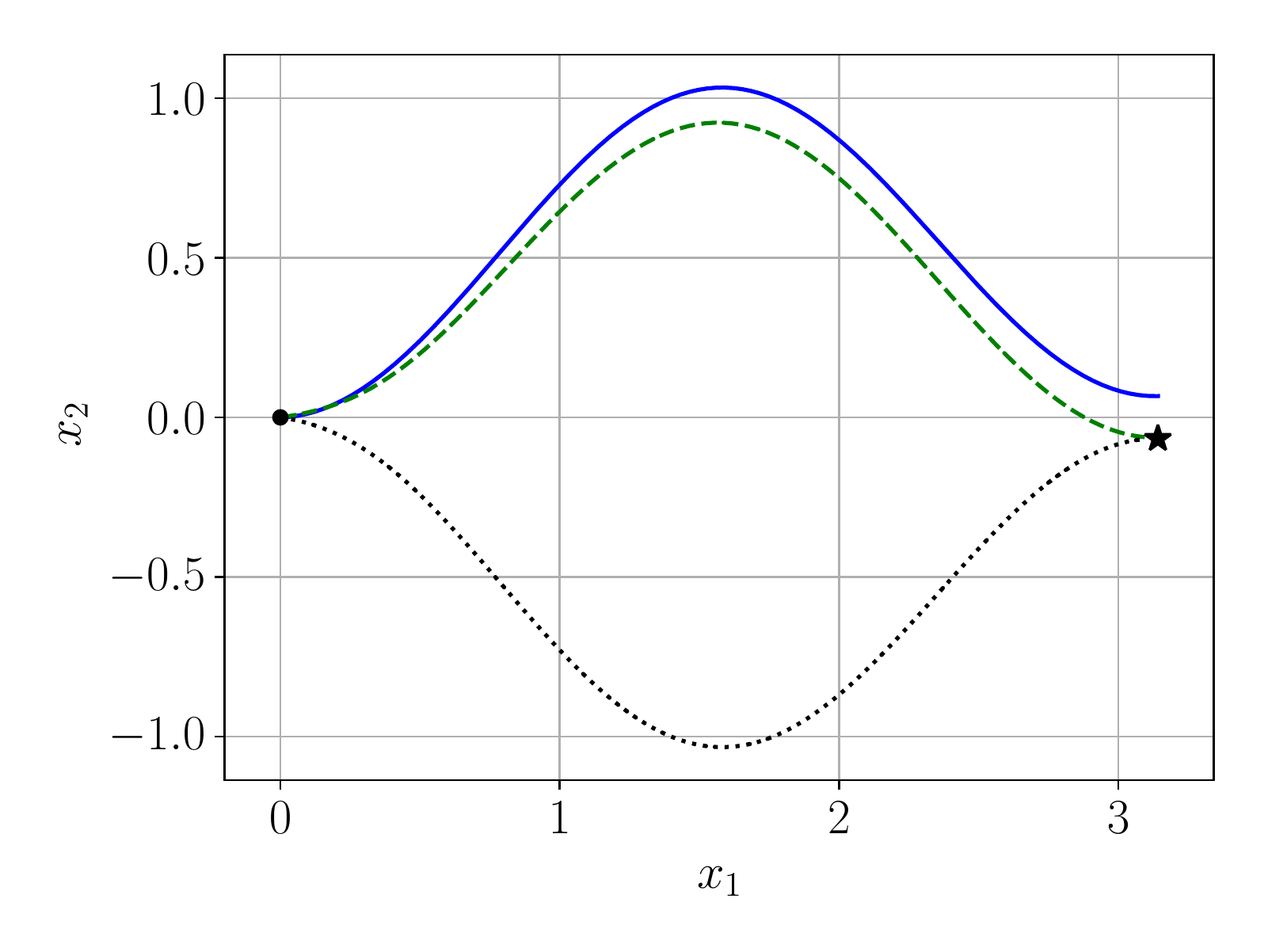}}
    \caption{Comparison between DMPs' formulation \eqref{eqs:old_dmps_vector} and \eqref{eqs:new_dmps_vector}.
    In both plots, the blue solid line shows the learned trajectory.
    The dot and star mark, respectively, the new initial position $ \vect{x}_0 $ (which, for simplicity, in this example is kept the same as the learned one) and new goal position $ \vect{g} $.
    The black dotted line shows the generalization to the new goal using DMP formulation \eqref{eqs:old_dmps_vector}, while the dashed green line shows the generalization using formulation \eqref{eqs:new_dmps_vector}.
    The grid is plotted to emphasize that in Figure~\ref{subfig:ovn_under} the vertical component of the new goal position is of opposite sign than the learned one.}
    \label{fig:old_porblems}
\end{figure*}

When dealing with $d$-dimensional trajectories, we make $d$ decoupled copies of system \eqref{eqs:dmp_old_form}, obtaining the following vector formulation:
\begin{subnumcases}{\label{eqs:old_dmps_vector}}
    \tau \dot{\vect{v}} = \mtrx{K} (\vect{g} - \vect{x}) - \mtrx{D} \vect{v} + (\vect{g} - \vect{x}_0) \odot \vect{f}(s) \label{eq:old_dmp_vect_acc}\\
    \tau \dot{\vect{x}} = \vect{v}
\end{subnumcases}
where \( \vect{x}, \vect{v}, \vect{g}, \vect{x}_0, \vect{f}(s) \in \RR ^ d \) and \( \mtrx{K} , \mtrx{D} \in \RR^{d \times d} \) are diagonal matrices, so to maintain each component decoupled from the others.
The operator $ \odot $ denotes the component-wise multiplication: given $ \vect{v} = [v_i], \vect{w}=[w_i] \in \RR^d $, we define
\( \vect{v} \odot \vect{w} = [v_i w_i]_i .\)

\noindent
Thanks to the decoupling of the system, the forcing term can be learned component by component.

Three main drawbacks characterize DMP formulation \eqref{eqs:old_dmps_vector} (see \cite{HPPS09}).
Firstly, if in any component the goal position coincides with the starting position, $ g = x_0 $, the perturbation term $ f $ clearly does not contribute to the evolution of the dynamical system.
Secondly, if $ g - x_0 $ is ``small'' the scaling of the perturbation term $f$ by the quantity $ g - x_0 $ may produce unexpected (and undesired) behaviors.
Finally, if the scaling factor $ g - x_0 $ changes sign from the learned trajectory to the new one, the trajectory will result mirrored.

\noindent
In Figure~\ref{fig:old_porblems} we present an example similar to that in Figure~2 of \cite{HPPS09} to show the aforementioned drawbacks.
In both tests, the vertical component of the difference $ \vect{g}- \vect{x}_0 $ is quite small: \( (\vect{g} - \vect{x}_0) |_{2} = \frac{1}{15} \approx 0.0667 \).
Figure~\ref{subfig:ovn_high} shows that even a slight change in goal position results in a complete different trajectory when using formulation \eqref{eqs:dmp_old_form}.
Figure~\ref{subfig:ovn_under}, instead, shows the `mirroring' problem.
In this case, the vertical component of $ \vect{g} - \vect{x}_0 $ changes sign from the learned to the new execution. Thus, the trajectory results mirrored w.r.t. the horizontal axis $ x_2 = 0 $.

To overcome these disadvantages, the following formulation was proposed in \cite{PHPS08, PHAS09, HPPS09}:

\begin{subnumcases}{\label{eqs:new_dmps_vector}}
    \!\!\!  \tau \dot{\vect{v}} = \mtrx{K} (\vect{g} - \vect{x}) - \mtrx{D} \vect{v} - \mtrx{K} (\vect{g}- \vect{x}_0)s + \mtrx{K} \vect{f}(s) \label{eq:new_dmp_vect_acc}\\
    \!\!\! \tau \dot{\vect{x}} = \vect{v}
\end{subnumcases}
where the evolution of $s$ is still described by the canonical system \eqref{eq:canonical_system}, and the forcing term is still written in terms of basis functions as in \eqref{eq:forcing_term}.

By removing the scaling term $ \vect{g}-\vect{x}_0 $ from the forcing term $\vect{f}$, this new formulation solves the aforementioned problems of \eqref{eqs:dmp_old_form}, as it can be seen in Figure~\ref{fig:old_porblems}.
However, while solving the drawbacks characterizing formulation \eqref{eqs:old_dmps_vector}, DMPs' formulation \eqref{eqs:new_dmps_vector} still presents an important drawback: the generalization to different spatial scales is not always feasible since the forcing term does not have any dependence on the relative position between starting and ending points.
We will discuss the details of this aspect in Section~\ref{sec:linear_invariance} when presenting our proposed modifications to solve this drawback.

% ---------------------------------------------------------------------------- %
% CONTRIBUTION
% ---------------------------------------------------------------------------- %

% ---------------------------------------------------------------------------- %
% BASIS FUNCTIONS
% ---------------------------------------------------------------------------- %

\section{New Set of Basis Functions}\label{sec:basis}

A change of the set of basis functions is motivated by the fact that a desirable property of basis functions is to be compactly supported \cite{WWCT16} since compactly supported basis functions provide an easier update of the learned trajectory.
Indeed, if only a portion of the trajectory has to be modified, only a subset of the weights needs to be re-computed (we will discuss the details of this aspect in Section~\ref{subsec:trj_update}).
Moreover, while with GBFs the forcing term never vanishes, compactly supported basis functions allow the forcing term to be zero outside the support, thus providing a faster convergence of the system to the goal position.

In \cite{WWCT16} the authors proposed to use \emph{Truncated Gaussian Basis Functions} (TGBFs):
\begin{equation}
    \label{eq:truncated_gaussian}
    \tilde{\psi}_i(s) =
    \begin{cases}
        \exp \br{ - \frac{h_i}{2} \, (s - c_i) ^ 2 } & \text{if } s - c_i \le \theta_i \\
        0 & \text{otherwise}
    \end{cases},
\end{equation}
where $c_i$ and $h_i$ are defined as in \eqref{eq:basis_centers_def} and \eqref{eq:basis_width_gaussian_def} respectively, and \( \theta_i = K / \sqrt{h_i} \).
Unfortunately, this approach has two main drawbacks.
Firstly, a discontinuity is introduced at the truncation points $\theta_i$, which may produce unexpected behaviors.
Secondly, in order to work properly, this approach requires the introduction of \emph{biases terms} in \eqref{eq:forcing_term}, which now reads
\begin{equation}
    \label{eq:forcing_term_biases}
        f (s) = \frac{ \sum_{i=0}^N (\omega_i s + \beta_i) \, \tilde\psi_i(s) }{ \sum_{i=0}^N \tilde\psi_i(s) },
\end{equation}
thus doubling the number of parameters that need to be learned (a weight $\omega_i$ and a bias $\beta_i$ for each basis function), and increasing the overall computational cost.

\noindent
Finally, we remark that these basis functions are not compactly supported since their support is an interval $ (-\infty, L] $ with $L \in \RR$.
Moreover, since the truncation point is at the ``right'' in $s$ (i.e. $ \tilde{\psi_i}(s) \equiv 0 $ when $s$ is above a certain quantity), and $s$ is a decreasing function in $t$, we remark that these basis functions never vanish as $ t \to \infty $.

In this part, we propose a new set of basis functions that improves the TGBFs \eqref{eq:truncated_gaussian}.
Indeed, our proposed set contains compactly supported basis functions, which then vanish in both directions.
Moreover, since no truncation (and, thus, discontinuities) are introduced, there is no need to double the number of parameters, and our formulation will require the learning of only the weights $ \omega_i $ in \eqref{eq:forcing_term}.

We will also show that our proposed set of basis functions results in a minimization problem that is both numerically more stable and faster to solve.

\subsection{Mollifier-like and Wendland Basis Functions}\label{sec:comp_supp_basis}

In this work, we propose a new set of basis functions that are both smooth and compactly supported.

\noindent
Consider the \emph{mollifier} $ \varphi :\RR \to \RR $ defined as
\begin{equation}
    \label{eq:mollifier_def}
    \varphi (x) =
    \begin{cases}
        \frac{1}{I_n} \exp \br{ - \frac{ 1 }{ 1 - |x|^2 } } & \text{if }|x| < 1 \\
        0 & \text{otherwise}
    \end{cases},
\end{equation}
where $ I_n $ is set so that $\int_\RR \varphi (x) \diff x = 1$. Function $\varphi$ is smooth and compactly supported (its support is the interval $[-1, 1]$).

\noindent
It is then natural to define the set of \emph{mollifier-like basis functions} $ \set{ \varphi_i (s) }_{i = 0, 1, \ldots, N} $ given by
\begin{equation}
    \label{eq:mollifier_basis_def}
    \varphi_i(s) =
    \begin{cases}
        \exp \br { - \frac{ 1 }{ 1 - r ^ 2 } } & \text{if }r < 1 \\
        0 & \text{otherwise}
    \end{cases},
\end{equation}
where $ r $ is a function of $s$ defined as
\begin{equation}
    \label{eq:r_def}
    r = | a_i (s - c_i) |,
\end{equation}
where, given the centers $c_i$ as in \eqref{eq:basis_centers_def}, we define the widths $a_i$ as
\begin{equation}
    \label{eq:mollififer_basis_widths_def}
    \begin{array}{l}
        a_i = \dfrac{ \tilde{h} }{ \vert c_{i} - c_{i-1} \vert },\quad i = 1,2,\ldots, N,\\
        a_0 = a_{1}.
    \end{array}
\end{equation}
As in \eqref{eq:basis_width_gaussian_def}, scalar $ \tilde{h} > 0 $ determines the overlapping between basis functions.
In all our tests, we fix $ \tilde{h} $ to value one: $ \tilde{h} = 1 $.
We remark that the usual normalization term $I_n$ in \eqref{eq:mollifier_def} is omitted in \eqref{eq:mollifier_basis_def} since it does not enter our approach.

\begin{figure}[t]
    \centering
    \includegraphics[width = 0.9 \columnwidth]{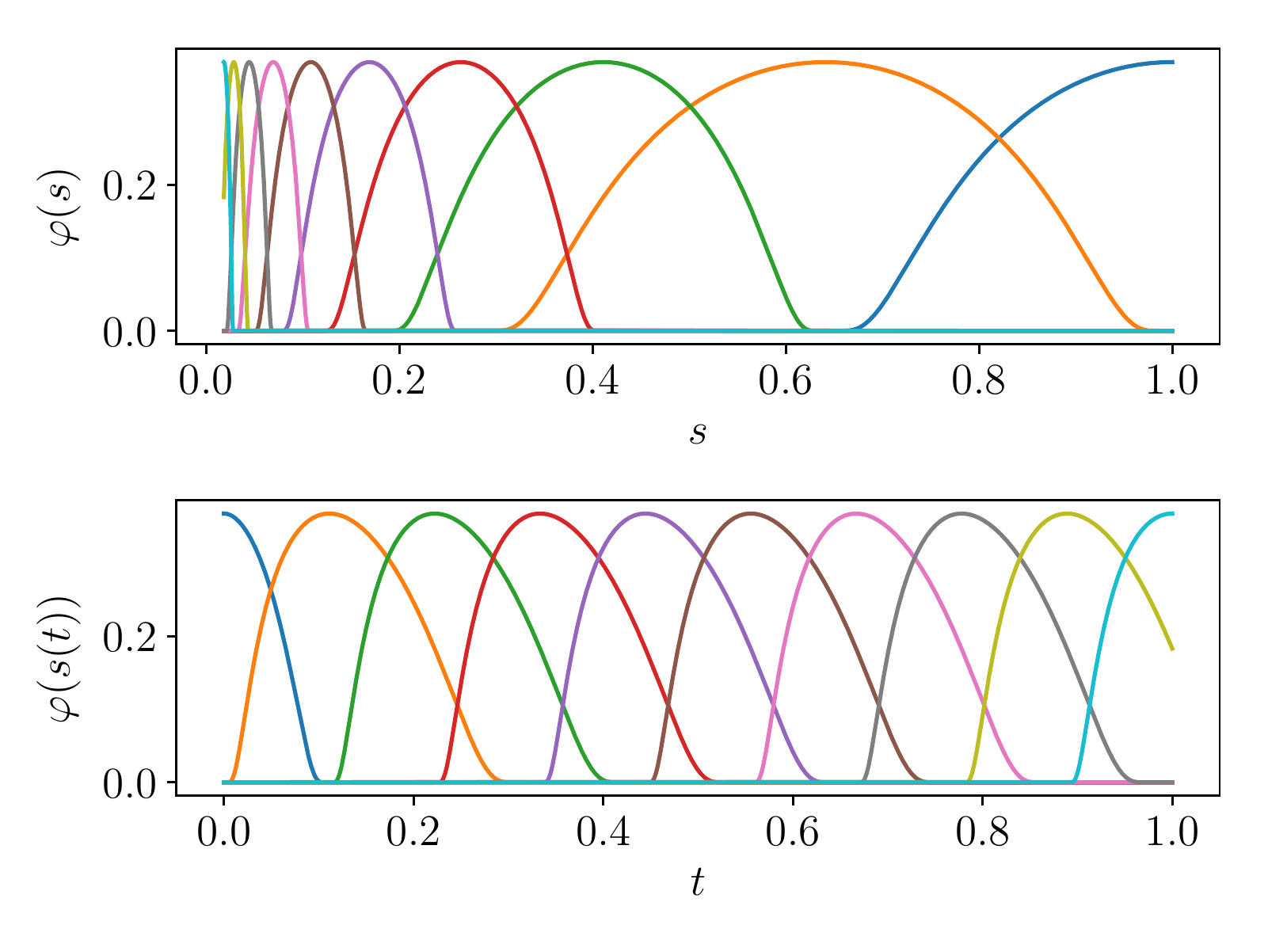}
    \caption{Plot of mollifier-like basis functions as defined in \eqref{eq:mollifier_basis_def}.
    In this example $N=9$, $\alpha = 4$ and $T = 1$.
    The first plot shows the basis functions as a function of $s$, while the second plot shows them as functions of time $t$.
    }
    \label{fig:mollifier_basis_plot}
\end{figure}

\noindent
In Figure~\ref{fig:mollifier_basis_plot} we plot an example of mollifier-like basis functions both as function of $s$ and of $t$.
We remark that the basis functions are equispaced in $t$.
Moreover, since $s$ is a strictly decreasing function of $t$, their order changes from $s$ to $t$ (the first basis function in $s$ is the last in $t$ and vice versa), as it was for the GBFs used so far in the literature.

\begin{table*}[h]
    \small\sf\centering
    \caption{Summary of the properties (regularity and support compactness) of the different sets of basis functions presented in Section~\ref{sec:basis}.
    Bold font is used to mark the more desirable properties (smoothness and compact support).
    \label{tab:basis_prop}}
    \begin{tabular}{cccc}
        \hline
        \multicolumn{2}{c}{Function}                      & Regularity                  & Compactly Supported \\
        \hline
        Gaussian           & $\psi_i (s) (\cdot) $         & $\bm{\mathcal{C}^{\omega}(\RR)}$ & No \\ \hline
        Truncated Gaussian & $\tilde{\psi}_i (s) (\cdot) $ & Discontinuous               & No \\ \hline
        Mollifier-like     & $\varphi_i (s) (\cdot) $         & $\bm{\mathcal{C}^{\infty}(\RR)}$ & \textbf{Yes} \\ \hline
        \multirow{7}{*}{Wendland}  
        & $ \phi ^ {(2)} ( \cdot ) $    & $ \mathcal{C}^0(\RR) $      & \textbf{Yes}  \\
        & $ \phi ^ {(3)} ( \cdot ) $    & $ \mathcal{C}^0(\RR) $      & \textbf{Yes}  \\
        & $ \phi ^ {(4)} ( \cdot ) $    & $ \mathcal{C}^2(\RR) $      & \textbf{Yes}  \\
        & $ \phi ^ {(5)} ( \cdot ) $    & $ \mathcal{C}^2(\RR) $      & \textbf{Yes}  \\
        & $ \phi ^ {(6)} ( \cdot ) $    & $ \mathcal{C}^4(\RR) $      & \textbf{Yes}  \\
        & $ \phi ^ {(7)} ( \cdot ) $    & $ \mathcal{C}^4(\RR) $      & \textbf{Yes}  \\
        & $ \phi ^ {(8)} ( \cdot ) $    & $ \mathcal{C}^6(\RR) $      & \textbf{Yes}  \\
        \hline
    \end{tabular}
\end{table*}

Other well known regular (see Table~\ref{tab:basis_prop} for details on the regularity) and compactly supported basis functions are \emph{Wendland functions} \cite{Wen95, Sch07}.
In Section~\ref{sec:basis_results} we test the following Wendland basis function (where the operator $(\cdot)_+$ denotes the \emph{positive part function}: \( (x)_+ = \max\{ 0, x \}\)):
\begin{subequations}
    \label{eqs:wendland_set}
    \begin{align}
        \phi_i^{(2)}(r) & = (1-r)_+ \! {}^2 ,                               \label{eq:wen_30} \\
        \phi_i^{(3)}(r) & = (1-r)_+ \! {}^3 ,                               \label{eq:wen_50} \\
        \phi_i^{(4)}(r) & = (1-r)_+ \! {}^4 \, (4r + 1),                    \label{eq:wen_32} \\
        \phi_i^{(5)}(r) & = (1-r)_+ \! {}^5 \, (5r + 1),                    \label{eq:wen_52} \\
        \phi_i^{(6)}(r) & = (1-r)_+ \! {}^6 \, (35r^2 + 18r + 3),           \label{eq:wen_34} \\
        \phi_i^{(7)}(r) & = (1-r)_+ \! {}^7 \, (16 r^2 + 7 r + 1),          \label{eq:wen_54} \\
        \phi_i^{(8)}(r) & = (1-r)_+ \! {}^8 \, (32r^3 + 25r^2 + 8r + 1).    \label{eq:wen_36}
    \end{align}
\end{subequations}
where $c_i$ are the centers as in \eqref{eq:basis_centers_def} and $a_i$ are the widths as in \eqref{eq:mollififer_basis_widths_def}, and $r$ is defined as in \eqref{eq:r_def}.

In Table~\ref{tab:basis_prop} we summarize the properties of the basis functions we presented in this section.
As can be seen, mollifier-like basis functions are the only ones being both smooth and compactly supported.

%%% ---------------- %%%
%%%  WEIGHTS UPDATE  %%%
%%% ---------------- %%%
\subsection{Trajectory Update}
\label{subsec:trj_update}

In \cite{WWCT16} was pointed out that if only a portion of the trajectory has to be re-learned, only a subset of weights has to be re-computed.
To be more precise, assume that a desired trajectory $ \gamma (t) $ has to be updated in the time interval $ [t_0, t_1] \subset [0, T] $.
Then, not all the weights $ \omega_i $ have to be re-computed, but only those whose basis functions $ \varphi_i $ satisfy (where the support has to be intended on the values of $t$ and not $s$)
\begin{equation*}
    \text{\normalfont{supp}} (\varphi_i) \cap [t_0, t_1] \neq \emptyset.
\end{equation*}
From this, it is easy to notice that for classical GBFs this means that all weights have to be re-computed, since their support is the whole real line $ \RR $: \( \text{\normalfont{supp}} (\psi_i) = \RR \).
For compactly supported basis functions, this intersection will be usually smaller than the whole set of indexes $ \{ 0, 1, \ldots, N \} $, and it will depend only on the length $ t_1 - t_0 $ of the interval.

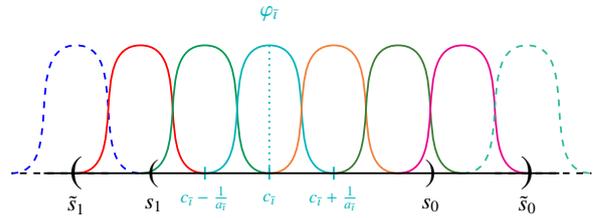
\begin{figure}[t]
    \centering
    \resizebox{1.0\linewidth}{!}{
        \begin{tikzpicture}
            \draw[thick, blue, dashed] (-1.5, 0) to[out = 0, in = 180]++ (1, 2) to[out = 0, in = 180]++ (1, -2);
            \draw[thick, red] (-0.5, 0) to[out = 0, in = 180]++ (1, 2) to[out = 0, in = 180]++ (1, -2);
            \draw[thick, ForestGreen] (0.5, 0) to[out = 0, in = 180]++ (1, 2) to[out = 0, in = 180]++ (1, -2);
            \draw[thick, Aquamarine] (1.5, 0) to[out = 0, in = 180]++ (1, 2) to[out = 0, in = 180]++ (1, -2);
            \draw[thick, Orange] (2.5, 0) to[out = 0, in = 180]++ (1, 2) to[out = 0, in = 180]++ (1, -2);
            \draw[thick, OliveGreen] (3.5, 0) to[out = 0, in = 180]++ (1, 2) to[out = 0, in = 180]++ (1, -2);
            \draw[thick, Magenta] (4.5, 0) to[out = 0, in = 180]++ (1, 2) to[out = 0, in = 180]++ (1, -2);
            \draw[thick, SeaGreen, dashed] (5.5, 0) to[out = 0, in = 180]++ (1, 2) to[out = 0, in = 180]++ (1, -2);

            \draw[thick, Aquamarine, opacity=1, dotted] (2.5, 2) -- (2.5, 0);
            \draw[thick, Aquamarine] (1.5, 0.1) -- (1.5, -0.1);
            \draw[thick, Aquamarine] (2.5, 0.1) -- (2.5, -0.1);
            \draw[thick, Aquamarine] (3.5, 0.1) -- (3.5, -0.1);
            \node() at (2.5, 2.5) {\color{Aquamarine}$\varphi_{\bar{\imath}}$};
            \node() at (2.5, -0.4) {\footnotesize\color{Aquamarine}$c_{\bar{\imath}}$};
            \node() at (1.5, -0.4) {\footnotesize\color{Aquamarine}$c_{\bar{\imath}} - \frac{1}{a_{\bar{\imath}}}$};
            \node() at (3.5, -0.4) {\footnotesize\color{Aquamarine}$c_{\bar{\imath}} + \frac{1}{a_{\bar{\imath}}}$};

            \node() at (-0.5, 0) {\LARGE$($};
            \node() at (6.5, 0) {\LARGE$)$};
            \node() at (-0.5, -0.5) {$\tilde{s}_1$};
            \node() at (6.5, -0.5) {$\tilde{s}_0$};
            \draw[thick] (-1, 0) -- (7, 0);
            \draw[thick, dashed] (0, 0) --++ (-1.5, 0);
            \draw[thick, dashed] (6, 0) --++ (+1.5, 0);
            \node() at (0.7, 0) {\Large$($};
            \node() at (5, 0) {\Large$)$};
            \node() at (0.7, -0.5) {$s_1$};
            \node() at (5, -0.5) {$s_0$};
        \end{tikzpicture}
    }
    \caption{
        Depiction of compactly supported basis functions whose support intersects an interval $(s_1, s_0)$.
        The basis functions drawn using solid lines are those whose weights have to be updated.
        As one can observe, they satisfy condition \eqref{eq:w_up_condition}.
        Interval $ (\tilde{s}_1, \tilde{s}_0) $ shows the domain in which the integrals in \eqref{eqs:integrals_min_problem} have to be computed.
    }
    \label{fig:weights_update_depiction}
\end{figure}

\noindent
The support, in $s$, of each basis function $ \varphi_i $ is the interval
\( \left( c_i - \nicefrac{1}{a_i} , c_i + \nicefrac{1}{a_i} \right) . \)
Therefore, the set of weights $ \{\omega_i\}_{i \in I} $ that has to be updated are those for which
\begin{equation}
    c_i - \frac{1}{a_i} \le s_0 \quad \text{and} \quad c_i + \frac{1}{a_i} \ge s_1,
    \label{eq:w_up_condition}
\end{equation}
where $ s_0 = \exp ( -\alpha \, t_0 ) $ and $ s_1 = \exp ( -\alpha \, t_1 ) $.

\noindent
Once the set of indexes $ I $ has been identified using \eqref{eq:w_up_condition}, one must solve a linear problem as in \eqref{eq:lin_p_1}, where $ \mtrx{A} $ is a $ |I| \times |I| $ matrix,
\( \mtrx{A} \in \RR^{|I| \times |I|} , \)
and $\vect{b}$ is a vector with $|I|$ components,
\(\vect{b} \in \RR^{|I|} .\)
The components of matrix $ \mtrx{A} $ and vector $ \vect{b} $ are given in \eqref{eqs:integrals_min_problem}.
We remark that the integrals in \eqref{eqs:integrals_min_problem} have to be evaluated on the interval
\begin{equation*}
    ( \tilde{s}_1 , \tilde{s}_0 ) = \bigcup_{i \in I} \text{supp}(\varphi_i),
\end{equation*}
where the support has to be intended in $s$.
By solving the linear problem, the weights $ \omega_i $, $ i \in I $ are updated.

\begin{figure*}[t]
    \centering
    \subfloat[Error as function of the number of basis functions. \label{subfig:err_n_bfs}]{\includegraphics[width=0.9\columnwidth]{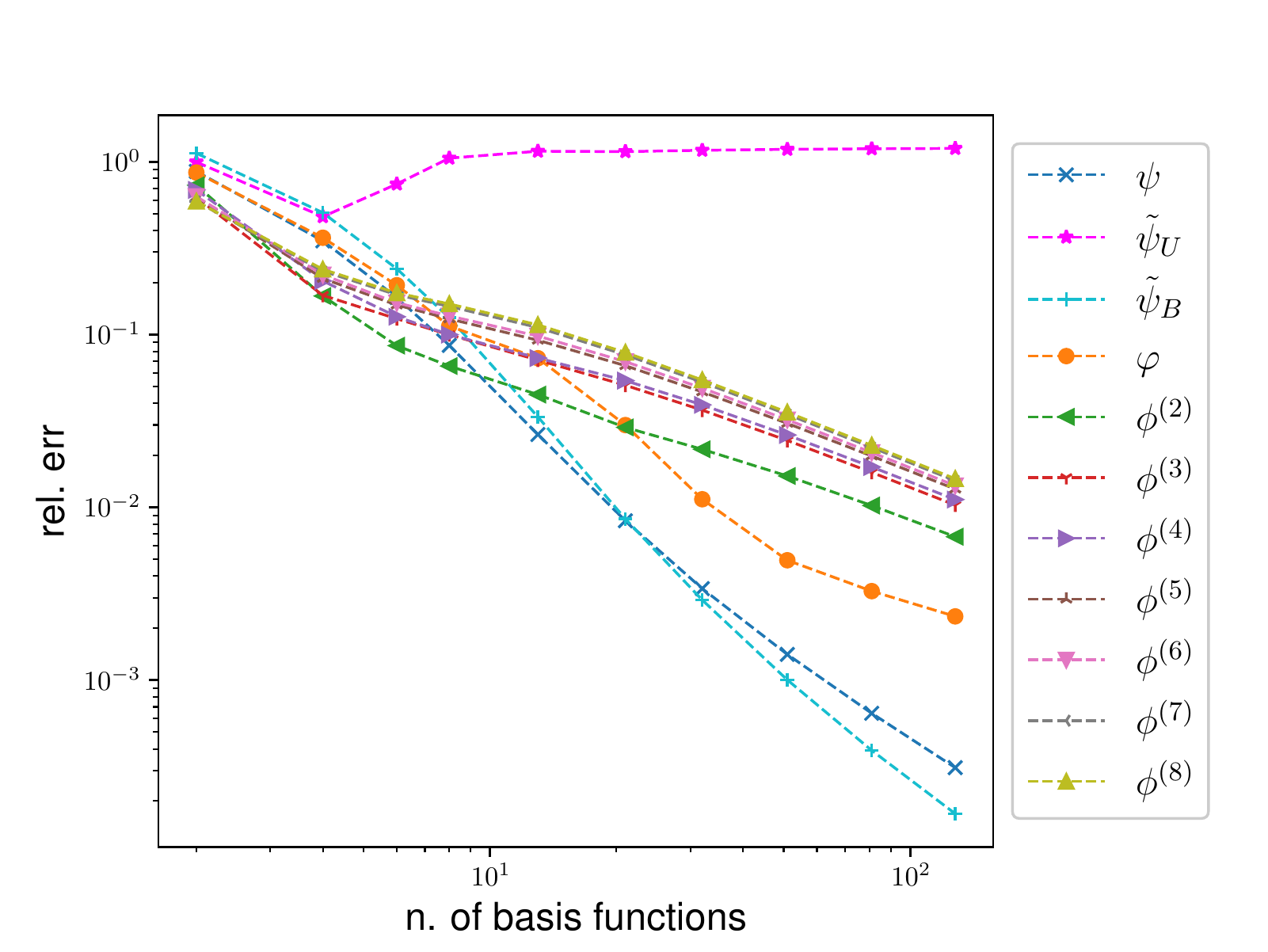}}
    \hspace{1.25cm}
    \subfloat[Error as function of the number of parameters. \label{subfig:err_n_param}]{\includegraphics[width=0.9\columnwidth]{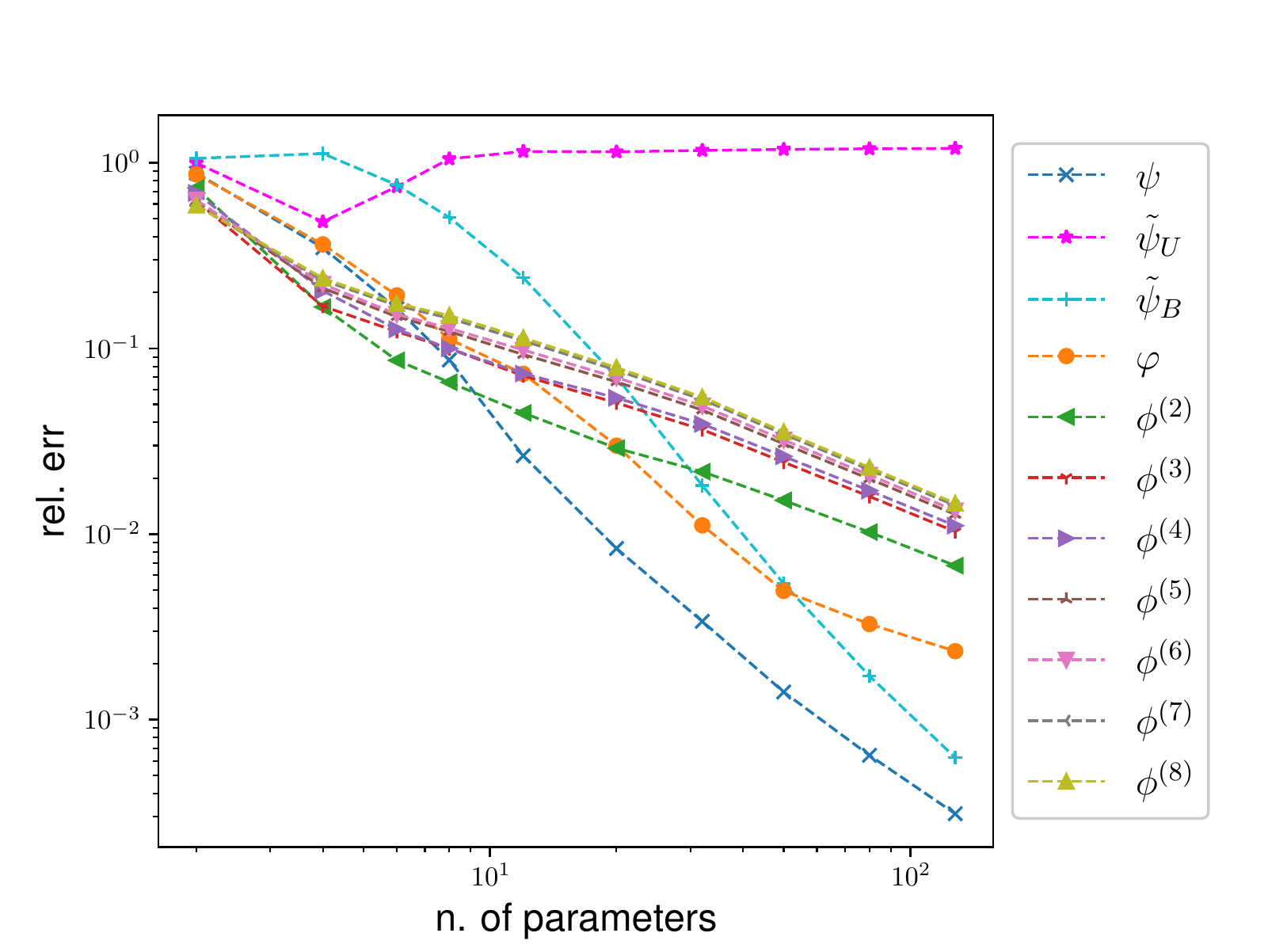}}
    \caption{Plot of the $L^2$ error done by approximating the hat function \eqref{eq:target_function_basis_test}.
    The first plot shows the error w.r.t. the number of basis functions, while the second plot shows the error w.r.t. the number of parameters that has to be learned.
    TGBFs are tested both using the (classical) unbiased formulation \eqref{eq:forcing_term} and the biased one \eqref{eq:forcing_term_biases}.
    These two different approaches are denoted respectively by $\tilde \psi_U$ and $\tilde{\psi_B}$ in the legend.}
    \label{fig:test_basis}
\end{figure*}

\noindent
Figure~\ref{fig:weights_update_depiction} gives an intuition on how to identify the set $ \{ \varphi_i \}_{i \in I} $ of basis functions satisfying \eqref{eq:w_up_condition} and the interval $ ( \tilde{s}_1 , \tilde{s}_0 ) $.

\begin{rmk}
    In the case of TGBFs, the set of weights to update will depend not only on the length of the interval but also on its `position'.
    For instance, if the last part of the trajectory has to be updated, i.e. $ [t_0, t_1] = [t_0, T] $, all the weights must be updated since $ T \in \text{\normalfont{supp}} (\tilde{\psi}_i) $ for any $ i = 0, 1, \ldots, N $.

    \noindent
    Thus, in general, when a portion of the trajectory has to be modified, the number of weights to re-compute will be less when using compactly supported basis functions than when using TGBFs.
\end{rmk}

% \begin{algorithm}[t]
%     \caption{Weights update}
%     \label{alg:weights_update}
%     \begin{algorithmic}[1]
%         \Require{Family of basis functions $ \varphi $, centers set $ \{ c_i \}_{i = 0,1,\ldots,N} $, widths set $ \{ a_i \}_{i=0,1,\ldots,N} $, time interval $ (t_0, t_1) $, new desired trajectory $ \tilde{x}(t) $, DMPs' parameters.}
%         \Ensure{List $I$ of indexes, updated weights $ \omega_i, i \in I $.}
%         \State{Compute the interval in $s$:
%             \[ s_0 = \exp(-\alpha \, t_0), \quad s_1 = \exp(-\alpha \, t_1) \]}
%         \State{Compute the set $I$ of indeces $i$ as those satisfying $ c_i - \nicefrac{1}{a_i} \le s_0 $ and $ c_i + \nicefrac{1}{a_1} \ge s_1 $.}
%         \State{Compute the desired forcing term $ \tilde{f}(s) $.}
%         \State{Compute the matrix $\mtrx{A}$ and vector $\vect{b}$ of the minimization problem as in \eqref{eqs:integrals_min_problem}:
%             \begin{align*}
%                 a_{h k} & = \int_{s_1}^{s_0} \frac{\psi_h(s) \psi_k(s)}{\br{ \sum_{i=0}^N \psi_i(s) }^2}\, s^2 \, \diff s \\
%                 b_h & = \int_{s_1}^{s_0} \frac{\psi_h(s) }{ \sum_{i=0}^N \psi_i(s) }\, f(s)\, s \, \diff s ,
%             \end{align*}
%             for $ h, k \in I $.
%         }
%         \State{Solve the linear problem \( \mtrx{A} \tilde{\bm{\omega}} = \vect{b} \).
%         The components in $ \tilde{\bm{\omega}} $ are the updated weights $ \omega_i $, $ i \in I $.}
%     \end{algorithmic}
% \end{algorithm}

%%% -------- %%%
%%%  RESULT  %%%
%%% -------- %%%

\subsection{Results}
\label{sec:basis_results}

We now investigate the goodness of the approximations obtained using the various examples of basis functions we introduced.
In particular, we test both the classical \eqref{eq:gaussian_basis_def} and the truncated \eqref{eq:truncated_gaussian} Gaussian basis functions, the mollifier-like basis functions \eqref{eq:mollifier_basis_def}, and the Wendland functions \eqref{eqs:wendland_set}.
We test three numerical aspects: in Section~\ref{sec:num_acc} the approximation error on particular target functions, in Section~\ref{sec:cond_num} the condition number of matrix $ \mtrx{A} $ in \eqref{eq:lin_p_1}, and in Section~\ref{subsec:time_eff} we test the computtional time needed to solve the minimization problem \eqref{eq:lin_p_1}.
Additionally, in Section~\ref{subsec:trj_up_result} we present a test on the trajectory update property presented in Section~\ref{subsec:trj_update}.

\subsubsection{Numerical Accuracy.}
\label{sec:num_acc}
We test the accuracy of the approximation of the following function:
\begin{equation}
\label{eq:target_function_basis_test}
    \eta(t) = t ^ 2 \, \cos (\pi \, t), \quad t \in [0, 1].
\end{equation}

\begin{figure*}[t]
    \centering
    \subfloat[\texttt{pushing\_needle}.\label{subfig:error_real_push}]{\includegraphics[width=0.9\columnwidth]{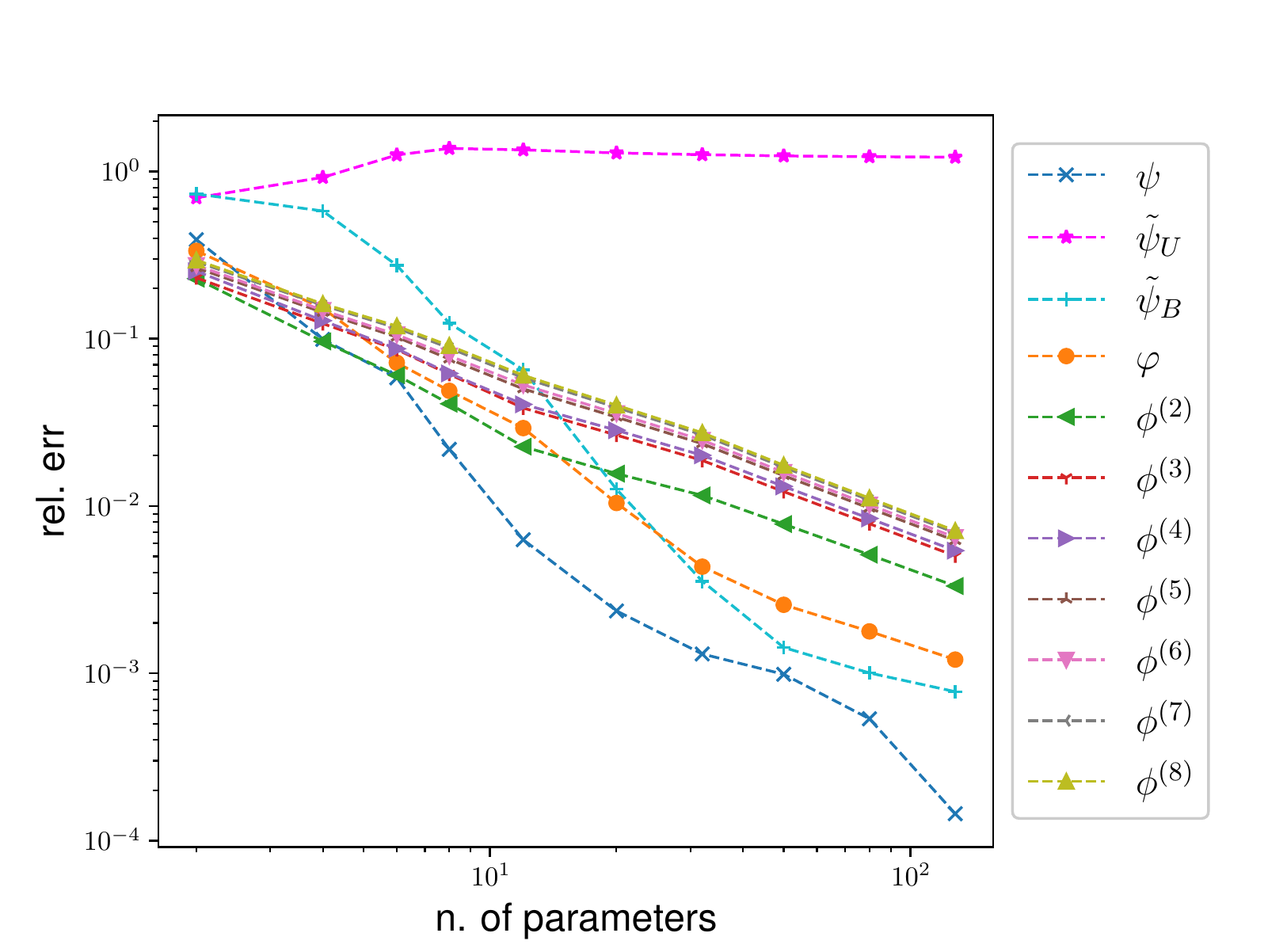}}
    \hspace{1.25cm}
    \subfloat[\texttt{positioning\_needle}.\label{subfig:error_real_pos}]{\includegraphics[width=0.9\columnwidth]{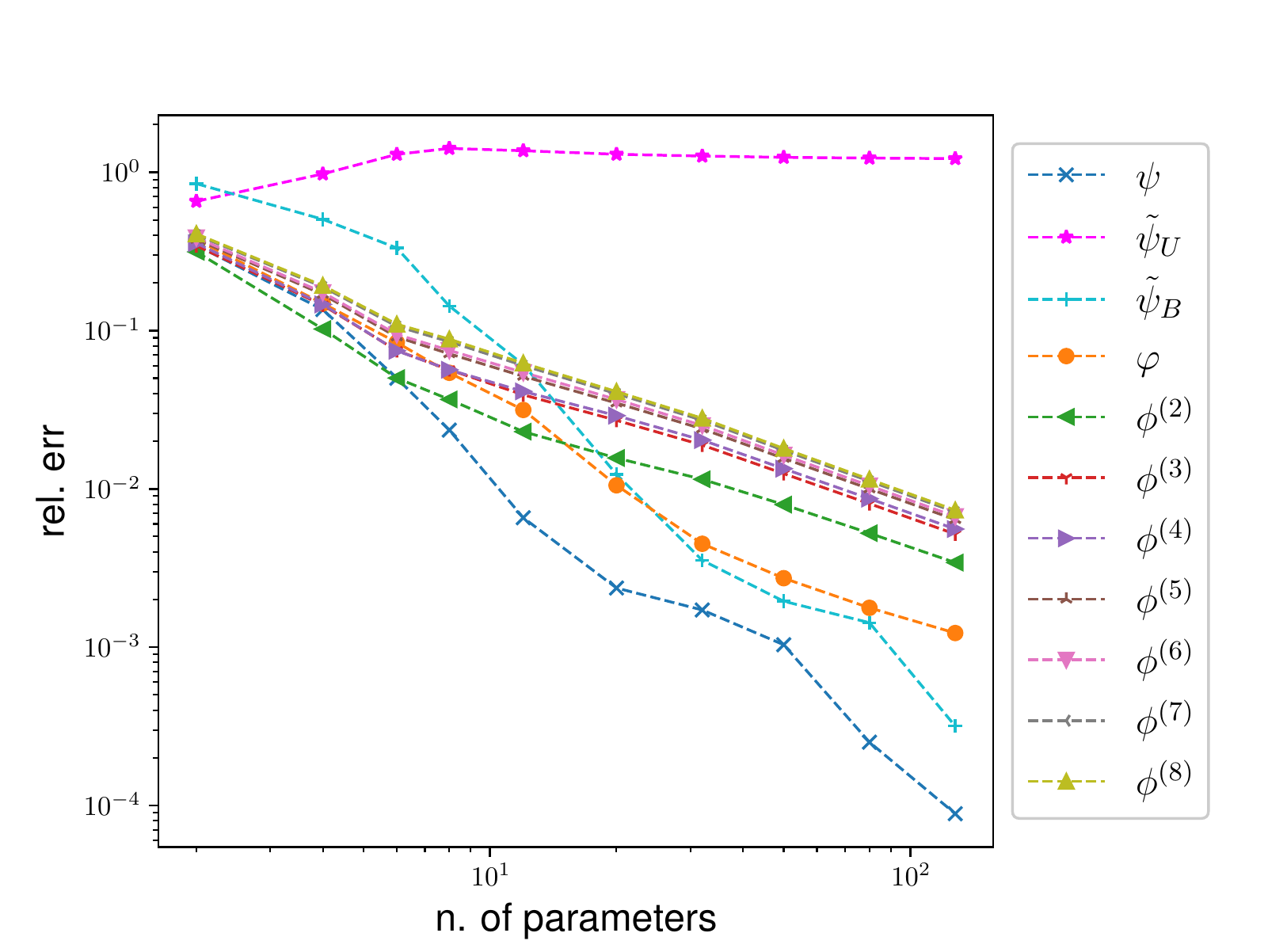}}
    \caption{Plot of the $L^2$ error done approximating two real gestures.
    The error is given as a function of the number of parameters that need to be learned.
    TGBFs are tested both using the (classical) unbiased formulation \eqref{eq:forcing_term} and the biased one \eqref{eq:forcing_term_biases}.
    These two different approaches are denoted respectively by $\tilde \psi_U$ and $\tilde{\psi_B}$ in the legend.
    }
    \label{fig:err_basis_real}
\end{figure*}

Figure~\ref{fig:test_basis} shows the approximation error, measured w.r.t. the $L^2$-norm, both w.r.t. the number of basis functions (Figure~\ref{subfig:err_n_bfs}), and w.r.t. the number of parameters that need to be learned (Figure~\ref{subfig:err_n_param}).
We recall that for TGBFs the number of parameters is double the number of basis functions since every basis function requires one weight and one bias term.
On the other hand, for all other classes of basis functions, the number of parameters coincide with the number of basis functions, since for each basis function only one parameter (the weight) has to be computed.

\noindent
The plot shows that TGBFs \eqref{eq:truncated_gaussian} work properly only using the biased formulation (denoted by $ \tilde{\psi}_B $ in the legend) for the forcing term \eqref{eq:forcing_term_biases}, which means that given $N$ basis functions we are solving a $2N$-length linear system when computing the weights and the biases.
On the other hand, when using the unbiased formulation \eqref{eq:forcing_term} (denoted by $ \tilde{\psi}_U $ in the legend) the approximant does not converge to the desired forcing term.

\noindent
We observe that when comparing the error w.r.t. the number of basis functions, the error obtained by using TGBFs is comparable to the error done using classical GBFs.
On the other hand, when comparing the error w.r.t. the number of parameters that have to be computed, TGBFs approximate worse than mollifier basis functions when the number of parameters is below $ 60 $.
Usually, in the applications, no more than fifty basis functions are used.
For this particular value and target function, mollifier-like basis functions and truncated Gaussian have almost identical approximation errors.
However, we remark that these results depend on the particular choice of the target function $ \eta $ in \eqref{eq:target_function_basis_test}, and that different target functions may give different results.
Tests performed with different target functions show similar results: classical Gaussian functions remain the best approximator, while truncated Gaussian and mollifier-like give similar approximation accuracy and usually work better than Wendland functions.

Additionally, we performed this test on trajectories obtained from real task executions.
In particular, we relied on the JIGSAW dataset \cite{GVRAVLTZKH14}.
This dataset contains data on three elementary surgical tasks (suturing, knot-tying, and needle-passing).
The demonstrations were collected from eight different subjects of different experience in robotic surgery.
In Figure~\ref{fig:err_basis_real} we show the convergence error for two gestures extracted from the dataset.
In particular, Figure~\ref{subfig:error_real_push} shows the results for the \texttt{pushing\_needle} gesture, while Figure~\ref{subfig:error_real_pos} shows the results for the \texttt{positioning\_needle} gesture.
Both gestures were extracted from the same user (`D') performing the same task (suturing).
As it can be seen, the results are similar to those obtained by approximating function $ \eta $ in \eqref{eq:target_function_basis_test}: GBFs \eqref{eq:gaussian_basis_def} are the best approximators, TGBFs \eqref{eq:truncated_gaussian} and mollifier-like basis functions \eqref{eq:mollifier_basis_def} have similar convergence result, and Wendland basis functions \eqref{eqs:wendland_set} are the less accurate basis functions.

In all tests, we tested classical Gaussians, Wendland, and mollifier-like basis functions also using the biased formulation \eqref{eq:forcing_term_biases}, without noticing any difference in the goodness of the approximation.

\subsubsection{Condition Number.}
\label{sec:cond_num}

\begin{figure}[t]
    \centering
    \includegraphics[width=0.9\columnwidth]{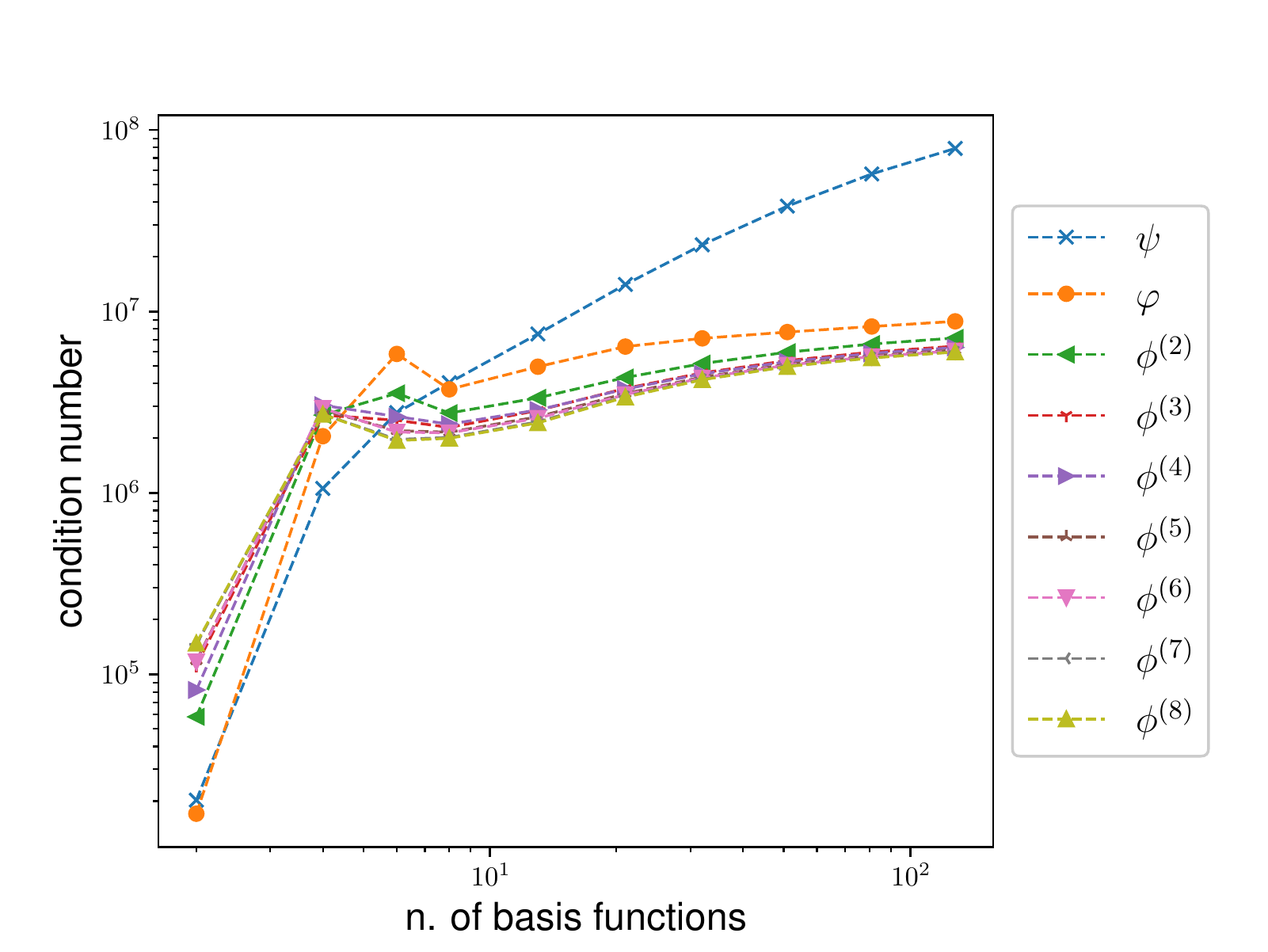}
    \caption{Condition number of the matrix $\mtrx{A}$ with elements $a_{hk} $ defined in \eqref{eq:regression_problem_linear_part} w.r.t. the number of basis functions.}
    \label{fig:condition_number_test}
\end{figure}

We now discuss the numerical accuracy of the minimization problem \eqref{eq:lin_p_1} used to compute the weights $\omega_i$ in \eqref{eq:forcing_term}.
To do so, we investigate the condition number of matrix $\mtrx{A}$, \( \cond (\mtrx{A}) \defas \|\mtrx{A}\| \, \| \mtrx{A}^{-1}\| \), in \eqref{eq:lin_p_1}.
The importance of this test is given from the fact that when one solves numerically a linear system, the approximation error is directly proportional to the condition number.
Indeed, when numerically solving a linear system $ \mtrx{A} \vectgreek{\omega} = \vect{b} $, whose exact solution is $ \tilde{\vectgreek{\omega}} $, and numerical solution is $ \hat{\vectgreek{\omega}} $, the following inequality holds:
\begin{equation}
    \label{eq:cond_num_result}
    \frac{ \| \tilde{\vectgreek{\omega}} - \hat{\vectgreek{\omega}} \| }{ \| \hat{\vectgreek{\omega}} \|} \le \cond (\mtrx{A}) \frac{ \| \vectgreek{\varsigma} \| }{ \| \vect{b} \| } ,
\end{equation}
where $ \vectgreek{\varsigma} \defas \vect{b} - \mtrx{A} \tilde{\vectgreek{\omega}}$ is the \emph{residual}.
Inequality \eqref{eq:cond_num_result} says that the relative error done when numerically solving a linear system is amplified by the condition number of the matrix $ \mtrx{A} $.

\noindent
In this test, we do not consider TGBFs since, in this case, the components $\mtrx{A}$ and $ \vect{b}$ of the linear problem \eqref{eq:lin_p_1} have different formulations, having to learn both the weights and the biases.

Figure~\ref{fig:condition_number_test} shows that the condition number \( \cond (\mtrx{A}) \) of matrix $\mtrx{A}$ increases faster for GBFs, while the compactly supported basis functions show a slower increase.
Moreover, even with only twenty basis functions, the condition number obtained with GBFs is significantly bigger than any compactly supported basis function.
This means that solving the minimization problem \eqref{eq:lin_p_1} using GBFs results in a more severe numerical cancellation error than mollifier-like or Wendland functions.

\begin{figure}[t]
    \centering
    \resizebox*{0.7\columnwidth}{!}{
        \begin{tikzpicture}
            \clip (-5.5, 5.5) rectangle (5, -5);
            \node () at (0,0) {\includegraphics{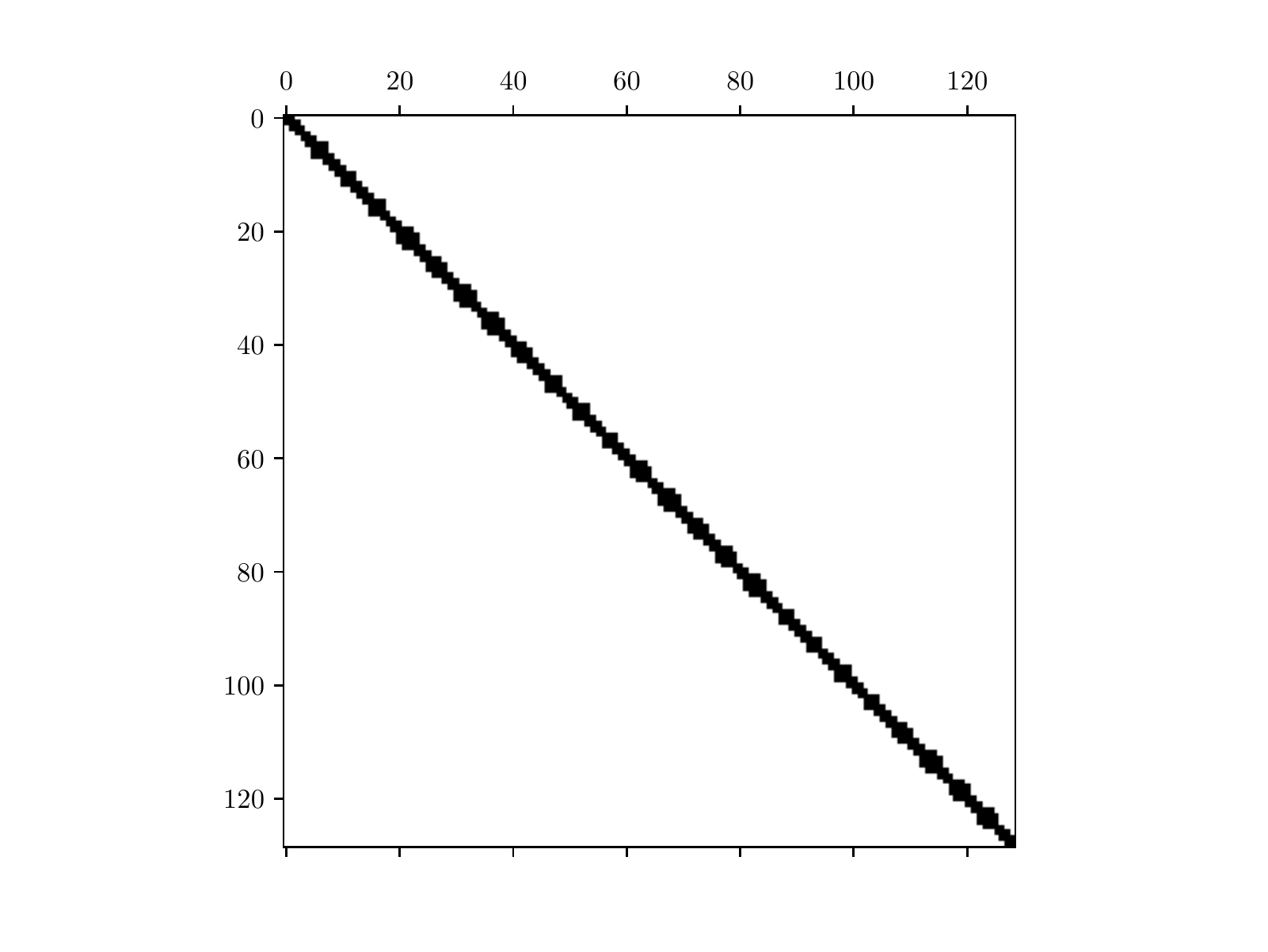}};
        \end{tikzpicture}
    }
    \caption{Sparsity pattern of matrix $\mtrx{A}$ with elements $a_{hk} $ defined in \eqref{eq:A_component} in the case of $N = 2^{7} = 128$ using mollifier-like basis functions.}
    \label{fig:sparsity}
\end{figure}

\noindent
The lower condition number can be explained by the fact that mollifier-like and Wendland basis functions are compactly supported, and then the resulting matrix $ \mtrx{A} $ will have many off-diagonal components equal to zero, as it can be seen in the sparsity pattern shown in Figure~\ref{fig:sparsity}.
On the other hand, when one uses GBFs, all the components of $\mtrx{A}$ are non-zero, since $ \psi_i(s) > 0, \forall s\in \RR $.
Thus, $\mtrx{A}$ is a full matrix when using GBFs, while it is a sparse $n-$diagonal matrix when using compactly supported basis functions; and, in general, full matrices have a bigger condition number than sparse $n-$diagonal ones.

We remark that convergence analysis shown in Figures~\ref{fig:test_basis} and \ref{fig:err_basis_real} is done on particular choices of target functions.
The convergence error may differ depending on the forcing term that needs to be approximated.
However, we stress that the condition number shown in Figure~\ref{fig:condition_number_test} does not depend on the target function but only on the choice of basis functions.

\subsubsection{Computational Time}
\label{subsec:time_eff}

An additional advantage in using compactly supported basis functions lies in the improvements in computational time when solving the linear problem \eqref{eq:lin_p_1}.
Indeed, when the number of basis functions $N$ increases, the entries in matrix $\mtrx{A}$ increases quadratically: $ \mathcal{O}(N^2) $.
When GBFs are used, all entries in $\mtrx{A}$ are non-zero.
On the other hand, when compactly supported basis functions are adopted, the number of non-zero entries in $ \mtrx{A} $ increases linearly, $ \mathcal{O}(N \, \ell ) $, where $ \ell $ is the number of non-zero diagonals in $ \mtrx{A} $, and is determined by the overlapping between basis functions, which depends on parameter $ \tilde{h} $ in \eqref{eq:mollififer_basis_widths_def}.

\noindent
Consequently, as the number of basis functions $N$ increases, the number of operations that are needed to solve the linear problem \eqref{eq:lin_p_1} increases cubically, $ \mathcal{O}(N^3) $, when the matrix is full; and only quadratically, $\mathcal{O}(N^2\ell)$, when the matrix is sparse.
Thus, one should expect the minimization problem \eqref{eq:lin_p_1} to be faster to solve when compactly supported basis functions are used.

\begin{figure}[t]
    \resizebox*{0.9\columnwidth}{!}{
        \begin{tikzpicture}
            \clip (-8.1, 4.8) rectangle (7.45, -6.1);
            \node at (0,0) {\includegraphics{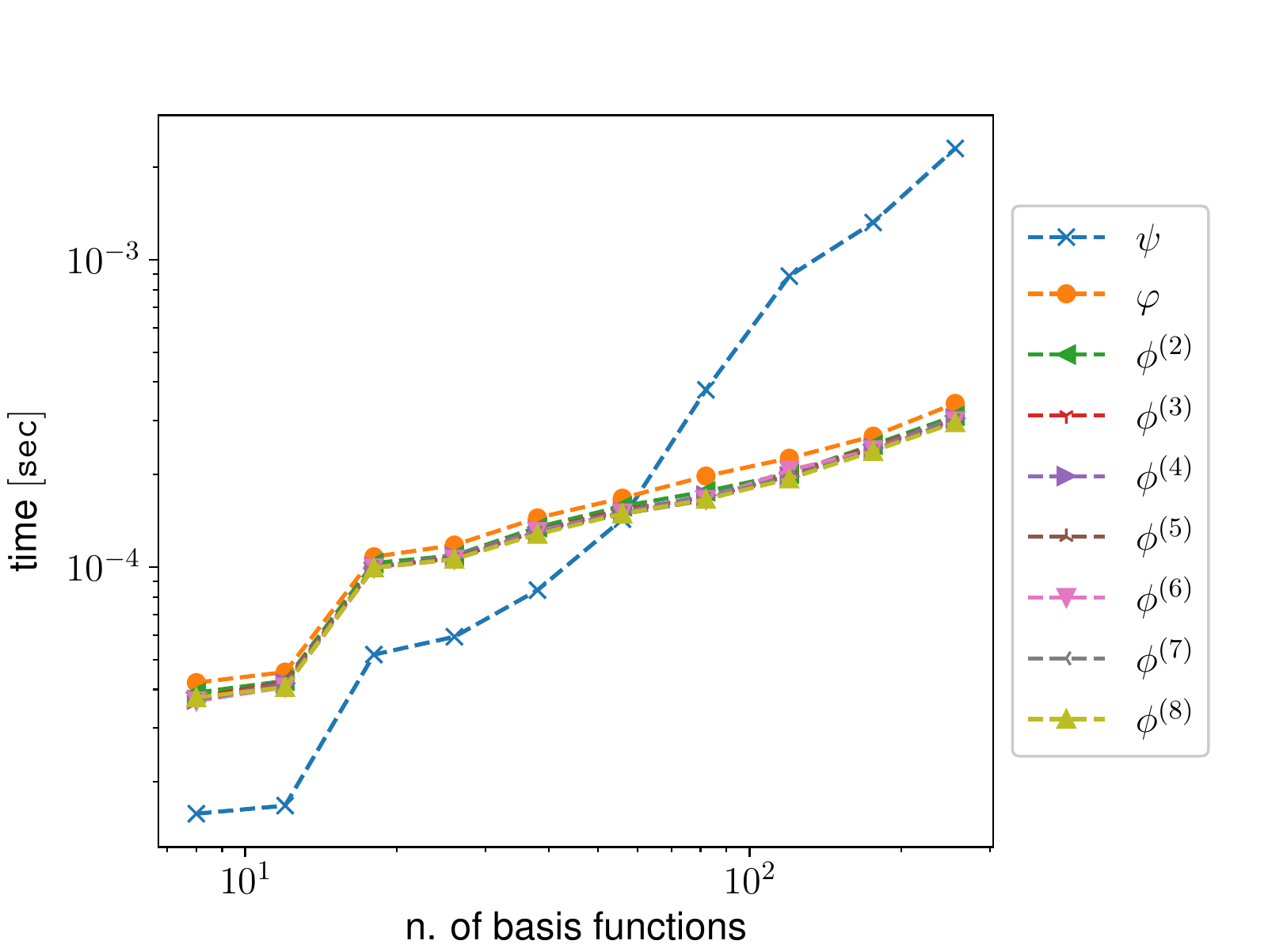}};
        \end{tikzpicture}
    }
    \caption{Average computational time when solving the minimization problem \eqref{eq:lin_p_1} as function of the number of basis functions $N$.}
    \label{fig:basis_time}
\end{figure}

To test the improvement in computational time, we perform the following test.
For different values of the number of basis functions $N$, we compute the matrix $ \mtrx{A} $ in \eqref{eq:lin_p_1}.
Then, for each value of $N$, we consider thirty different values for vector $ \vect{b} $ and solve the linear problem \eqref{eq:lin_p_1} saving the time needed to solve it.
We perform this test for the GBFs \eqref{eq:gaussian_basis_def}, mollifier-like basis functions \eqref{eq:mollifier_basis_def}, and Wendland basis functions \eqref{eqs:wendland_set}.

\noindent
Figure~\ref{fig:basis_time} shows the result of this test.
In particular, the log-scale plot shows the average computational time for each value of $N$.
As it can be seen, as the number of basis functions increases, the computational time needed to solve the linear problem \eqref{eq:lin_p_1} has a greater growth when GBFs are used.
On the other hand, when compactly supported basis functions are adopted, the increment in computational time is reduced.

\noindent
Tests were performed on a notebook with a quad-core Intel Core i7-7000 CPU with 16 GB of RAM.

%%% ---------------- %%%
%%%  UPDATE RESULTS  %%%
%%% ---------------- %%%
\subsubsection{Trajectory Update}
\label{subsec:trj_up_result}

\begin{figure}[t]
    \centering
    \resizebox*{0.9\columnwidth}{!}{
        \begin{tikzpicture}
            \clip (-8.0, 5) rectangle (7, -6.2);
            \node () at (0, 0) {\includegraphics{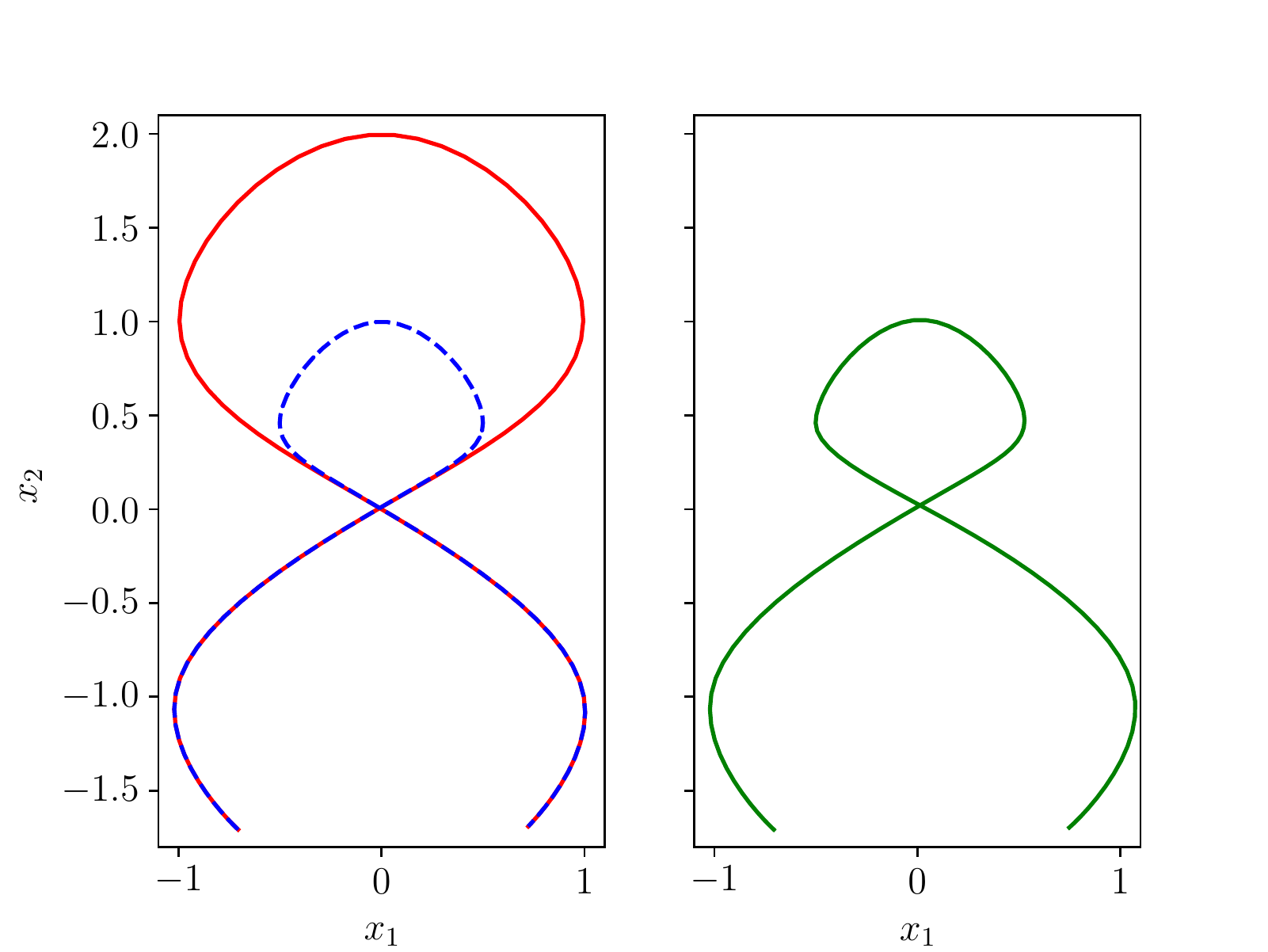}};
        \end{tikzpicture}
    }
    \caption{
        Result for the `trajectory update' property of compactly supported basis functions.
        On the left, the two trajectories are shown, showing that they differ only in the central portion.
        On the right, the execution of the obtained DMP is shown.
        The DMP is initially learned from the `big' trajectory, and then updated using the middle portion of the small one.
    }
    \label{fig:traj_update}
\end{figure}

In this Section, we present a synthetic test to show the ``trajectory update'' property presented in Section~\ref{subsec:trj_update}.
To do so, we generate two trajectories, which can be seen on the left plot in Figure~\ref{fig:traj_update}, using clamped splines (to have null initial and final velocities).
The trajectories are identical on the `tails' but they differ in the central portion (one is larger than the other).
To show the update properties of compactly supported basis functions, we start by learning a DMP from the `largest' trajectory (shown in red in Figure~\ref{fig:traj_update}).
DMPs parameters are $ \mtrx{K} = K \, \midentity_2 $ and $ \mtrx{D} = D \, \midentity_2 $ with $ K = 150 $ and $ D = 2\sqrt{K} \approx 24.49 $, $ \alpha = 4 $, and $ N = 100 $.

\noindent
Next, we identificate, using \eqref{eq:w_up_condition}, the set of basis functions that have to be updated when the middle portion of the trajectory has to be updated.
This result in the set of indexes $ I = \{ 24, 25, \ldots, 63 \} \subset \{ 0, 1, \ldots, N \} = \{ 0, 1, \ldots, 100 \} $.
Then, we re-compute the set of weights $ \{ \vectgreek{\omega}_i \}_{i \in I} $.

\noindent
The plot on the right in Figure~\ref{fig:traj_update} shows the execution of the resulting, updated DMP.
As one can observe, by updating only a subset of the weights the resulting trajectory mimics the new desired behavior.

%%% ---------------- %%%
%%%  CONSIDERATIONS  %%%
%%% ---------------- %%%
\subsection{General Considerations}

The tests performed in Section~\ref{sec:basis_results} showed the advantages of using compactly supported basis functions.

\noindent
Firstly, these sets of functions give a sparse matrix in minimization problem \eqref{eq:lin_p_1}.
As a consequence, the learning phase is both numerically faster and more stable than when using GBFs \eqref{eq:gaussian_basis_def}.

\noindent
Secondly, compactly supported basis functions allow to `update' a DMP when only a portion of the trajectory has to be modified, instead of learning a new one from scratch.

\noindent
Finally, the usage of compactly supported basis functions ensures a faster convergence of the DMP system to the goal position since the resulting forcing term is null after a finite amount of time, \( f(s(t)) = 0, \forall t > T^\star \) \cite{WWCT16}.
This is not true for (both classical and truncated) Gaussian basis functions \eqref{eq:gaussian_basis_def}, \eqref{eq:truncated_gaussian}.
Indeed, the support of the classical GBFs is the whole real line $ \RR $; while the support for TGBFs is an interval $ (- \infty, L] $, $ L \in \RR $ in $s$, which is an interval $ [L', + \infty) $, $ L' \in \RR $ in $t$.

Among the families of compactly supported basis functions we presented in this work, our proposed mollifier-like basis functions \eqref{eq:mollifier_basis_def} are both the most regular and the best approximants.

For these reasons, we argue that mollifier-like basis functions may become the new standard when using DMPs.

\begin{rmk}
    The numerical advantages (stability and time efficiency) showed by compactly supported basis functions influence only the learning phase.
    Indeed, since the approximation errors are similar and, overall, small among the different classes of basis functions, the resulting DMP execution will have no meaningful difference.
\end{rmk}

As a final observation, we remark that other approaches based upon basis functions approximation (such as ProMPs \cite{PDPN13}) could take advantage of the computational stability and efficiency of compactly supported basis functions.

% ---------------------------------------------------------------------------- %
% INVARIANCE
% ---------------------------------------------------------------------------- %

\section{Invariance Under Invertible Linear Transformations}\label{sec:linear_invariance}

Invariance under translations of DMPs formulations \eqref{eqs:dmp_old_form} and \eqref{eqs:new_dmps_vector} is straightforward.
Indeed, by changing the initial position from $\vect{x}_0$ to $\vect{x}_0'$, and by considering, instead of $\vect{g}$, the new goal position $\vect{g}' = \vect{g} + (\vect{x}_0' - \vect{x}_0)$, the whole trajectory is translated by a quantity $\vect{x}_0' - \vect{x}_0$.

DMPs are able to adapt to small changes of the relative position between goal and start, $ \vect{g} - \vect{x}_0 $; however, the robustness of DMPs w.r.t. sensible changes of vector $ \vect{g} - \vect{x}_0 $ heavily depends on the choice of the hyperparameters $\mtrx{K}, \mtrx{D}$ in \eqref{eqs:new_dmps_vector} and $\alpha$ in \eqref{eq:canonical_system} (see, for instance, the tests performed in Figures~\ref{fig:generalization}, \ref{fig:generalization_2}).
Generalization to arbitrary changes of quantity $\vect{g} - \vect{x}_0$ can be achieved by exploiting the invariance property of equations \eqref{eqs:new_dmps_vector} w.r.t. (invertible) affine transformations, see \cite{HPPS09}.
At the best of the authors' knowledge, this well-known property has never been exploited in order to make DMPs globally robust w.r.t. arbitrary changes of $ \vect{g} - \vect{x}_0 $.

\noindent
The invariance of \eqref{eqs:new_dmps_vector} can be easily proven (see \cite{HPPS09}).
Consider the invertible transformation matrix $\mtrx{S} \in \RR^{d \times d}$.
By substituting
\begin{equation}
    \label{eq:transformation}
    \begin{aligned}
        \vect{x}' & = \mtrx{S} \vect{x}, & \vect{v}' & = \mtrx{S} \vect{v}, & \vect{x}_0' & = \mtrx{S} \vect{x}_0, \\
        \vect{g}' & = \mtrx{S} \vect{g}, & \mtrx{K}' & = \mtrx{S} \mtrx{K} \mtrx{S}^{-1}, & \mtrx{D}' & = \mtrx{S} \mtrx{D} \mtrx{S}^{-1},
    \end{aligned}
\end{equation}
in \eqref{eqs:new_dmps_vector} we obtain the transformation law of the forcing term
\begin{equation}
    \label{eq:force_transformation}
    \vect{f}' = \mtrx{S} \vect{f}.
\end{equation}
Thus, if we want to generate a new trajectory $\mtrx{S} \vect{x}$, it is sufficient to apply the transformations in \eqref{eq:transformation} and \eqref{eq:force_transformation} to \eqref{eqs:new_dmps_vector}.
The resulting DMP system, thus, reads
\begin{subnumcases}{\label{eqs:extended_dmps_vector}}
    \begin{multlined}
        \tau \dot{\vect{v}} =
            \mtrx{K} ' (\vect{g} - \vect{x}) -
            \mtrx{D} ' \vect{v} -
            \mtrx{K} ' (\vect{g}- \vect{x}_0)s + \\
            \mtrx{K} ' \vect{f}(s)
    \end{multlined}
        \label{eq:extended_dmps_vector_acc} \\
    \tau \dot{\vect{x}} = \vect{v} \label{eq:extended_dmps_vector_vel}
\end{subnumcases}

We remark that if the elastic and damping terms are the same for each degree-of-freedom, then $\mtrx{K}$ and $\mtrx{D}$ are multiple of the identity matrix and thus $\mtrx{K} = \mtrx{K}'$ and $\mtrx{D} = \mtrx{D}'$.

\subsection{Invariance Under Roto-Dilatation}\label{sec:inv_theory}

A case of particular interest is when $ \mtrx{S} $ in \eqref{eq:transformation}, \eqref{eq:force_transformation} represents a \emph{roto-dilatation}.

\noindent
Consider a trajectory which is learned together with the relative position between the goal and the starting point, i.e. the vector $\vect{g} - \vect{x}_0$; and a new trajectory, with starting and ending points $ \vect{x}_0' $ and $ \vect{g} ' $ respectively has to be reproduced.

\noindent
We first compute the rotation matrix $ \mtrx{R}_{ \widehat{\vect{g} - \vect{x}_0 } } ^ { \widehat{ \vect{g}' - \vect{x}_0'}} $, which maps the unit vector $ \widehat{\vect{g} - \vect{x}_0} $ to $ \widehat{\vect{g}' - \vect{x}_0' }$, where operator $ \widehat{\cdot} $ denotes the normalization: \( \widehat{\vect{v}} \defas \vect{v} / \| \vect{v} \| \).
Rotation matrix $ \mtrx{R}_{ \widehat{\vect{g} - \vect{x}_0 } } ^ { \widehat{ \vect{g}' - \vect{x}_0'}} $ can be computed using the algorithm presented in \cite{Zhe17}.

\noindent
After performing the rotation, we perform a dilatation of \( \| \vect{g}' - \vect{x}_0' \| / \| \vect{g} - \vect{x}_0 \| \) obtaining the transformation matrix
\begin{equation}
    \label{eq:rotodilatation_matrix}
    \mtrx{S} = \frac{ \norm{ \vect{g}' - \vect{x}_0' } }{ \norm{ \vect{g} - \vect{x}_0 } } \, \mtrx{R}_{\widehat{\vect{g} - \vect{x}_0}} ^ { \widehat{ \vect{g}' - \vect{x}_0'} }.
\end{equation}
Therefore, when a learned DMP has to be used to generate a new trajectory, characterized by the parameters $\vect{x}_0'$ and $\vect{g}'$, we compute the matrix $\mtrx{S}$ as in \eqref{eq:rotodilatation_matrix}, and then the quantities $\mtrx{K}'$, $ \mtrx{D}'$, and $\vect{f}'$ as in \eqref{eq:transformation} and \eqref{eq:force_transformation}.

\noindent
This approach can be used even if the goal position changes with time during the execution of the trajectory simply by updating the matrix $\mtrx{S}$ at each time.
This, in particular, means that $ \mtrx{S} $ depends on time.
In Section~\ref{sec:invariance_results} we will test this approach both for static and moving goal positions.

We remark that this method cannot be performed when starting and ending points coincide: $ \vect{x}_0 = \vect{g} $, since the matrix $\mtrx{S}$ in \eqref{eq:rotodilatation_matrix} would result in a null matrix, $\mtrx{S} = \mtrx{0}$.
However, in such cases, one should use \emph{rhythmic} (or \emph{periodic}) DMPs instead of discrete ones, which are beyond the scope of this paper.

\begin{rmk}
    The choice to use a roto-dilatation as the transformation matrix $\mtrx{S}$ is not a unique possibility.
    Indeed, any invertible matrix can be used.
    For example, if we write $\mtrx{S}$ as the diagonal matrix \( \mtrx{S} = \text{diag}(s_1, s_2, \ldots, s_d) \) in which each component of the diagonal is defined as (assuming that the learned goal and starting positions differ in each component)
    \[ s_i = \frac{ \vect{g}'_i - \vect{x}_{0\, i}' }{ \vect{g}_i - \vect{x}_{0\, i} } ,\quad i = 1,2,\ldots, d , \]
    we would obtain a formulation similar to \eqref{eqs:old_dmps_vector}, in which each component is scaled by $ g - x_0 $.
\end{rmk}

\begin{rmk}
    Choosing roto-dilatation has the advantage that the evaluation of the inverse transformation does not require numerically inverting the matrix.
    Indeed, since the matrix $ \mtrx{S} $ of the transformation can be written as a non-zero scalar value multiplied by an orthogonal matrix
    \[ \mtrx{S} = a \, \mtrx{R}, \quad a \in \RR \setminus\{0\}, \,\mtrx{R} \in \RR^{d\times d}, \, \mtrx{R}^{-1} = \mtrx{R}\transpose, \]
    where $\mtrx{R}$ is the rotation matrix, the inverse is simply
    \[ \mtrx{S}^{-1} = \frac{1}{a} \, \mtrx{R}^{-1} = \frac{1}{a} \, \mtrx{R}\transpose. \]
    Thanks to this property, the computation of the inverse matrix $ \mtrx{S}^{-1} $ in \eqref{eq:transformation} can be performed efficiently (since transposing a matrix is faster than inverting one).
\end{rmk}

\subsection{Results}\label{sec:invariance_results}

In this Section, we present several synthetic tests and robotic experiments to show how DMP formulation \eqref{eqs:extended_dmps_vector} results in better behaviors when compared to formulation \eqref{eqs:new_dmps_vector}.

In the following, we refer to \emph{classical} DMPs when talking about the implementation without exploiting the invariance property.
Similarly, we refer to \emph{extended} DMPs when talking about the formulation \eqref{eqs:extended_dmps_vector} we introduced in Section~\ref{sec:inv_theory}, in which the invariance property is exploited by using, as linear invertible transformation, the roto-dilatation defined in \eqref{eq:rotodilatation_matrix}.

\begin{figure*}[t]
    \centering
    \subfloat[Rotation\label{subfig:rotation}]{\includegraphics[width=0.8\columnwidth]{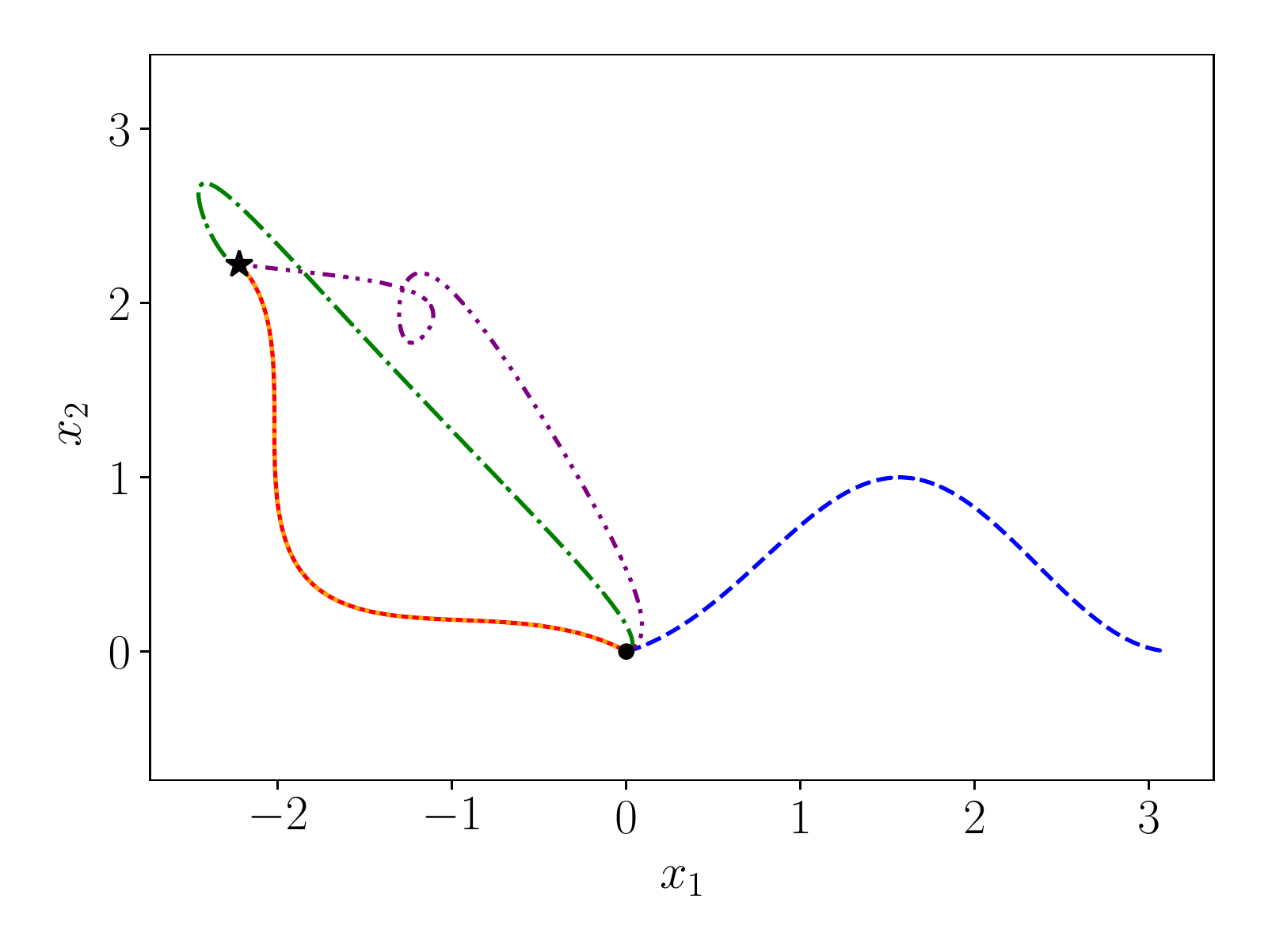}}
    \hspace{2cm}
    \subfloat[Dilatation\label{subfig:dilate}]{\includegraphics[width=0.8\columnwidth]{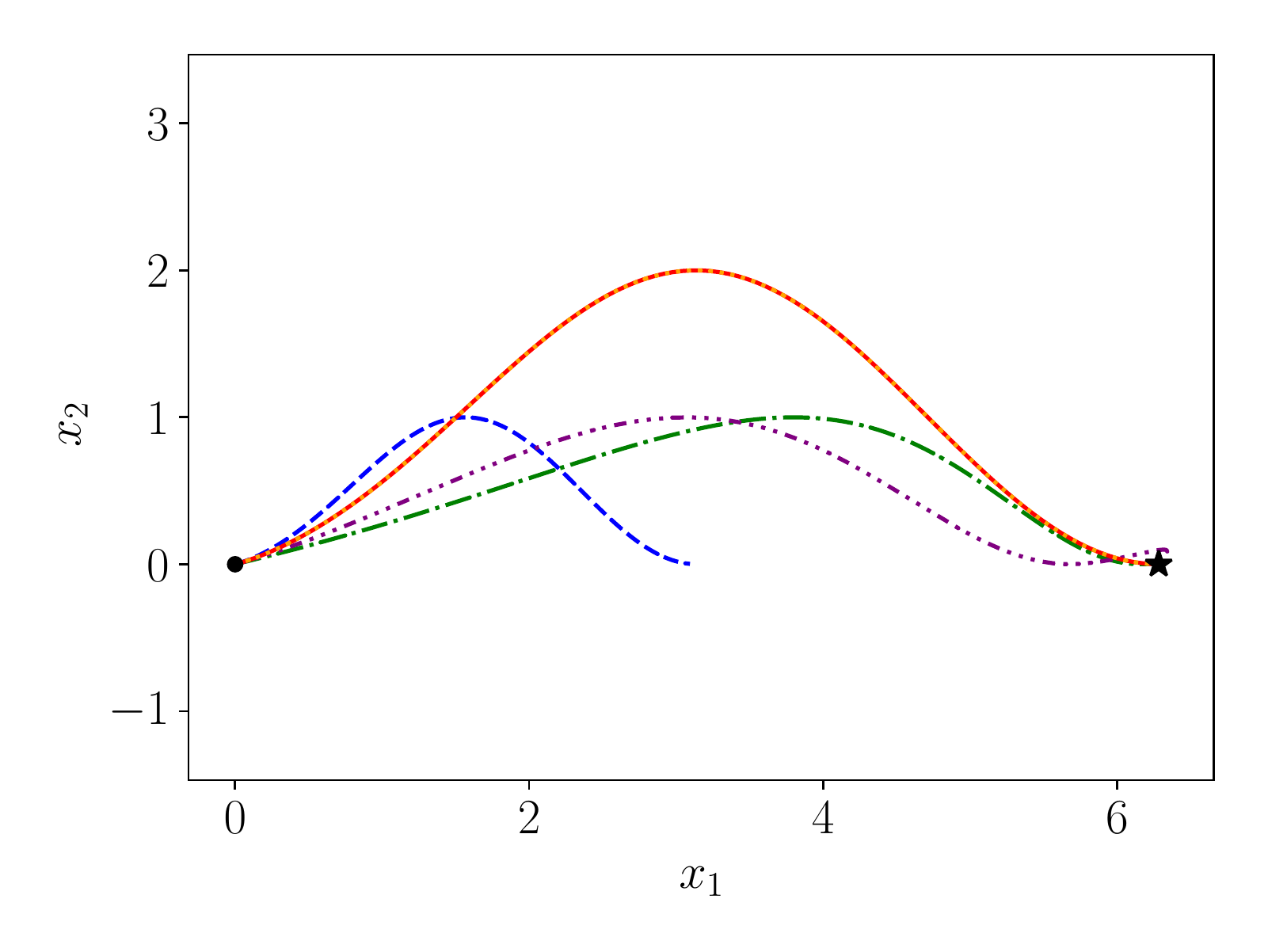}}\\
    \subfloat[Shrinking\label{subfig:shrink}]{\includegraphics[width=0.8\columnwidth]{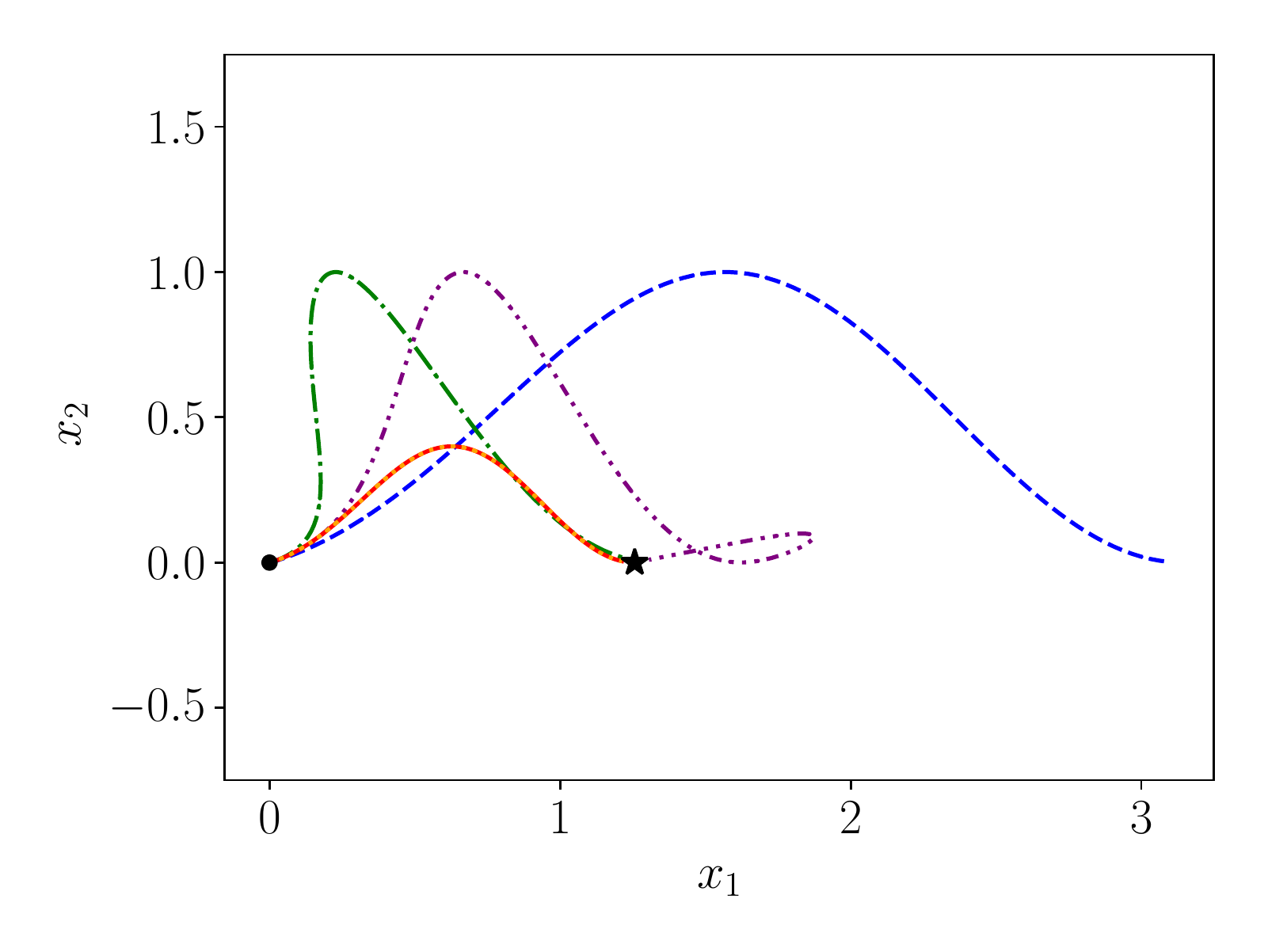}}
    \hspace{2cm}
    \subfloat[Moving goal\label{subfig:moving}]{\includegraphics[width=0.8\columnwidth]{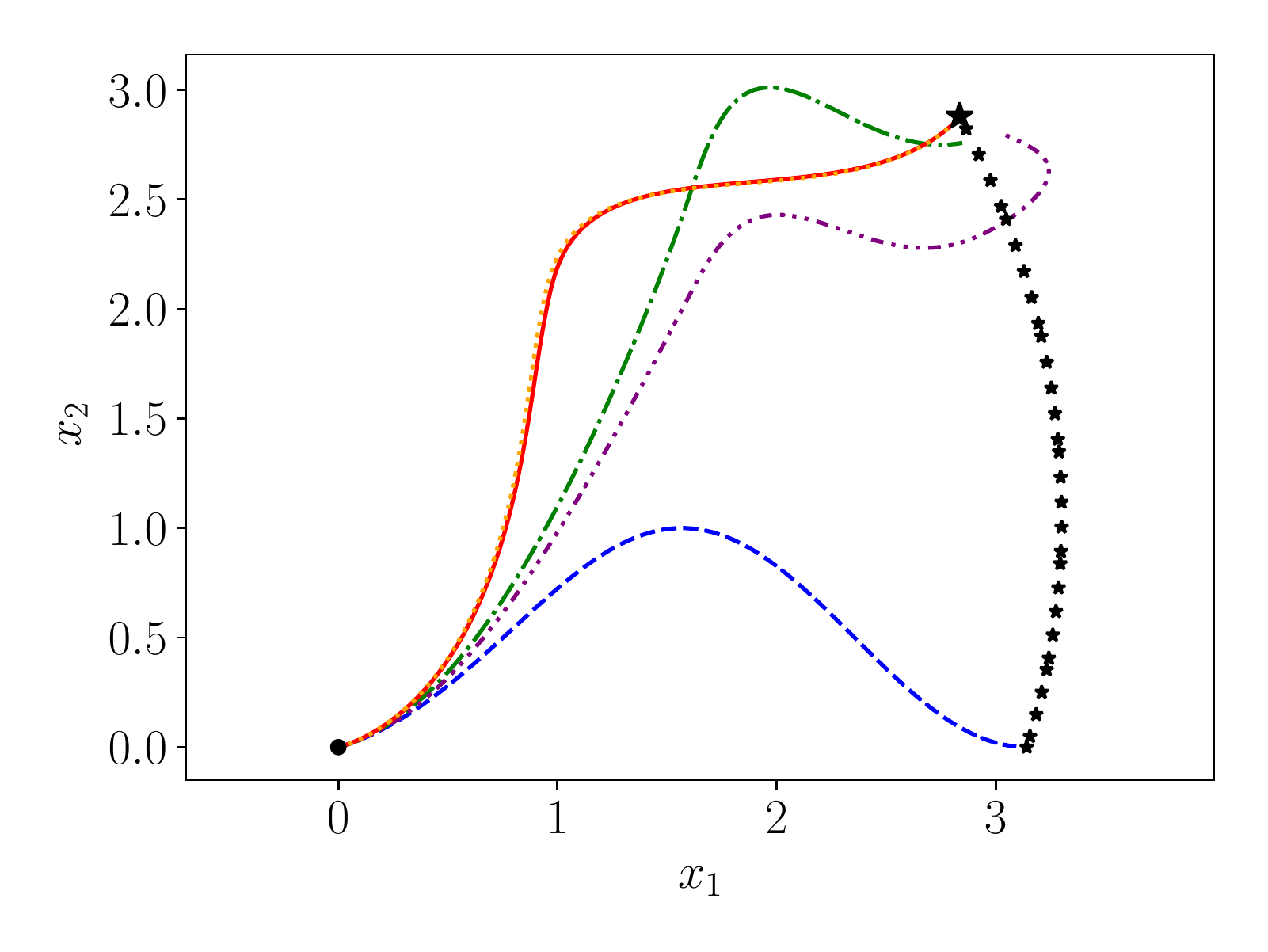}}
    \caption{
        Different behaviors obtained when changing the goal position.
        In Figures~\ref{subfig:rotation}, \ref{subfig:dilate}, and \ref{subfig:shrink} the goal is different from the learned one and still.
        In Figure~\ref{subfig:moving} the goal starts in the learned position and moves continuously.
        The desired curve is given by \eqref{eq:demo_scaling_1}.
        In all four plots, the desired curve is plotted using the blue dashed line.
        The execution of extended DMP is shown by the red full line for $ \alpha = 4 $, and by the orange dotted line for $ \alpha = 2 $ (in all these experiments, they overlap).
        The execution of classical DMPs are marked with the dash-dotted green line when $ \alpha = 4 $, and with the purple one dash three dots line when $ \alpha $ is set to $2$.
        The dot and star mark, respectively, the desired starting and goal positions $ \vect{x}_0 $ and $ \vect{g} $.
        The stars trial in Figure~\ref{subfig:moving} shows the movement of the goal.
    }
    \label{fig:generalization}
\end{figure*}

\subsubsection{Synthetic Tests.}

\begin{figure*}[t]
    \centering
    \subfloat[Rotation\label{subfig:rotation_2}]{\includegraphics[width=0.8\columnwidth]{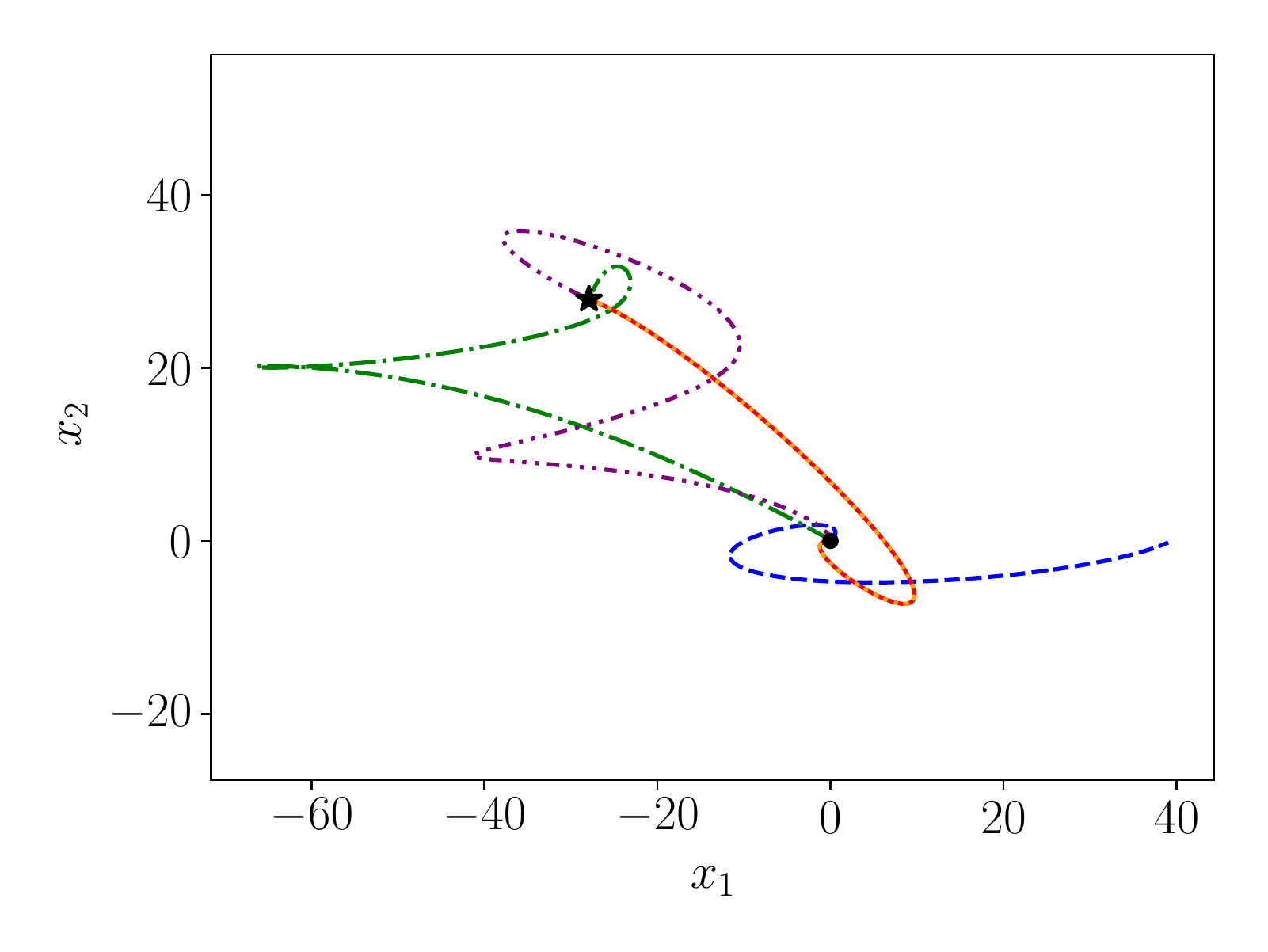}}
    \hspace{2cm}
    \subfloat[Dilatation\label{subfig:dilate_2}]{\includegraphics[width=0.8\columnwidth]{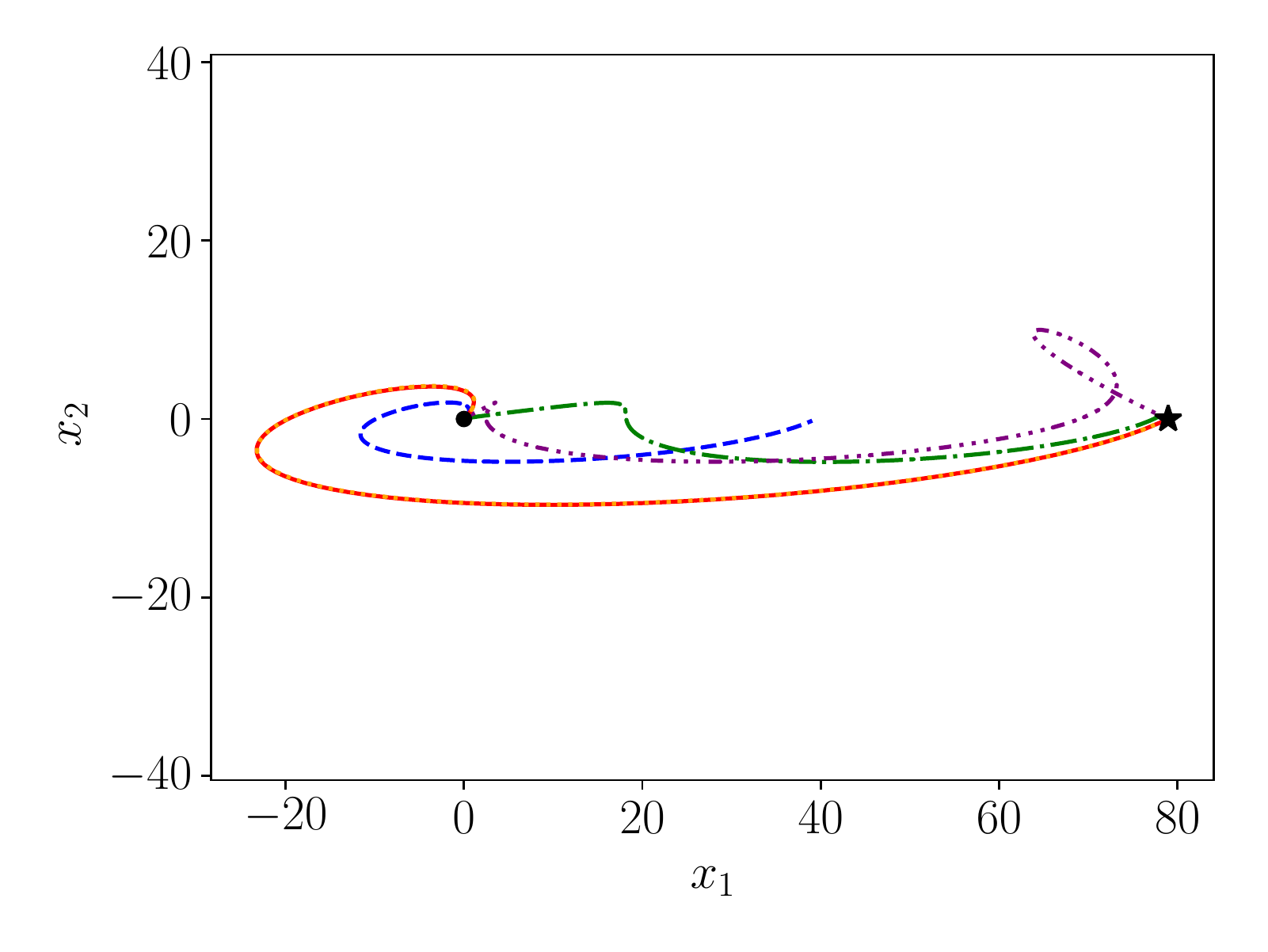}}\\
    \subfloat[Shrinking\label{subfig:shrink_2}]{\includegraphics[width=0.8\columnwidth]{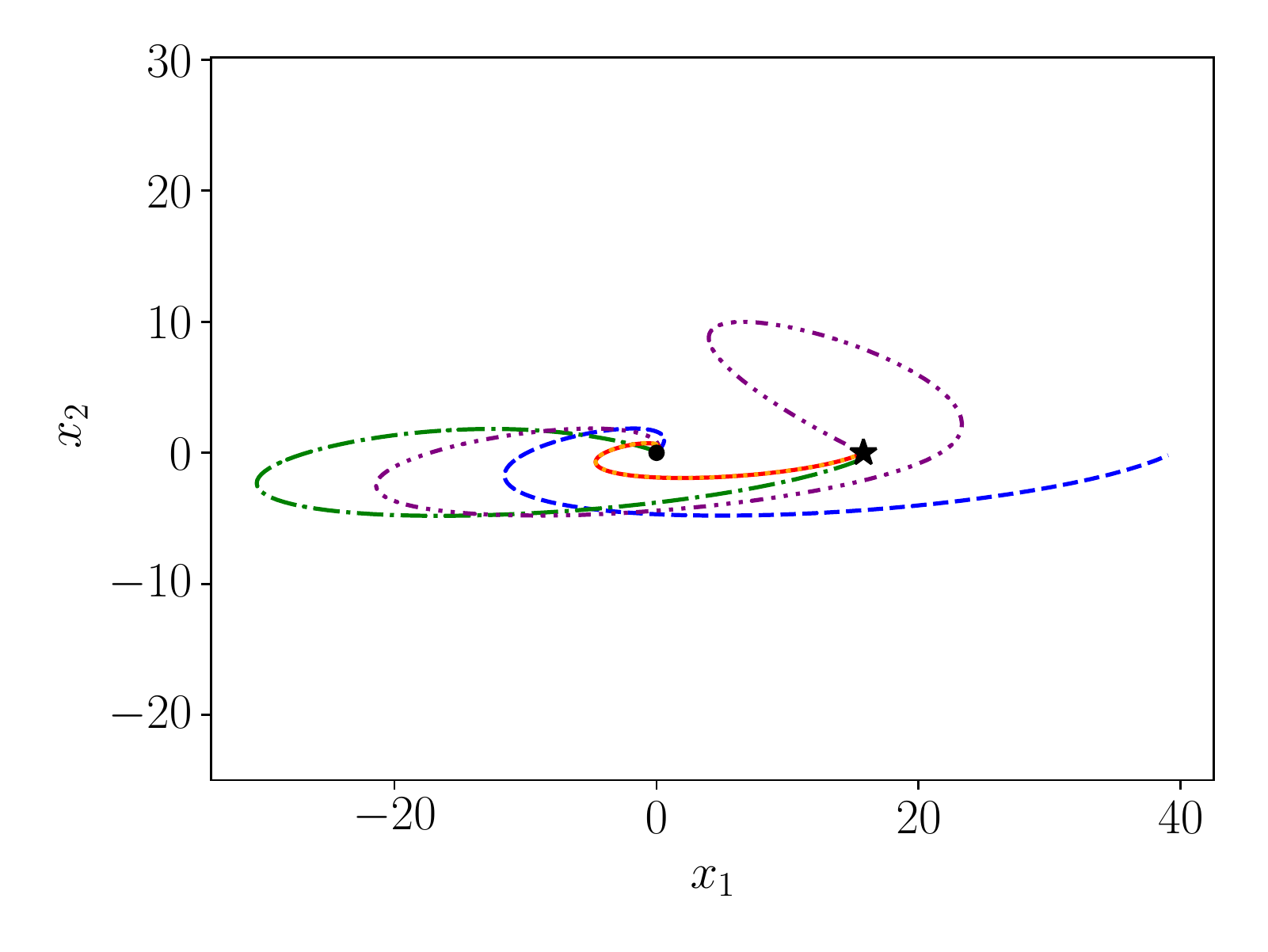}}
    \hspace{2cm}
    \subfloat[Moving goal\label{subfig:moving_2}]{\includegraphics[width=0.8\columnwidth]{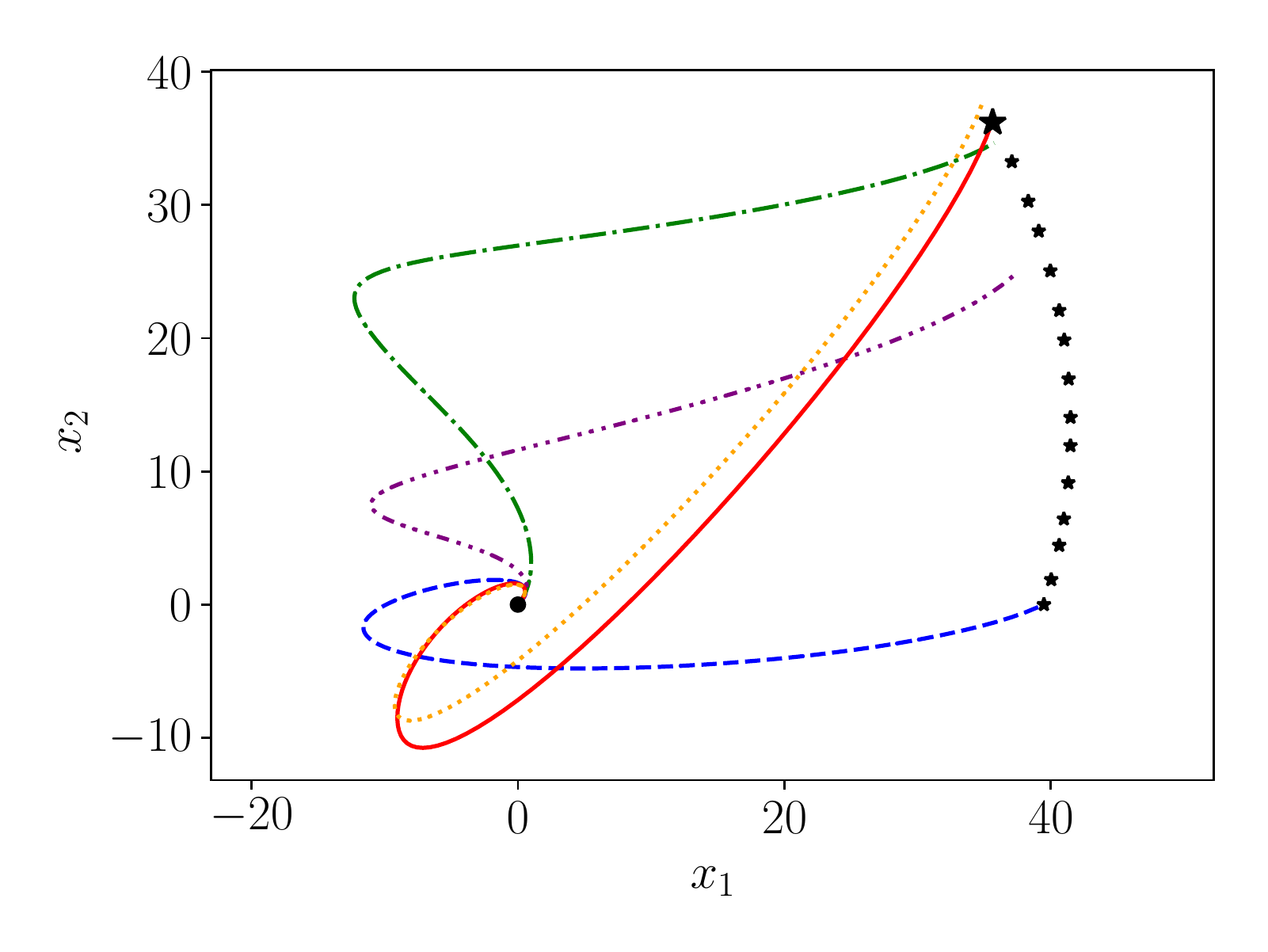}}
    \caption{
        Different behaviors obtained when changing the goal position.
        In Figures~\ref{subfig:rotation_2}, \ref{subfig:dilate_2}, and \ref{subfig:shrink_2} the goal is different from the learned one and still.
        In Figure~\ref{subfig:moving_2} the goal starts in the learned position and moves continuously.
        The desired curve is given by \eqref{eq:demo_scaling_2}.
        In all four plots, the desired curve is plotted using the blue dashed line.
        The execution of extended DMP is shown by the red full line for $ K=150 $, and by the orange dotted line for $ K = 15 $ (in Figures~\ref{subfig:rotation_2}--\ref{subfig:shrink_2}, they overlap).
        The execution of classical DMPs are marked with the dash-dotted green line when $ K=150 $, and with the purple one dash three dots line when $ K $ is set to $15$.
        The dot and star mark, respectively, the desired starting and goal positions $ \vect{x}_0 $ and $ \vect{g} $.
        The stars trial in Figure~\ref{subfig:moving_2} shows the movement of the goal.
    }
    \label{fig:generalization_2}
\end{figure*}

We investigate the robustness gain ensured by the invariance property under rotation and dilatation of the reference system by considering two examples in which a trajectory is learned and then executed changing the relative position between goal and starting point $\vect{g} - \vect{x}_0$.
In these tests, we do not change $ \vect{x}_0 $ since DMPs are natively translational invariant.
We compare the different behaviors obtained with and without the scaling term $ \mtrx{S} $ defined in \eqref{eq:rotodilatation_matrix}.\footnote{
We will not compare the results with the original DMPs \eqref{eqs:old_dmps_vector} since the drawbacks of such formulation have already been discussed in the literature \cite{PHPS08, HPPS09, PHAS09}.}
For all the tests presented in this Section, we use mollifier-like basis functions \eqref{eq:mollifier_basis_def}.
However, we remark that tests done with other classes of basis functions give similar results.

In the first simulation, we generate the desired curve in the plane:
\begin{equation}
    \label{eq:demo_scaling_1}
    \vect{x}(t) = \br{t, \sin^2(t) },\quad t \in [0, \pi] .
\end{equation}    
Then we perform the learning step to compute the weights $\vectgreek{\omega}_i$, and we test the generalization properties of DMPs by changing in different manners the goal position.
In particular, in Figures~\ref{subfig:rotation}, \ref{subfig:dilate}, and \ref{subfig:shrink} the goal is set to a different configuration and is still.
Instead, in Figure~\ref{subfig:moving}, the goal starts in the learned position and moves towards a final target configuration while the DMP is being executed.
All tests are executed using both classical and extended DMPs.

\noindent
We performed these tests with elastic term $K = 150 $, and damping term $D = 2\sqrt{150} \approx 24.49$ for both components $x_1$ and $x_2$, that is $ \mtrx{K} = K \mtrx{I}_2 $ and $ \mtrx{D} = D \mtrx{I}_2 $, being $ \mtrx{I}_2 $ the $ 2 \times 2 $ identity matrix.
We test the generalization by using two different values of $\alpha$ in the canonical system \eqref{eq:canonical_system}, $\alpha = 4$ and $\alpha = 2$.
Both these values for $ \alpha $ result in a canonical system that vanishes at the final time within a tolerance smaller than $ 1\% $.
Indeed, since the final time is $ t_1 = \pi $ in \eqref{eq:demo_scaling_1}, for $ \alpha = 2 $ we have \( \exp( -\alpha \, T ) = \exp (-2\pi) \approx 1.9 \texttt{e} -03 \); while for $ \alpha = 4 $ we have \( \exp ( -\alpha\,T ) = \exp(-4\pi) \approx 3.5\texttt{e}-06 .\)
Figure~\ref{fig:generalization} shows the results of our tests.

\noindent
In the second simulation, the desired curve is:
\begin{equation}
    \label{eq:demo_scaling_2}
    \vect{x}(t) = \br{t^2 \, \cos(t), t \, \sin(t) },\quad t \in [0, 2\pi] .
\end{equation}
Also in this case we test the generalization properties of both classical DMPs and extended DMPs when the relative position $ \vect{g} - \vect{x}_0 $ is changed.
In Figures~\ref{subfig:rotation_2}, \ref{subfig:dilate_2}, and \ref{subfig:shrink_2} the goal is set to a different configuration before we start executing the DMP.
Instead, in Figure~\ref{subfig:moving_2}, the goal starts in the learned position and moves towards a final target configuration while the DMP is being executed.
The main difference is that in this second test we set $ \alpha = 4 $, and we use two different values for $ \mtrx{K} $ (and, thus, for $ \mtrx{D} $).
Indeed, we set $\mtrx{K} = K \mtrx{I}$ and set $K$ to assume value $150$ and $15$.
In both cases, the damping term is set to $ \mtrx{D} = \sqrt{K} \, \mtrx{I}_2 $.

Both these tests show how extended DMPs are more robust than the classical DMP formulation since the trajectories generated by taking advantage of the invariance property have a shape that closely resembles the learned trajectory, while classical DMPs may generate trajectories that do not resemble the learned behavior.
Moreover, we notice that the goodness of the generalization of classical DMPs heavily depends on the choice of the parameters.
For instance, in the cases of dilatation and moving goal for \eqref{eq:demo_scaling_1}, they generalize well when $\alpha = 4$, but fails when $\alpha = 2$ (see Figure~{\ref{subfig:dilate}} and \ref{subfig:moving}).
Similarly, we observe that the generalization of classical DMPs heavily depends also on the choice of parameter $\mtrx{K}$ (and, consequently, $\mtrx{D}$).
For instance, Figure~\ref{subfig:moving_2} shows that the trajectories are different, even if the change in goal position is the same.
On the other hand, we see that the generalization of the trajectory is robust against the particular choice of hyperparameters when using extended DMPs.
Indeed, in almost all our tests, the generalizations obtained by extended DMPs completely overlap when changing the goal position.
The only case in which there is an actual difference, even if hardly noticeable, is for the test in Figure~\ref{subfig:moving_2}, i.e. when the goal is moving and we compare two DMPs with different values of $ \mtrx{K} $ (and, thus, $ \mtrx{D} $).
This is due to two factors: the scaling matrix $\mtrx{S}$ depends explicitly on time (since $ \vect{g} $ is moving), and different values for $\mtrx{K}$ influence the un-perturbed evolution of dynamical system \eqref{eqs:new_dmps_vector}.
On the other hand, we observe that, in the case presented in \ref{subfig:moving}, there is no difference between the two trajectories, even if $ \mtrx{S}$ depends on time because the un-perturbed evolution of the system is not influenced by changes in $\alpha$.

\subsubsection{Tests on Real Robots.}
\label{subsec:scaling_test_real}

\begin{figure}[t]
    \centering
    \subfloat[Panda setup.\label{subfig:panda_setup}]{\includegraphics[height=0.47\columnwidth]{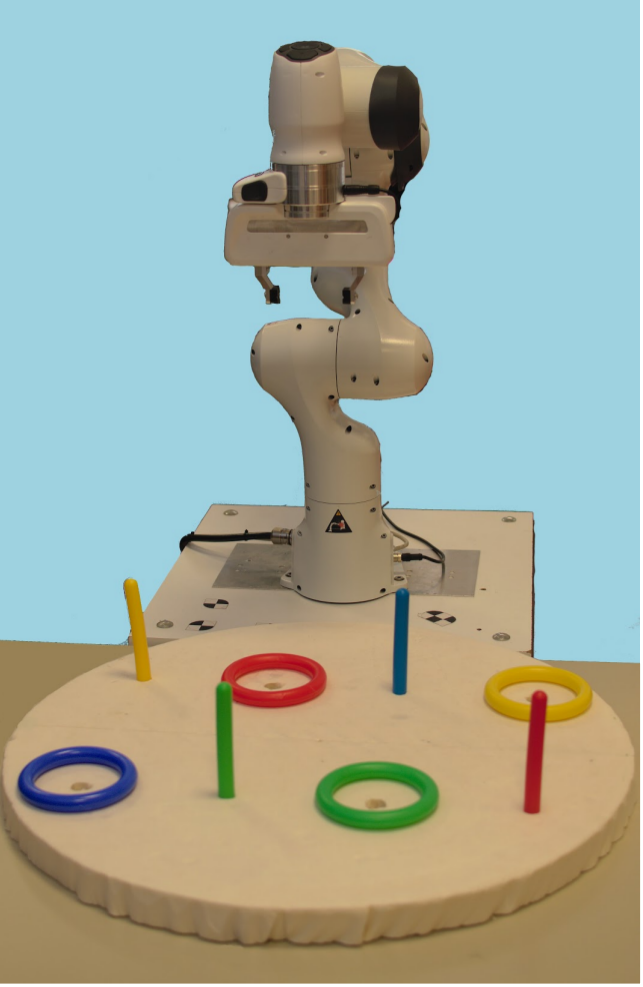}}
    \hspace{0.5cm}
    \subfloat[daVinci setup.\label{subfig:davinci_setup}]{\includegraphics[height=0.47\columnwidth]{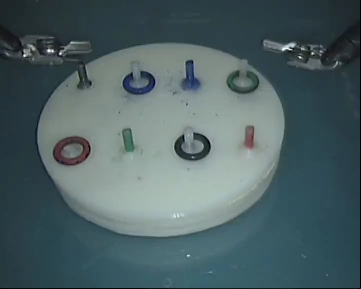}}
    \caption{Setups of the Peg \& Ring task.}
\end{figure}

On a real setup, we test our new formulation by performing a task on a 7 Degrees-of-Freedom (DoF) industrial manipulator, a Panda robot from Franka Emika shown in Figure~\ref{subfig:panda_setup}.
We will perform two tests.
In the first test, the learned behavior will be used to execute the task on the same robot, and we will show how extended DMPs \eqref{eqs:extended_dmps_vector} result in a better adaptation than classical DMPs \eqref{eqs:new_dmps_vector}.
The second test, instead, will show how extended DMPs \eqref{eqs:extended_dmps_vector} can be used to \emph{transfer} the learned behavior to a different setup, in our case, the daVinci surgical robot from Intuitive.

\noindent
The \emph{Peg \& Ring} task consists of grabbing, one at a time, four colored rings and moving them to the same-colored peg.
The task consists of two gestures, namely \texttt{move}, in which the end-effector has to reach the ring, and \texttt{carry}, in which the ring is moved toward the peg.
Automation of this task on surgical setups is a crucial aspect in Autonomous Robotic Surgery since it presents several challenges of real surgery (e.g. grasping) \cite{GMNRF19, GMRSF20a}.
During the learning phase, we learn a $3-$dimensional DMP for both gestures, on the Panda, via kinesthetic teaching.
For both tests, DMPs parameters are $ \mK = K\,\midentity_3 $ and $ \mD = D \, \midentity_3 $, with $ K=150 $ and $ D = 2\sqrt{K} \approx 24.49 $, and $ \alpha = 4 $.
% In the first test, we learn one DMP for each gesture, \texttt{move} and \texttt{carry}.
As it can be seen from Figure~\ref{subfig:panda_setup}, the four \texttt{carry} movements are qualitatively different for each color: the red and blue rings must be moved `in diagonal' along the base to be put in the corresponding pegs, while the green and yellow rings must be moved `horizontally'.

\begin{figure}[t]
    \centering
    \resizebox{0.9\linewidth}{!}{
    \begin{tikzpicture}
        \clip(-5.5, 4) rectangle (5, -5);
        \node() at (0,0) {\includegraphics{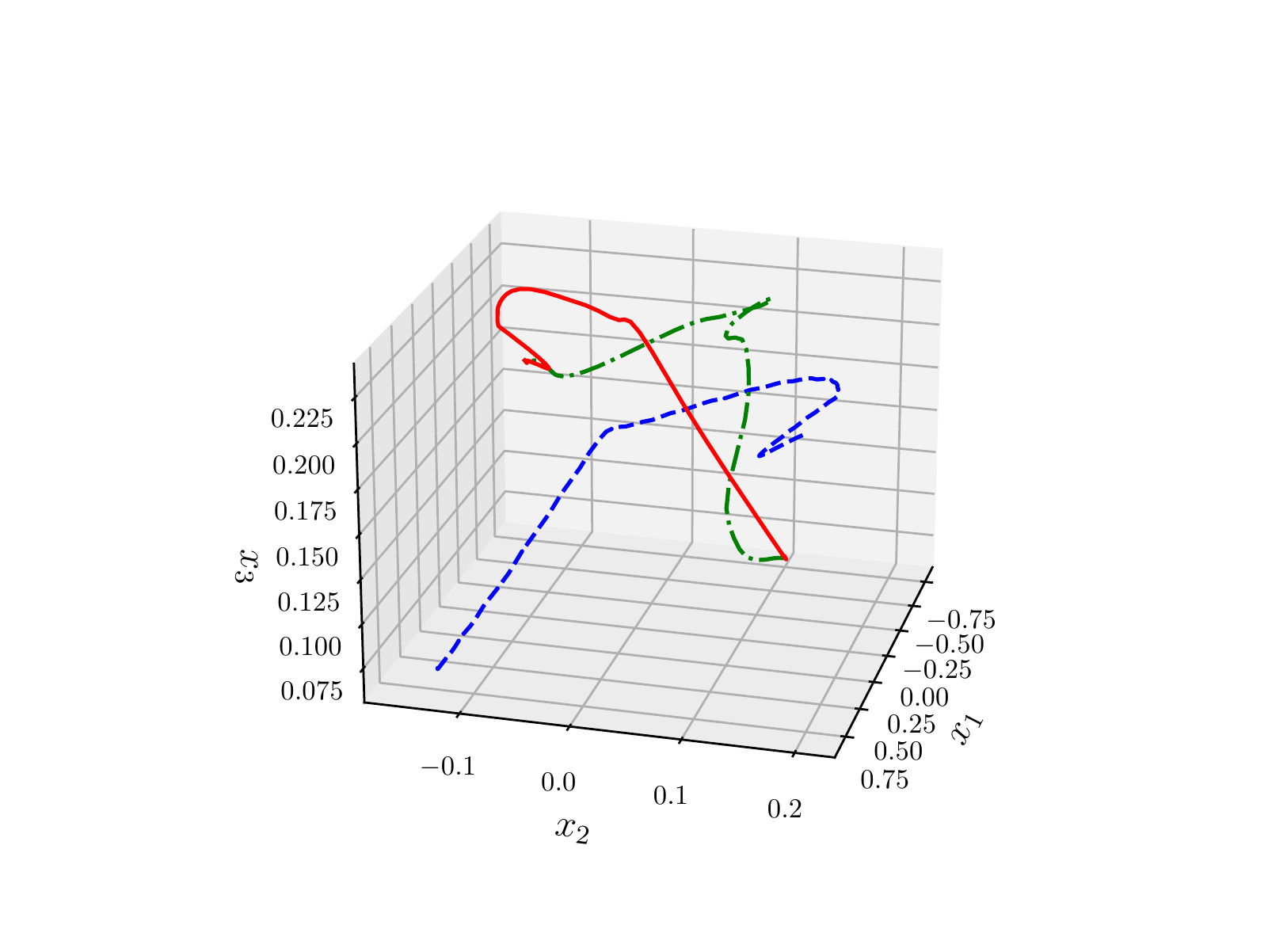}};
    \end{tikzpicture}
    }
    \caption{
        Generalization of extended DMPs \eqref{eqs:extended_dmps_vector}.
        The blue dashed line shows the desired behavior, learned from the \texttt{carry} movement for the red ring.
        The solid red line shows the executed extended DMP \eqref{eqs:extended_dmps_vector} for the starting and ending points for the \texttt{carry} movement for the blue ring.
        The green dash-dotted line shows the behavior obtained using classical DMPs \eqref{eqs:new_dmps_vector} adapted to the critical points of the blue ring.
    }
    \label{fig:transfer_l_r}
\end{figure}

To prove the adaptability of extended DMPs, we learn the desired behavior from the execution of the \texttt{carry} gesture for the red ring and use it to execute the gesture for all the rings in the scene.
Figure~\ref{fig:transfer_l_r} shows the result of this test.
In particular, one can observe that the shape of the executed trajectory (for the blue ring) maintains a trajectory of similar shape to the learned behavior (from the execution for the red ring).
For completeness, we plot the behavior obtained with classical DMPs \eqref{eqs:new_dmps_vector}.
From this, it is clear that the behavior of extended DMPs success in maintaining the same shape as the learned behavior, while classical DMPs fail in doing so.

\begin{figure*}[t]
    \centering
    \resizebox*{0.9\linewidth}{!}{
    \begin{tikzpicture}
        \clip (-19, 7.5) rectangle (12, -13);
        \node () at (2,0.5) {\includegraphics[width=1.5\linewidth]{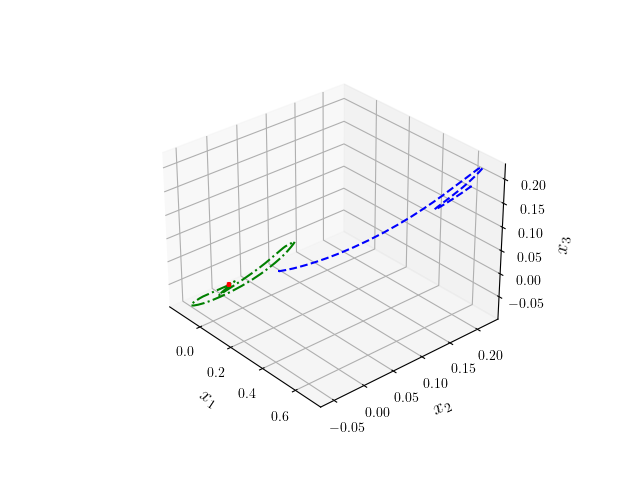}};
        \draw[ultra thick] (-3.7, -2.5) rectangle++ (+5.0, +3.6);
        % \draw[ultra thick] (-3.8, -2.3) rectangle++ (1, +1);
        % \draw (-3.8, -2.3) -- (-18, -4);
        % \draw (-3.8, -1.3) -- (-18, 4.5);
        % \draw (-2.8, -1.3) -- (-6, 4.5);
        % \draw (-2.8, -2.3) -- (-6, -4);
        % \draw[ultra thick, fill=white] (-18, -4) rectangle++ (12, 8.5);
        \draw (-3.7, 1.1) -- (-18.3, 10 - 3.0);
        \draw (-3.7, -2.5) -- (-18.3, 0 - 3.0);
        \draw (1.3, -2.5) -- (-5.6, 0 - 3.0);
        \draw (1.3, 1.1) -- (-5.6, 10 - 3.0);

        \draw[ultra thick, fill=white] (-18.3, 10- 3.0) rectangle++ (12.7, -10);
        \node () at (-13.0, 5.5 - 3.0) {\includegraphics[width=1.1\linewidth]{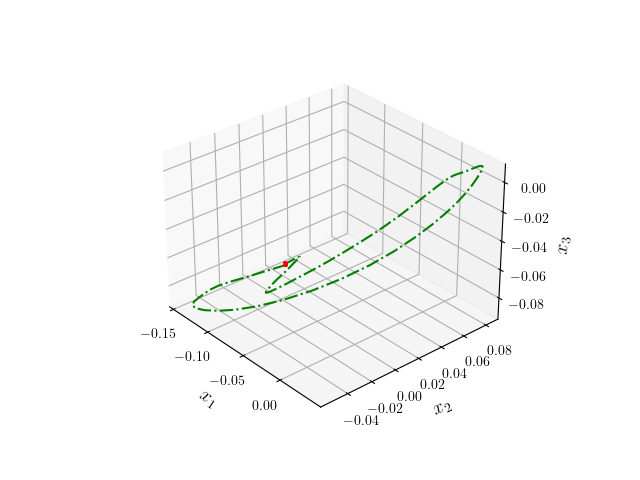}};
        \draw[ultra thick, dashed] (-14.5, 5.5- 3.0) rectangle++ (+1, -1);

        \draw[dashed] (-14.5, 5.5- 3.0) -- (-22.0 + 4.0, -1.5 - 2.0);
        \draw[dashed] (-14.5, 4.5- 3.0) -- (-22.0 + 4.0, -10.5 - 2.0);
        \draw[dashed] (-13.5, 5.5- 3.0) -- (-10.5 + 4.0, -1.5 - 2.0);
        \draw[dashed] (-13.5, 4.5- 3.0) -- (-10.5 + 4.0, -10.5 - 2.0);
        \draw[ultra thick, dashed, fill=white] (-22.0 + 4.0, -1.5 - 2.0) rectangle++ (11.5, -9.0);
        \node () at (-16.0 + 4.0, -5.5 - 2.0) {\includegraphics[width=1\linewidth]{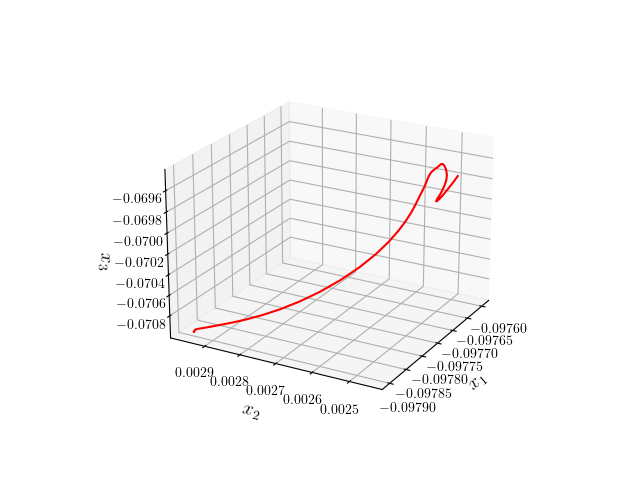}};
    \end{tikzpicture}
    }
    \caption{
        Generalization of DMPs between the learned trajectory on the Panda and the executed trajectory on the daVinci surgical robot.
        The blue dashed line shows the desired (and learned) trajectory, obtained on the Panda industrial manipulator via kinesthetic teaching.
        The red solid line shows the executed extended DMPs \eqref{eqs:extended_dmps_vector} on the daVinci surgical robot.
        The green dash-dotted line shows the adaptation of classical DMPs \eqref{eqs:new_dmps_vector} to the same starting and goal positions as the red line.
        }
    \label{fig:scaling_panda_davinci}
\end{figure*}
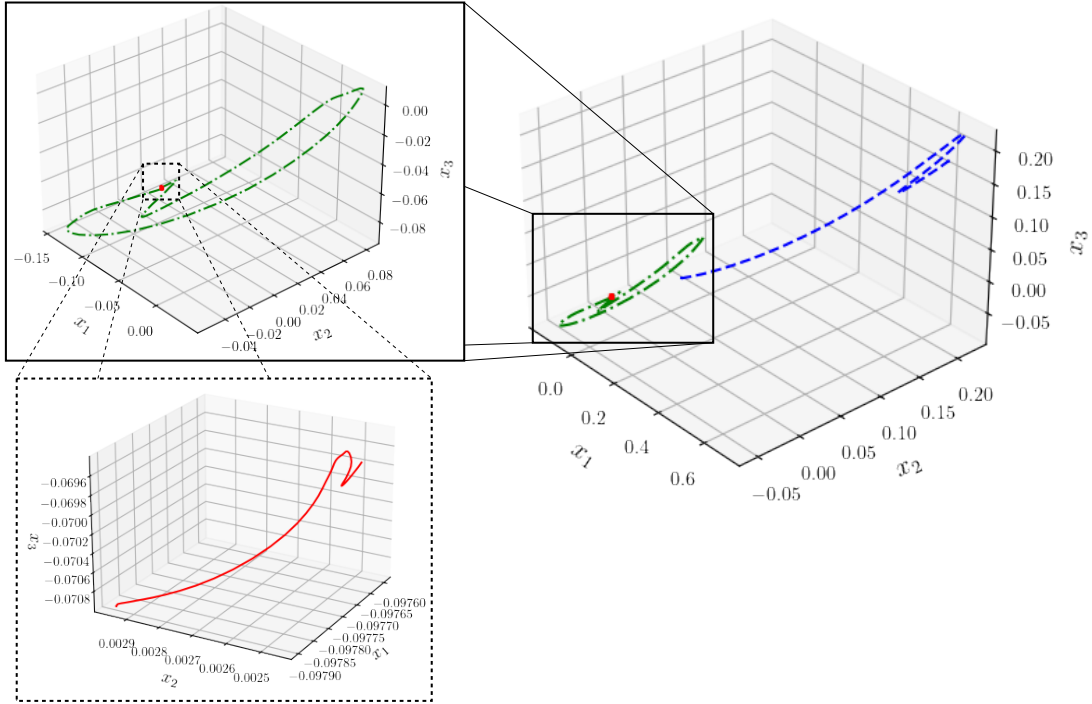

As the second test, we show how extended DMPs allow to `transfer' the desired behavior between different robotic setups.
To do so, we use the DMPs obtained from the execution performed on the Panda industrial manipulator to execute the Peg \& Ring task on the daVinci surgical robot (which setup can be seen in Figure~\ref{subfig:davinci_setup}).
Figure~\ref{fig:scaling_panda_davinci} shows the resulting trajectories for this test for an instance of the \texttt{move} gesture.
Additionally, we plot the trajectory obtained by adapting classical DMPs \eqref{eqs:new_dmps_vector} to the new starting and goal positions.
This experiment shows that extended DMPs \eqref{eqs:extended_dmps_vector} are able to generalize to a very different length-scale, while maintaining a shape similar to the learned behavior, effectively permitting to transfer movement from the industrial manipulator to the surgical robot.
Moreover, it is possible to observe that the trajectory obtained using classical DMPs fails in adapting to the new length-scale, generating a behavior that cannot be executed with the surgical robot.

% As it is shown in Figure~\ref{fig:scaling_panda_davinci} for the \texttt{move} gesture, the DMP is able to generalize to a very different length-scale, while maintaining a shape similar to the learned behavior, effectively permitting to transfer movement from the industrial manipulator to the surgical robot.

% ---------------------------------------------------------------------------- %
% REGRESSION
% ---------------------------------------------------------------------------- %

\section{Learning from Multiple Trajectories}\label{sec:regression}

DMPs have been used to learn trajectories from a single demonstration, because a set of demonstrated trajectories, in general, does not have the same starting and ending points, making the learning phase highly non-trivial.

So-called \emph{Stylistic DMPs} \cite{MHM10, MHM11} were developed to learn from multiple demonstrations by introducing an additional variable, called \emph{style parameter}, and by making the execution dependent on this new variable.
This approach is useful when the ``style'' is needed to describe a trajectory (for instance, the trajectory may depend on the height of an obstacle).
However, when the style is not an issue, this approach introduces additional and undesired variables that increase the complexity of the problem.

\subsection{Linear Regression Over Aligned DMPs}

We propose a technique that allows to extract a unique set of weights $ \vect{w}_i \in \RR ^ d $, $i = 0, 1, \ldots, N$, from a set of multiple observations as a single linear regression problem.

We consider a set of $M$ \emph{demonstrated trajectories} \( \set { \vect{x}^{(j)}(t) , t \in \left[ t_0^{(j)}, t_1^{(j)} \right] }_{j = 1}^M \).
Using the technique introduced in Section~\ref{sec:inv_theory} we transform each trajectory in such a way that all the starting and ending points are the same.
We choose $\vect{x}_0 = \vect{0}$ and $\vect{g} = \vect{1}$, but we emphasize that any choice satisfying $ \vect{x}_0 \neq \vect{g} $ would give the same results.
We compute the roto-dilatation matrices, for $j = 1,2,\ldots,M$,
\begin{equation}
    \label{eq:transformation_pre_regression}
    \mtrx{S}^{(j)} = \frac{ \norm{ \vect{1} - \vect{0} } }{ \norm{ {\vect{x}}^{(j)}\br{t_1^{(j)}} - {\vect{x}}^{(j)} \br{t_0^{(j)}} } } \, \mtrx{R}_{ \widehat{{\vect{x}}^{(j)} \br{t_1^{(j)}} - {\vect{x}}^{(j)} \br{t_0^{(j)}} }} ^ { \widehat{\vect{1} - \vect{0} } },
\end{equation}
and use these matrices to create the new set of transformed trajectories
\[ \set{ \check{\vect{x}}^{(j)} (t) = \mtrx{S}^{(j)} {\vect{x}} ^ {(j)}(t),t \in \left[ t_0^{(j)}, t_1^{(j)} \right] }_{j = 1, 2, \ldots, M}. \]

\noindent
Next, we perform a \emph{time scaling}, so that each curve has $ [0, T] $ as time domain, for a given $ T>0 $.
To do so, we define
\begin{equation}
    \label{eq:time_alignment_formula}
    \tilde{\vect{x}} ^ {(j)} (t) = \check{\vect{x}}^{(j)} \br{ \frac{t_1^{(j)} - t_0^{(j)} }{T} \, t + t_0^{(j)} }.
\end{equation}
It's easy to see that
\begin{align*}
    \tilde{\vect{x}} ^{(j)} (0) & = \check{\vect{x}} ^{(j)} \br { t_0 ^ {(j)} } ,\\
    \tilde{\vect{x}} ^{(j)} (T) & = \check{\vect{x}} ^{(j)} \br { t_1 ^ {(j)} } .
\end{align*}
From these two step, we obtain a new set of `transformed' curves $ \{ \tilde{\vect{x}}^{(j)}(t), t \in [0, T] \}_{j=1}^{M} $, each with time domain $ [0, T] $, and $ \vect{0} $ and $ \vect{1} $ as starting and ending points respectively.

\noindent
Equation~\eqref{eq:new_dmp_vect_acc} allows to compute the set of forcing terms $ \{ \vect{f}^{(j)}(s) \}_{j = 1,2, \ldots, M} $, and then we are able to compute the set of weights $ \{ \vect{w}_i \} _{i = 0,1,\ldots, N} $ that minimizes the sum of the squared errors (w.r.t. the $L_2$-norm) between the function generated using \eqref{eq:forcing_term} and the forcing terms $ \vect{f}^{(j)} (s) $, $j=1,2,\ldots,M$.

\noindent
For this purpose, we decompose the problem in each independent component.
For each component $p = 1,2,\ldots,d$, we look for the \emph{weight vector} $\vectgreek{\omega}^\star = [\omega_0 ^ \star, \omega_1 ^ \star, \ldots, \omega_N ^ \star] \transpose \in \RR^{N + 1} $ satisfying
\begin{equation*}
    \label{eq:regression_problem}
    \vectgreek{\omega}^\star = \arg \min_{\vectgreek{\omega} \in \RR^{N + 1}} \underbrace{ \sum_{j=1}^{M} \norm{ \frac{ \sum_{i=0}^{N} \omega_i \varphi_i(s) }{ \sum_{i=0}^{N} \varphi_i(s) } \, s - f_p^{(j)} (s) }_2^2}_{F(\vectgreek{\omega})},
\end{equation*}
where $ f_p^{(j)} ( s ) $ denotes the $p$-th component of $ \vect{f}^{(j)} (s) $ and is computed as
\begin{equation}
    \label{eq:forcing_term_new}
    f_p^{(j)} \big(s(t)\big) = \frac{\dot{v}_p^{(j)} + D_p v_p}{K_p} - \big( g_p^{(j)} - x_p(t) \big) + (g_p - x_{0,p})s(t),
\end{equation}
with $ s(t) = \exp(-\alpha t) $.

\noindent
The existence and uniqueness of a minimum for $F$ is guaranteed by its continuity and strict convexity.
Moreover, since $ F $ is smooth, $\vectgreek{\omega} ^ \star$ is the vector that nullifies its gradient: \( \nabla_{\vectgreek{\omega}} F( \cdot )|_{ \vectgreek{\omega}^\star} = \vect{0} \).
Let us denote with $ \vect{G} $ the gradient of $F$: $ \vect{G}( \cdot ) \doteq \nabla_{\vectgreek{\omega}} F( \cdot ) : \RR^{N+1} \to \RR^{N+1} $.
Its $h$-th component, denoted by $G_h$ is (recalling that \( \norm{\zeta (s)}_2^2 = \int_{s_0}^{s_1} (\zeta(s))^2 \diff s \))
\begin{align*}
    G_{h} (\vectgreek{\omega}) & = \pderiv{F}{\omega_h} (\vectgreek{\omega}) \nonumber \\
        & \begin{aligned}
        & = \sum_{j = 1}^{M} \, \bigintsss_{s_0}^{s_1} 2 \br{ \frac{ \sum_{i=0}^{N} \omega_i \varphi_i(s) }{ \sum_{i=0}^{N} \varphi_i(s) } \, s - f_p^{(j)} (s) } \\
                & \qquad\qquad\qquad\qquad\qquad\qquad\quad \br{ \frac{\varphi_h(s)}{\sum_{i=0}^{N} \varphi_i(s)}\, s }\diff s.
        \end{aligned}
\end{align*}
Function $G_h$ is linear in $ \vectgreek{\omega} $, thus the minimization problem can be written as a linear system:
\begin{equation}
    \label{eq:linear_problem}
    \mtrx{A}\,\vectgreek{\omega} = \vect{b}.
\end{equation}
The component in row $h$ and column $k$ of $\mtrx{A}$, $a_{hk}$, is, for $h,k \in \{ 0,1,\ldots,N \}$,
\begin{align}
    a_{h k} & = \pderiv{G_h}{\omega_k} ( \omega ) \nonumber \\
        & = \sum_{j=1}^{M} \, \bigintssss_{s_0}^{s_1} 2\frac{\varphi_h (s) \varphi_k (s)}{ \br{ \sum_{i=0}^{N} \varphi_i(s) }^2 } \, s^2 \diff s \nonumber \\
        & = 2 M \, \bigintssss_{s_0}^{s_1} \frac{\varphi_h (s) \varphi_k (s)}{ \br{ \sum_{i=0}^{N} \varphi_i(s) }^2 } \, s^2 \diff s, \label{eq:regression_problem_linear_part}
\end{align}
while the $h$-th component of the vector $\vect{b}$, $ b_h $, is, for $ h = 0,1,\ldots,N $,
\begin{align}
    b_h & = \sum_{j=1}^{M} \, \bigintssss_{s_0}^{s_1} 2 \, \frac{ \varphi_h(s) }{\sum_{i=0}^{N} \varphi_i(s)} \, f_p^{(j)} (s) \, s \diff s . \label{eq:regression_problem_affine_part}
\end{align}
Using any quadrature formula (e.g. Simpson's rule), the integrals in equations \eqref{eq:regression_problem_linear_part} and \eqref{eq:regression_problem_affine_part} can be computed for each $h, k = 0,1, \ldots, N$ giving matrix $ \mtrx{A} $ and vector $ \vect{b} $ of \eqref{eq:linear_problem}.
Thus, we can solve the linear problem and obtain $\vectgreek{\omega} ^ \star$.
We recall that this procedure has to be repeated for each component $p = 1,2,\ldots, d$.

\begin{algorithm}[t]
    \caption{Regression Over Multiple Demonstrations}
    \begin{algorithmic}[1]
        \Require{Set of desired trajectories $ \{ \vect{x}^{(j)} (t) \}_{j=1,2,\ldots, M} $, set of basis functions $ \{ \varphi_i(s) \}_{i=0,1,\ldots,N} $}, DMPs hyperparameter $ \mtrx{K, D} $ and $ \alpha $, final time $T > 0$;
        \Ensure{Set of weights $\{ \vectgreek{\omega}^{(p)} \}_p $ for each direction $ p = 1,2,\ldots, d $.}
        \LeftComment{Align the trajectories (both temporally and spatially)}
        \For{$j = 1,2 \ldots, M$}
            \State{Compute \( \mtrx{S}^{(j)} \) using \eqref{eq:transformation_pre_regression}}
            \State{Compute \( \check{\vect{x}}^{(j)}(t) = \mtrx{S}^{(j)} \vect{x} (t) \)}
            \State{Compute \( \tilde{\vect{x}}^{(j)} (t) \) using \eqref{eq:time_alignment_formula}}
        \EndFor
        \State{Evaluate the canonical system extrema \( s_0 = 1, s_1 = \ee ^ {- \alpha T} \)}
        \LeftComment{Compute the matrix $\mtrx{A} = [a_{h k}]_{h,k = 0,1,\ldots, N}$}
        \For{$ h = 0, 1, \ldots, N $}
            \State{\( a_{hh} = 2 M \int_{s_0}^{s_1} \frac{ \varphi_h(s) ^ 2}{ \br{\sum_{i=0}^{N} \varphi_i(s) }^2 } s^2 \diff s\)}
            \For{$ k = h + 1, h + 2, \ldots, N $}
                \State{\( a_{hk} = 2 M \int_{s_0}^{s_1} \frac{ \varphi_h(s) \varphi_k(s)}{ \br{\sum_{i=0}^{N} \varphi_i(s) }^2 } s^2 \diff s\)}
                \State{\( a_{kh} = a_{h k} \)}
            \EndFor
        \EndFor
        \LeftComment{For loop along the directions}
        \For{$ p = 1,2,\ldots, d $}
            \State{For each trajectory $ \tilde{\vect{x}}^{(j)}(t) $ compute the forcing term $ f^{(j)}(s) $ for direction $p$ using \eqref{eq:forcing_term_new}.}
            \LeftComment{Compute the vector $ \vect{b} = [b_h]_{h = 0,1,\ldots, N} $}
            \For{$ h = 0, 1, \ldots, N $}
                \State{\( b_h =  \sum_{j=1}^{M} \int_{s_0}^{s_1} 2 \, \frac{ \varphi_h(s) }{\sum_{i=0}^{N} \varphi_i(s)} \, f^{(j)} (s) \, s \diff s \)}
            \EndFor
            \State{Solve the minimization problem: \( \mtrx{A} \bm{\omega} ^{(p)} = \vect{b} \)}
        \EndFor
    \end{algorithmic}
    \label{alg:dmp_regression}
\end{algorithm}

\begin{figure}[t]
    \centering
    \subfloat[Synthetic test before rescaling.\label{subfig:synth}]{\includegraphics[width=0.8\columnwidth]{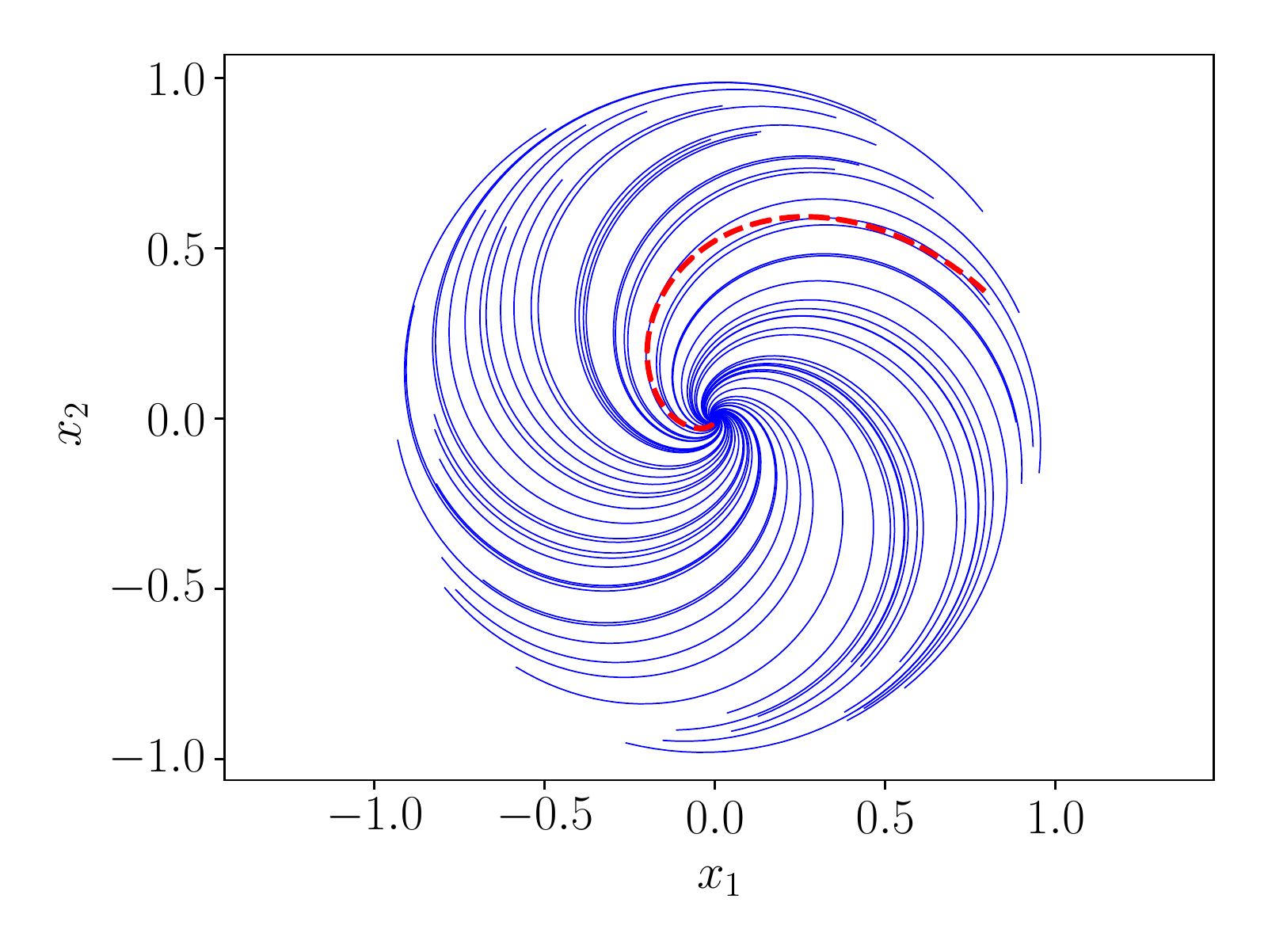}}
    % \hspace{2cm}
    \\
    \subfloat[Synthetic test after rescaling.\label{subfig:synth_scale}]{\includegraphics[width=0.8\columnwidth]{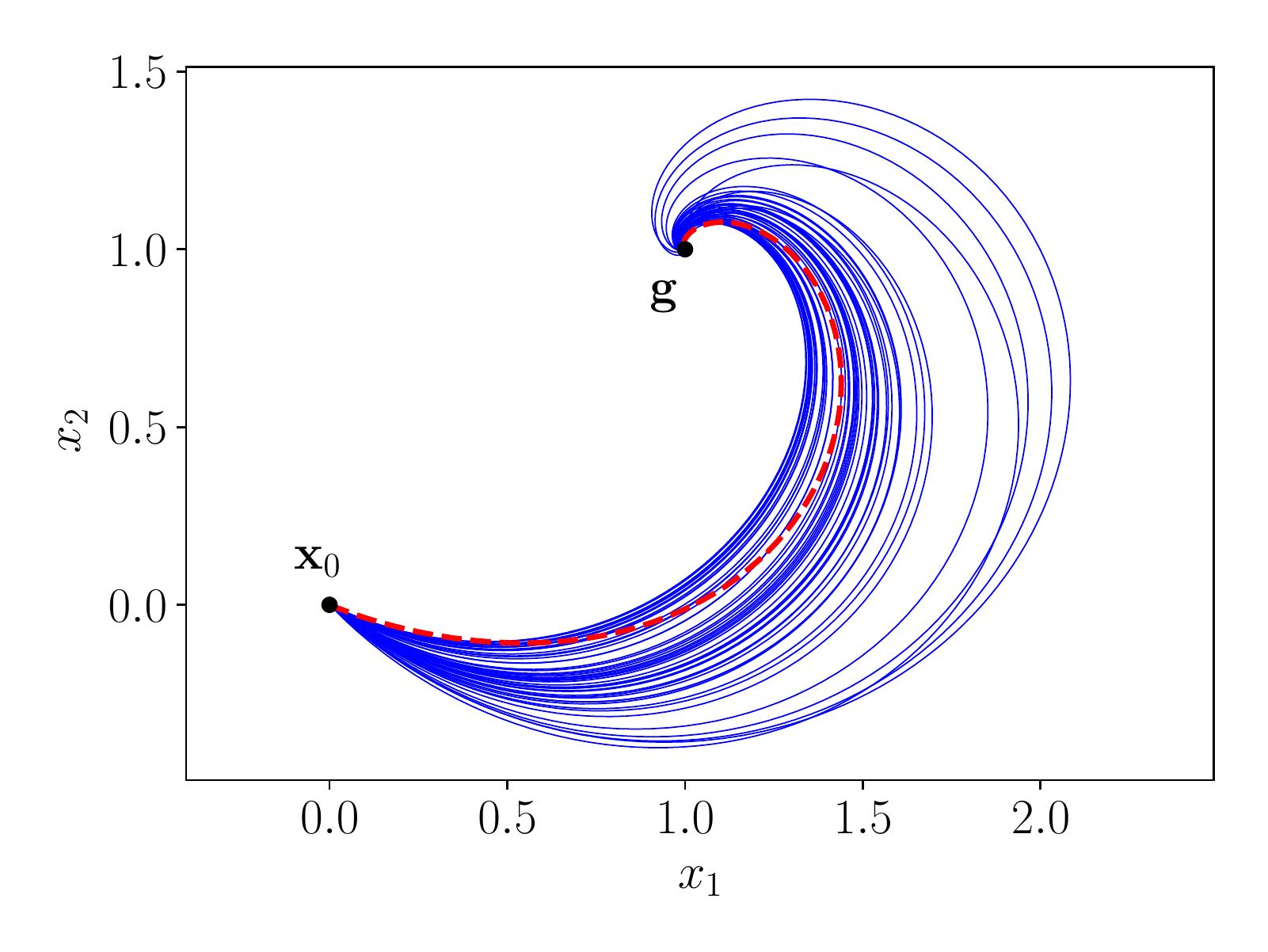}}
    \caption{Tests for our proposed approach to perform DMPs learning from multiple trajectories.
    The trajectories are obtained by integrating \eqref{eq:dyn_sist_test} with different initial conditions.
    In both plots the solid lines represent the demonstrated trajectories, while the dashed one is an execution of the learned DMP.
    The tests are plotted both before (Figure~\ref{subfig:synth}) and after (Figure~\ref{subfig:synth_scale}) the spatial scaling step.
    In Figure~\ref{subfig:synth_scale}, the dot represents the initial position $ \vect{x}_0 = (0,0)$, while the star marks the goal $\vect{g} = (1,1)$.
    % The tests on the real robot (Figure~\ref{subfig:move}, \ref{subfig:carry}) are plotted only after the rescaling step.
    }
    \label{fig:regression_demo}
\end{figure}

The algorithm to perform regression is summarized in Algorithm~\ref{alg:dmp_regression}.

\begin{rmk}
    Although we presented the algorithm using mollifier-like basis functions $\{\varphi_i\}_{i=0,1,\ldots,N}$, this algorithm works for any choice of the set of basis functions.
\end{rmk}

\begin{rmk}
    This approach does not encode information about `styles', differently from \cite{MHM11}.
    Thus, the generalization of the DMP depends only on the starting and goal positions $\vect{x}_0$ and $\vect{g}$.
\end{rmk}

\begin{rmk}
    The presented method performs classical linear regression over a set of observed trajectories, and it gives a non-probabilistic result, differently from other approaches, such as Probabilistic Movement Primitives \cite{PDPN13} and Kernelized Movement Primitives \cite{HRSC19}, which are purely probabilistic approaches.
\end{rmk}

\subsection{Results}\label{sec:regression_results}

In this Section, we test the ability to learn from multiple demonstrations by performing both synthetic tests and experiments with real robots.

\noindent
As first synthetic test, we create a set of trajectories $ \big\{{\vect{x}} ^{(j)}(t) = \big( x^{(j)}_1(t), x^{(j)}_2(t) \big) \big\}_j $ by integrating a known dynamical system.
The second synthetic test is performed using a well-know dataset containing trajectories extracted from surgical tasks.

On the real setup, we use real executions to automate the \emph{Peg \& Ring} task on poth the Panda industrial manipulator and the daVinci surgical robot.
In the experiment on the daVinci, we make the task more challenging by integrating extended DMPs \eqref{eqs:extended_dmps_vector} with obstacle avoidance methods.

\subsubsection{Synthetic Test.}

\begin{figure}[t]
    \centering
    \includegraphics[width=0.9\columnwidth]{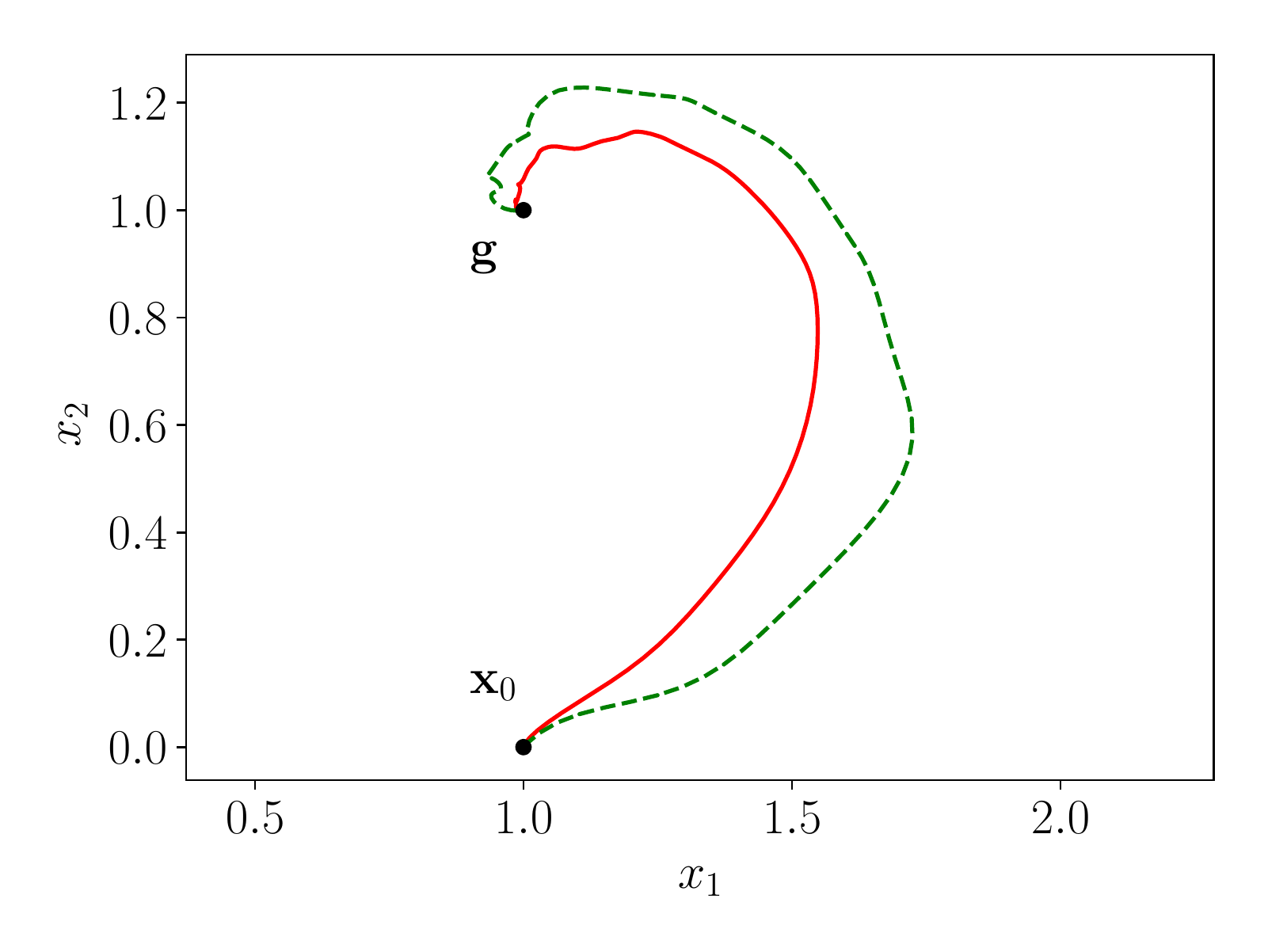}
    \caption{Comparison of two DMPs obtained by a noisy data-set.
    The solid red line shows the trajectory obtained using the proposed method for regression over multiple observations.
    The dashed green line shows the execution of a DMP learned from one sample of the dataset.
    }
    \label{fig:comparison_noise}
\end{figure}

For the test with synthetic data, we generate the trajectories computing the solution of the dynamical system
\begin{equation}
    \label{eq:dyn_sist_test}
    \begin{cases}
        \dot{x}_1 = x_1 ^ 3 + x_2 ^ 2 \, x_1 - x_1 - x_2 \\
        \dot{x}_2 = x_2 ^ 3 + x_1 ^ 2 \, x_2 + x_1 - x_2
    \end{cases},
\end{equation}
which is known to have a \emph{alpha-limit} on the circumference of radius $1$ and an attractive equilibrium at the origin.
The set of (fifty) trajectories is generated by choosing a random angle $\theta_0 \in [0, 2\pi)$ and a random radius $\rho_0 \in (0.8, 1)$, and then setting as initial position \( x_1(0) = \rho_0 \cos (\theta_0) \), \( x_2(0) = \rho_0 \sin (\theta_0) \).
Then the dynamical system is integrated using the \emph{classic fourth-order Runge-Kutta method} up to a random final time $ T \in (5, 10) $.
Choosing a final time greater than $5$ allows the system to reach the origin with an error smaller than $ 2\% $: \( \| \vect{x} - \vect{0} \| < 0.02 \).
Moreover, having the final time change within executions mimics the time inconsistencies that may arise when learning trajectories from real executions.
After generating the set of observations, we use Algorithm~\ref{alg:dmp_regression} to generate a DMP.
The DMP's hyperparameters are $ \mtrx{K} = K\,\mtrx{I}_2 ,\, \mtrx{D} = D\,\mtrx{I}_2$, with $K=150\, D=2\sqrt{K} \approx 24.49$, and $\alpha = 4$.
Finally, we randomly select an initial point $ \vect{x}_0 $ with $ \|\vect{x}_0\| \in (0.8,1) $ and execute the obtained DMP by setting as goal the attractive equilibrium of the system: $ \vect{g} = (0,0) $.

\noindent
Figure~\ref{fig:regression_demo} shows the results of these tests.
In particular Figure~\ref{subfig:synth} shows the set of trajectories obtained by integrating the ODE \eqref{eq:dyn_sist_test}, together with an execution of the obtained DMP. Figure~\ref{subfig:synth_scale} shows the same trajectories when spatially rescaled to have $ \vect{0} $ and $ \vect{1} $ as starting and ending points respectively.

To highlight the usefulness of learning from a set of observations, we added a Gaussian noise $ \epsilon \sim \mathcal{N} (0, \sigma^2) $ of null mean and variance $ \sigma^2 = 5 \cdot 10 ^ {-5} $ to the set of trajectories obtained by solving the dynamical system \eqref{eq:dyn_sist_test}.
We then use Algorithm~\ref{alg:dmp_regression} to extract a DMP (the hyperparameters are the same as the previous test).
Moreover, we learned a second DMP, with the same hyperparameters, using as desired trajectory a single trajectory of the dataset.

\noindent
In Figure~\ref{fig:comparison_noise} it can be seen that learning from a single sample of the dataset results in a DMP with undesired oscillation.
On the other hand, performing regression heavily reduces these oscillations.

\begin{figure}[t]
    \centering
    \resizebox*{0.9\columnwidth}{!}{
        \begin{tikzpicture}
            \clip(-7.5, 5) rectangle (7, -6.5);
            \node () at (0, 0) {\includegraphics{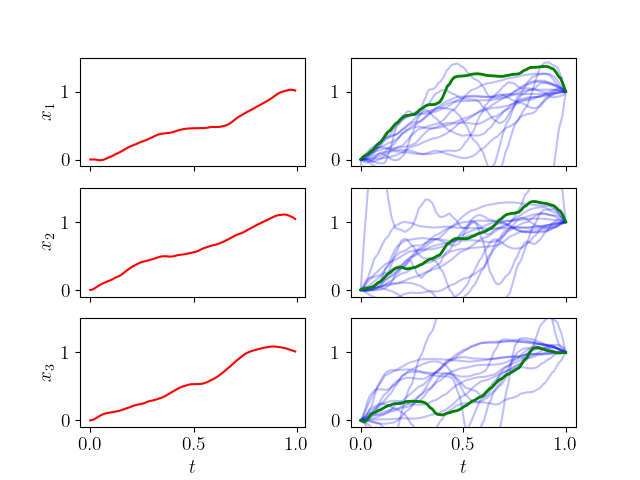}};
        \end{tikzpicture}
    }
    \caption{
        Comparison between two DMPs obtained by the JIGSAW dataset.
        The solid red line (on the left) shows the execution of the DMP obtained via regression.
        The dataset is shown on the right in blue.
        The green line marks a DMP obtained by a single element of the dataset.
    }
    \label{fig:error_regression_real}
\end{figure}

\noindent
To emphasize this aspect, we perform a similar comparison on the JIGSAW \cite{GVRAVLTZKH14} dataset.
In particular, we extract all the occurrences of the gesture `G1' (\emph{reaching needle with the right hand}) and use them to extract a unique DMP via Algorithm~\ref{alg:dmp_regression}.
DMPs' parameters are $ K = 150, D = 2\sqrt{K} \approx 24.49 $, and $ \alpha = 4 $.
Since this gesture involves only the right arm, we extract only the Cartesian components for it and we ignore the left end-effector.
As it can be seen in Figure~\ref{fig:error_regression_real}, in particular from the second component $ x_2 $, oscillations are reduced when a unique behavior is extracted from multiple demonstrations.

\subsubsection{Tests on Real Robots.}

In this Section, we present two experiments on real robotic setups to shows the ability of our proposed method for DMPs regression to extract a unique behavior from multiple observations.

\begin{figure}[t]
    \centering
    \subfloat[Initial setup.\label{subfig:panda_setup_1}]{
        \includegraphics[height=0.4\columnwidth]{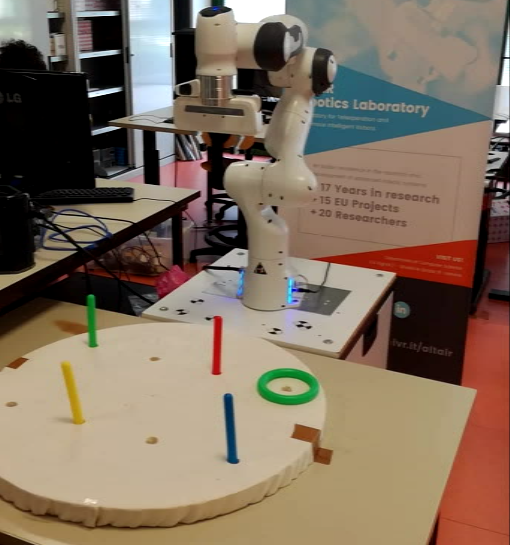}
    }
    \hspace{0.4cm}
    \subfloat[Grasped ring.\label{subfig:panda_grasped}]{
        \includegraphics[height=0.4\columnwidth]{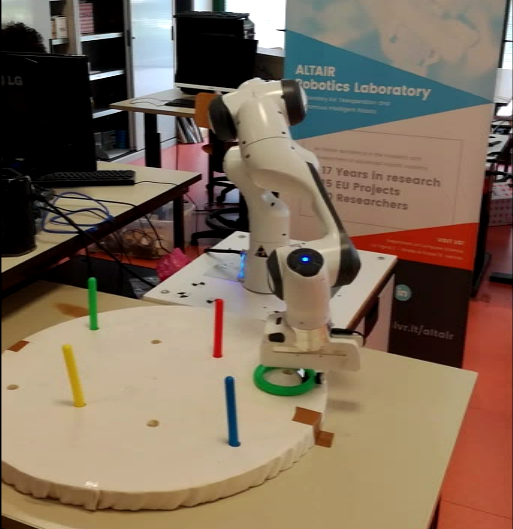}
    }
    \\
    \subfloat[Carrying the ring.\label{subfig:panda_avoided}]{
        \includegraphics[height=0.4\columnwidth]{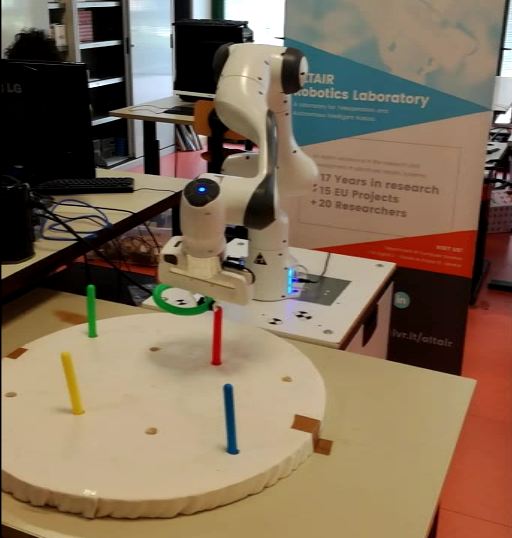}
    }
    \hspace{0.4cm}
    \subfloat[Released ring.\label{subfig:panda_released}]{
        \includegraphics[height=0.4\columnwidth]{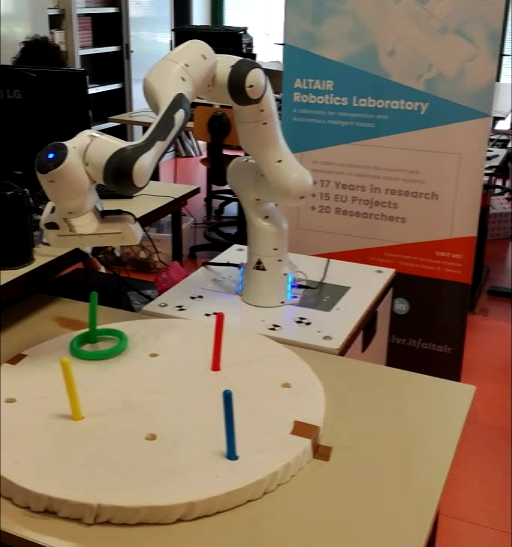}
    }
    \caption{The Peg\&Ring task with the Panda robot.}
    \label{fig:panda_frames}
\end{figure}

\begin{figure}[t]
    \centering
    \subfloat[\texttt{move}\label{subfig:move}]{
        \resizebox{0.8\columnwidth}{!}{
            \begin{tikzpicture}
                \clip (-5.75, 4.5) rectangle (5.25, -5.5);
                \node() at (0,0) {\includegraphics{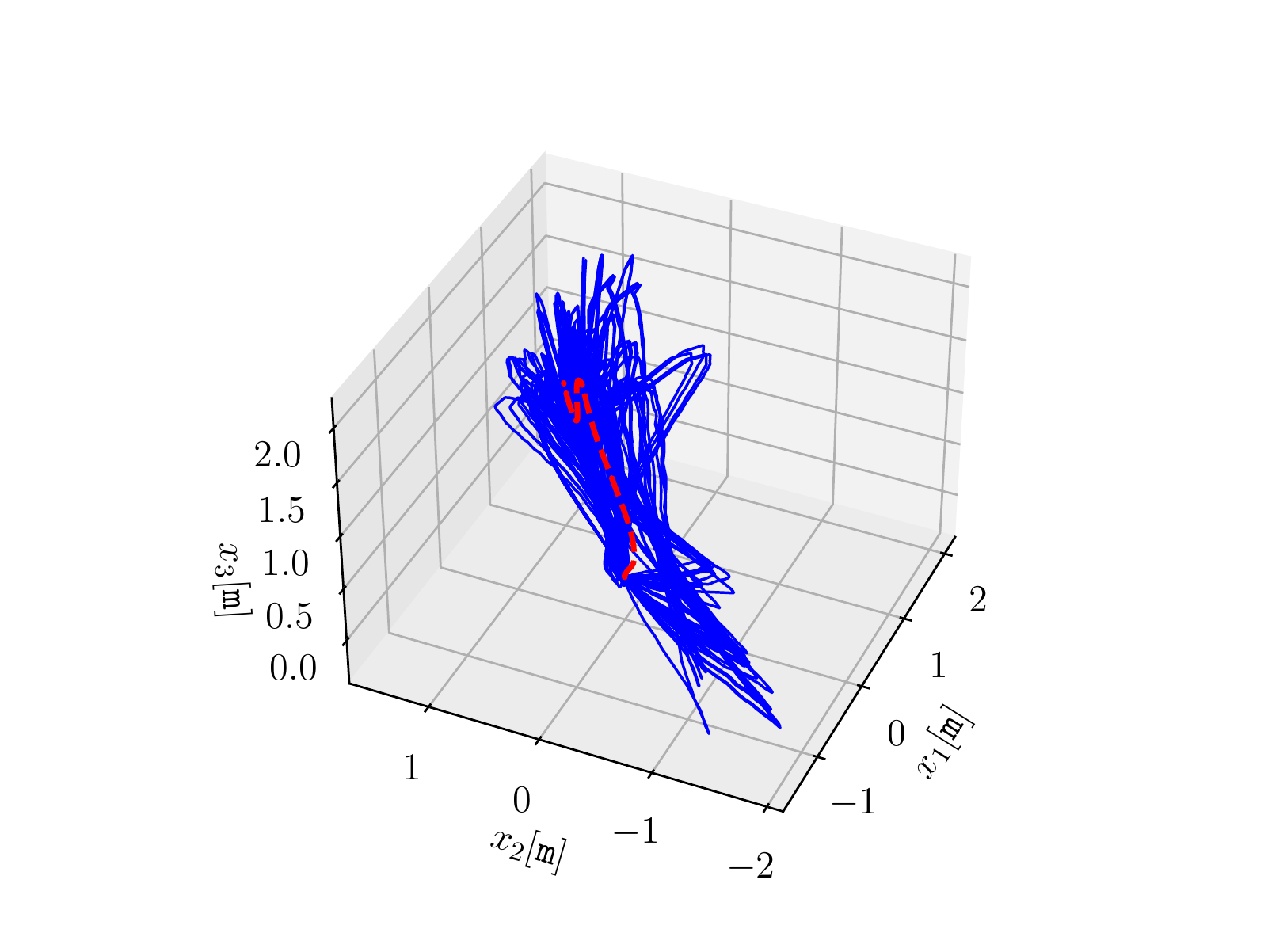}};
            \end{tikzpicture}
        }
        }
    % \hspace{2cm}
    \\
    \subfloat[\texttt{carry}\label{subfig:carry}]{
        \resizebox{0.8\columnwidth}{!}{
            \begin{tikzpicture}
                \clip (-4.5, 4.5) rectangle (6.5, -5.5);
                \node() at (0,0) {\includegraphics{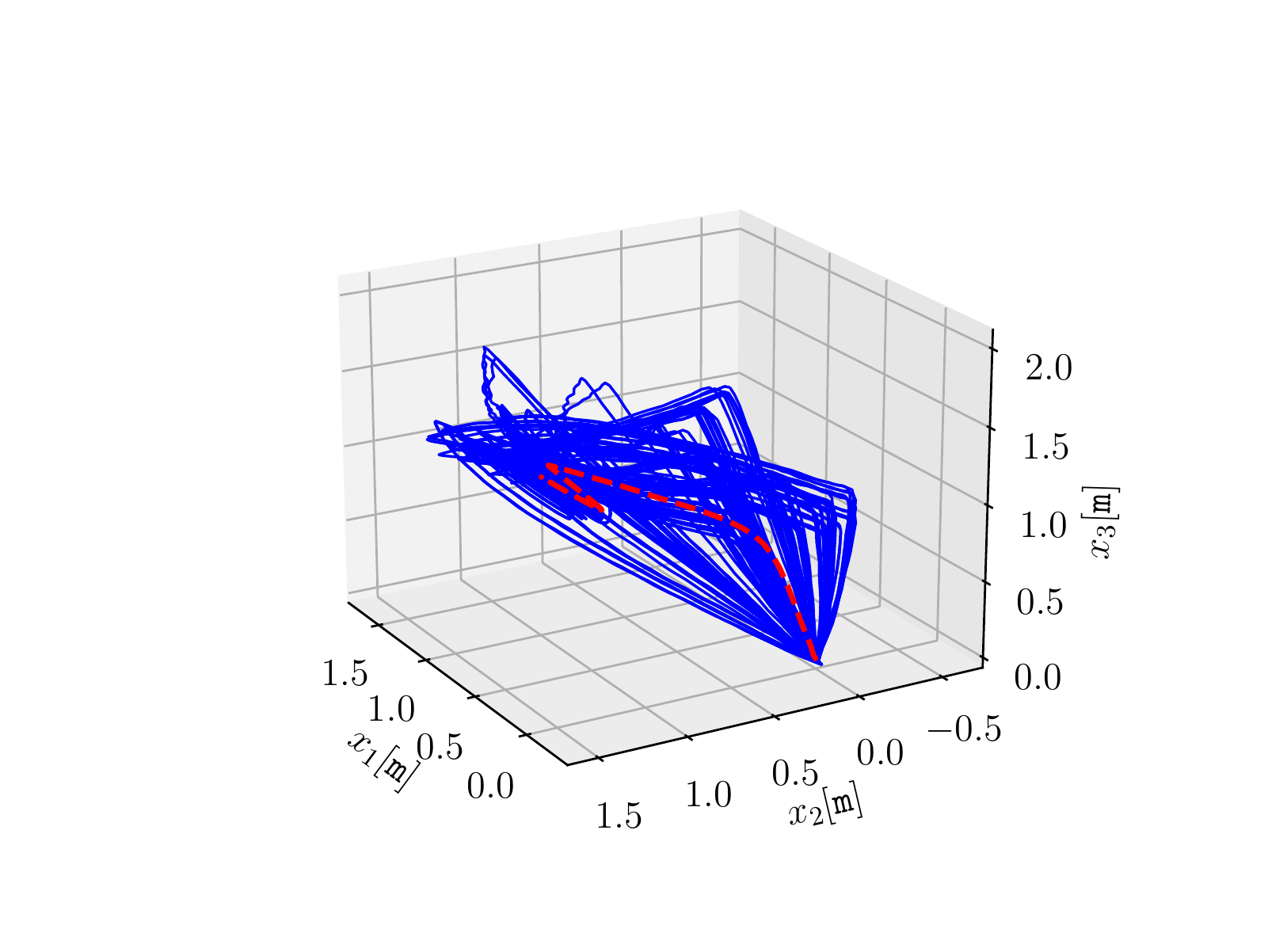}};
            \end{tikzpicture}
        }
        }
    \caption{
        Experiments on the real robot.
        For both gestures (\texttt{move} and \texttt{carry}), the trajectories of the dataset are plotted using solid lines, while the execution of the obtained DMP is plotted using a dashed line.
        All the trajectories are plotted after the rescaling, so that $\vect{x}_0 = \vect{0}$, and $\vect{g} = \vect{1}$ for both gestures.
    }
    \label{fig:regression_panda}
\end{figure}

\paragraph{Test on the Panda Industrial Manipulator}

\begin{figure}[t]
    \centering
    \resizebox{0.8\columnwidth}{!}{
        \begin{tikzpicture}
            \clip (-4.5, 4) rectangle (6, -5);
            \node() at (0,0) {\includegraphics{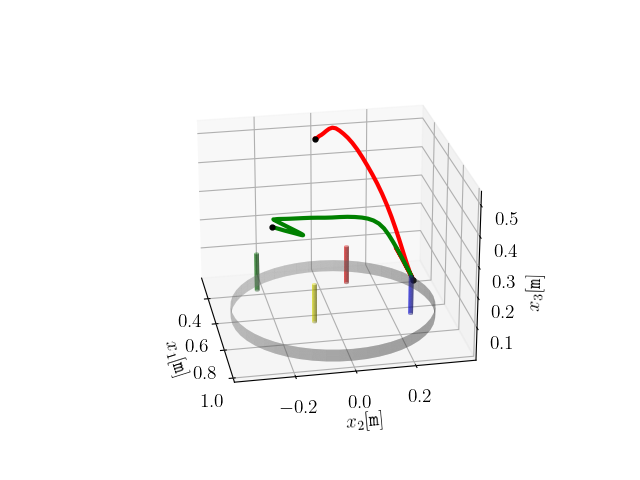}};
        \end{tikzpicture}
    }
    \caption{Plot of the \texttt{move} (in red) and \texttt{carry} (in green) gesture in an automatic execution of the Peg \& Ring task on the Panda robot.}
    \label{fig:task_autom}
\end{figure}

For the experiments on a real setup, we executed the Peg \& Ring task with a Panda industrial manipulator.
The main phases of the task are presented in Figure~\ref{fig:panda_frames}.
The task can be decomposed into two gestures, namely \texttt{move} and \texttt{carry}.
The first gesture is executed to reach the grasping point of the ring with the robot end-effector.
The second gesture, instead, is used to carry the ring toward the same colored peg.
The task uses four colored rings (and, thus, four same-colored pegs) and we execute the task a total of fifty times on the Panda robot, changing the grasping and releasing positions (i.e., the starting and final positions of each gesture).
Since we executed the task fifty times, we obtained a total of 200 samples for both the \texttt{move} and \texttt{carry} gestures.

\noindent
Figure~\ref{subfig:move} and \ref{subfig:carry} show the trajectories of, respectively, \texttt{move} and \texttt{carry} obtained with the industrial manipulator together with an execution of the obtained DMP with parameters $ K = 150, D = 2\sqrt{K} \approx 24.49 $, and $ \alpha = 4 $.
These are plotted after the rescaling so that $\vect{x}_0 = \vect{0}$, and $\vect{g} = \vect{1}$ for both gestures.

\noindent
We then used these DMPs to automate the Peg \& Ring task.
We implemented a Finite State Machine which takes as input the order of colors and then automatically executes the task by alternating the \texttt{move} and \texttt{carry} gestures.
The learned DMPs are used to model the Cartesian evolution of the robot's end-effector of the robot.
Figure~\ref{fig:panda_frames} shows the step of the task's execution for the green ring.
Figure~\ref{fig:task_autom} shows the movement of the end-effector.

\paragraph{Test on the daVinci Surgical Robot}

\begin{figure}[t]
    \centering
    \includegraphics[width=0.9\linewidth]{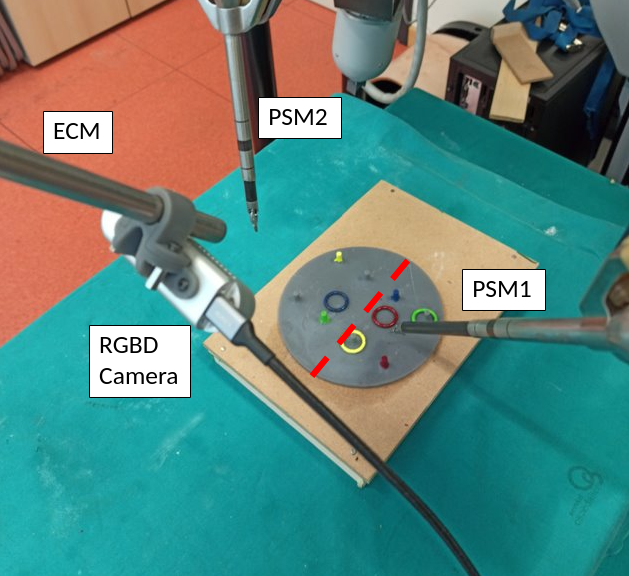}
    \caption{
        daVinci setup for the bimanual Peg \& Ring task.
        The red line splits the workspace in left and right portion.
    }
    \label{fig:davinci_split}
\end{figure}

\begin{figure}[t]
    \centering
    \subfloat[Original dataset.]{
        \resizebox{0.8\columnwidth}{!}{
            \begin{tikzpicture}
                \clip (-1.5, 1.2) rectangle (1.5, -1.25);
                \node () at (0,0) {\includegraphics[width=0.5\linewidth]{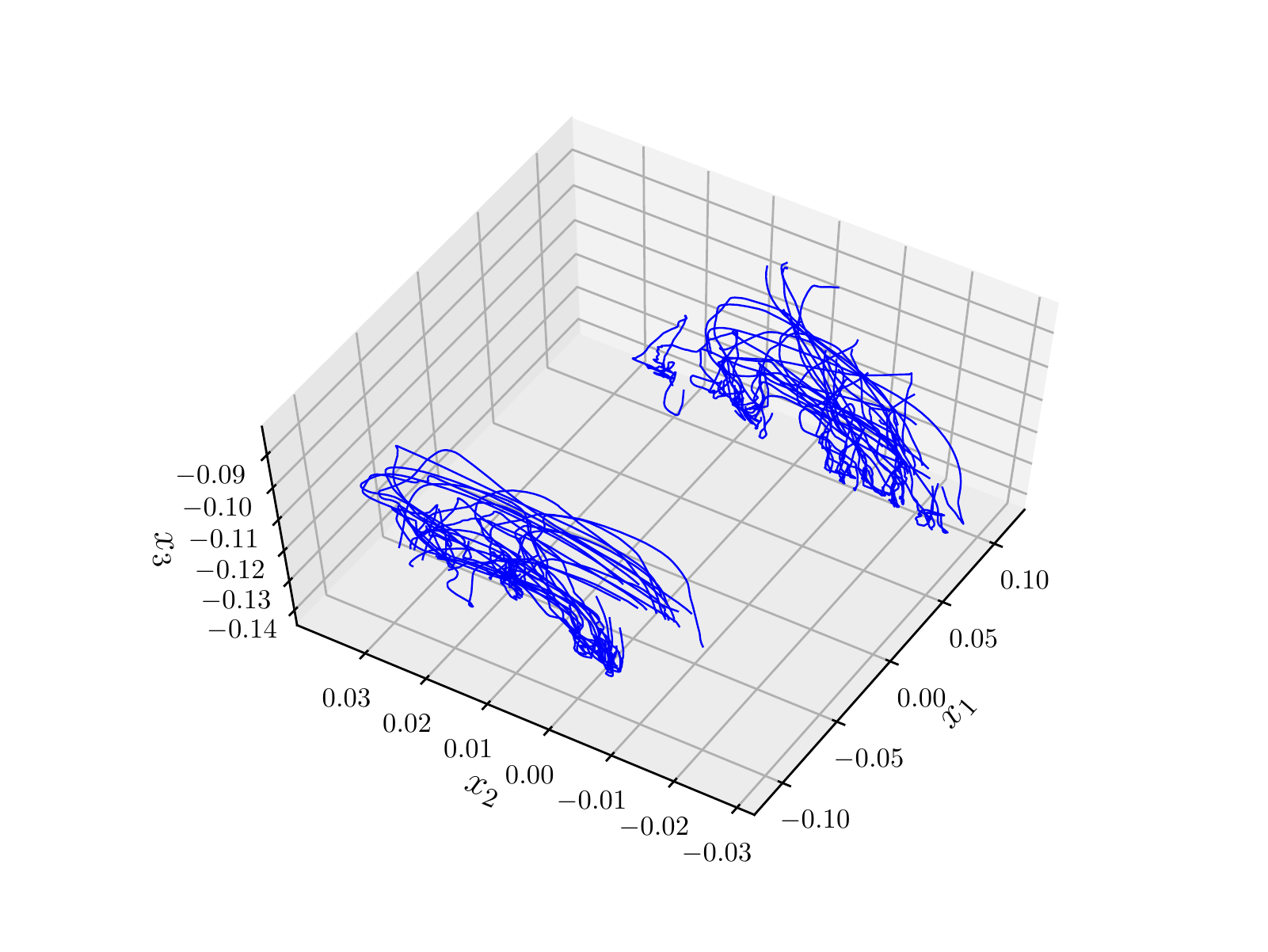}};
            \end{tikzpicture}
        }
    }
    % \hspace{2cm}
    \\
    \subfloat[Roto-dilatated dataset (in blu) and obtained DMP (in dashed red).]{
        \resizebox{0.8\columnwidth}{!}{
            \begin{tikzpicture}
                \clip (-1.5, 1.2) rectangle (1.5, -1.25);
                \node () at (0,0) {\includegraphics[width=0.5\linewidth]{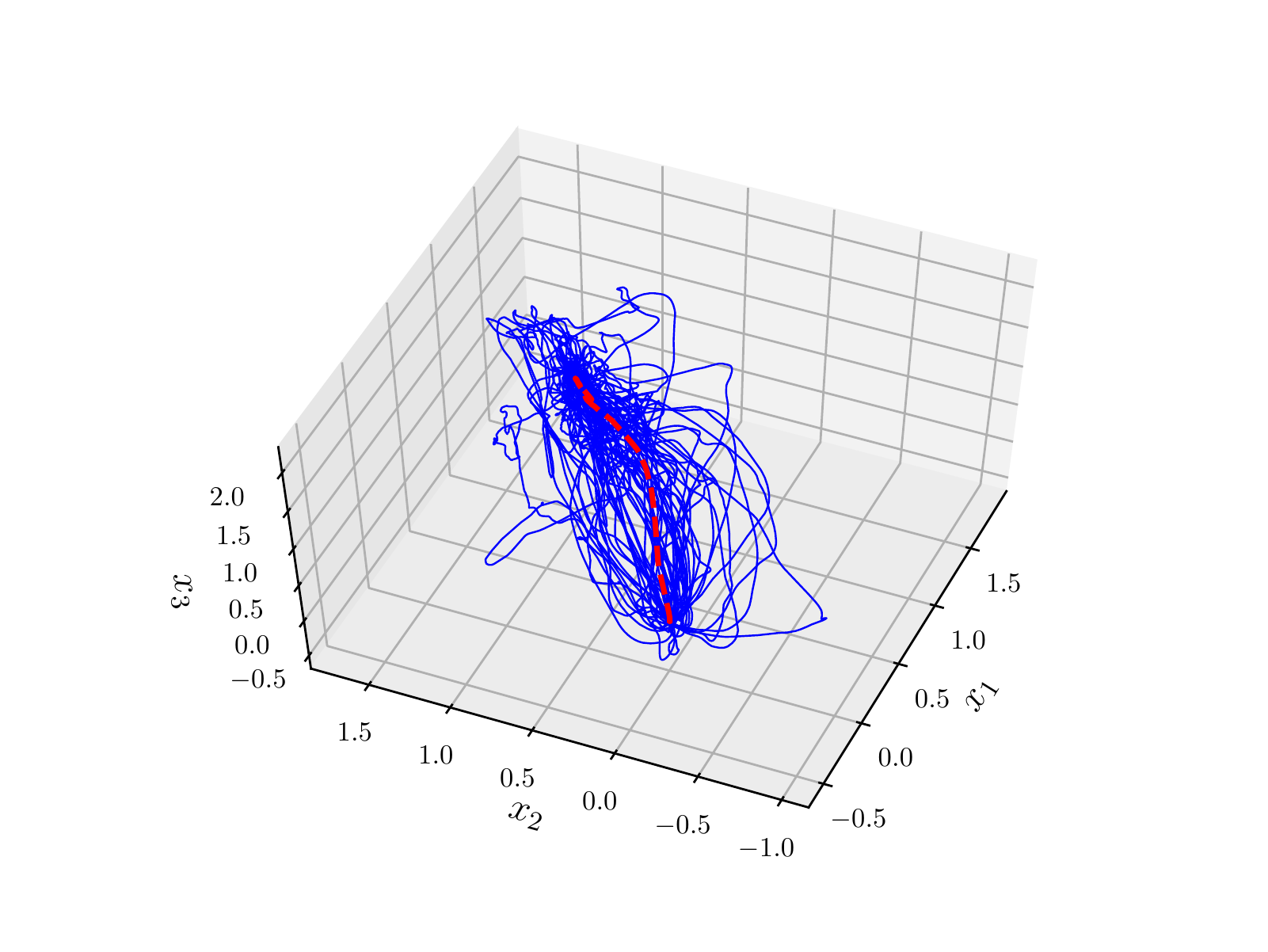}};
            \end{tikzpicture}
        }
    }
    \caption{
        Trajectory samples for the \texttt{move} gesture of the Peg \& Ring task on the daVinci surgical robot.
    }
    \label{fig:regression_test_dvrk}
\end{figure}

As a second experiment on real robots, we perform the Peg \& Ring task also on the daVinci surgical manipulator (which setup is shown in Figure~\ref{fig:davinci_split}).
In this case, the task is more complicated since it is executed with two end-effectors, and an additional movement, called \texttt{pass} has to be learned.
This gesture is used to transfer the ring from one end-effector to the other.
The workspace is split into two parts (see Figure~\ref{fig:davinci_split}), left and right and each end-effector must move only in one of the portions.
For this reason, if the ring initial position and the target-peg placement are in the same half, the end-effector picking the ring will be the same placing it, and the \texttt{pass} gesture will be unnecessary.
On the other hand, when the ring is in one half of the workspace and the same-colored peg is in the other, the ring will be transferred between end-effectors.

\noindent
In our test, we set the initial placement of the rings in such a way that the arm picking up the ring always differs from the hand placing it.
Thus, the ring has always to be passed from one end-effector to the other.
On this setup, the available workspace is proportionally smaller w.r.t. the robot dimension.
Moreover, the pegs and rings are proportionally bigger.
For this reason, we integrate the extended DMP system \eqref{eqs:extended_dmps_vector} with the obstacle avoidance method presented in \cite{GMCDSF19} in order to avoid collisions with the pegs.
Thus, \eqref{eq:extended_dmps_vector_acc} reads
\begin{equation*}
    \tau \dot{\vect{v}} =
        \mtrx{K} ' (\vect{g} - \vect{x}) -
        \mtrx{D} ' \vect{v} -
        \mtrx{K} ' (\vect{g}- \vect{x}_0)s + \mtrx{K} ' \vect{f}(s) +
        \vect{p} (\vect{x}),
\end{equation*}
where, intuitively, $ \vect{p} (\vect{x}) $ is a `repulsive term' that pushes the trajectory away from the obstacle.

\noindent
Since the \texttt{move} and \texttt{carry} gestures require the movement of a single end-effector at a time, we model both with a $ 3- $dimensional DMP.
On the other hand, the \texttt{pass} gesture requires that the two end-effectors move together towards the center of the workspace.
Thus, this third gesture is modeled using a single $6-$dimensional DMP.
In this way, the two end-effectors share the same canonical system so that their movement is synchronized.
Figure~\ref{fig:regression_test_dvrk} shows the dataset and obtained DMP for the \texttt{move} gesture.

The obtained DMPs allowed to successfully automate the task.

% ---------------------------------------------------------------------------- %
% CONCLUSIONS
% ---------------------------------------------------------------------------- %

\section{Conclusions}\label{sec:conclusions} %FIXME: future works?

In this work, we have highlighted some of the weak aspects of the DMP framework, proposing three major modifications to overcome them.

\noindent
At first, we proposed a new formulation for the set of basis functions.
We proved the proposed mollifier-like basis functions to be good approximators for the forcing term, while providing a faster and more stable learning phase.
Moreover, being compactly supported, this new set of basis functions allows to update a DMP when only a portion of the trajectory has to be modified.
Secondly, we proposed a strategy to exploit the invariance under affine transformations of the DMP formulation \eqref{eqs:new_dmps_vector}.
In particular, we showed how to generalize the learned trajectory to any length scale and any rotation of the reference frame maintaining the qualitative shape of the learned trajectory, specifying that the same approach can be extended to any invertible transformation.
Finally, we solved one of the major issues in DMPs, which is the inability of the original framework to learn a common behavior from multiple observations.

As future work, we aim to extend these improvements to the unit quaternion formulation of DMPs \cite{SFL19}, in order to make more robust also the orientation formulation of DMPs.

\noindent
Moreover, we aim to use the proposed formulation to improve the `Surgical Task Automation Frameworks' presented in \cite{GMNRF19} and \cite{GMRSF20a}, in which DMPs are used to generate surgical gestures.

% Bureaucracy

\section*{Funding}
    This project has received funding from the European Research Council (ERC) under the European Union's Horizon 2020 research and innovation programme, ARS (Autonomous Robotic Surgery) project, grant agreement No. 742671.

% ---------------------------------------------------------------------------- %
% BIBLIOGRAPHY
% ---------------------------------------------------------------------------- %

% \section*{References}
\bibliographystyle{elsarticle-num}
\bibliography{root}

\end{document}